\documentclass[10pt,twocolumn,letterpaper]{article}

\usepackage{wacv}
\usepackage{times}
\usepackage{epsfig}
\usepackage{graphicx}
\usepackage{amsmath}
\usepackage{amssymb}
\usepackage[accsupp]{axessibility}  % Improves PDF readability for those with disabilities.

\usepackage{amsthm}
\usepackage{caption}
\usepackage{subcaption}

\usepackage{url}
\usepackage[T1]{fontenc}

\usepackage{ctable} % for \specialrule command
\usepackage{bm}

\usepackage{amsmath,amsfonts,bm}

\def\eqref#1{equation~\ref{#1}}
\def\1{\bm{1}}

\def\vx{{\bm{x}}}

\DeclareMathAlphabet{\mathsfit}{\encodingdefault}{\sfdefault}{m}{sl}
\SetMathAlphabet{\mathsfit}{bold}{\encodingdefault}{\sfdefault}{bx}{n}

\newcommand{\R}{\mathbb{R}}

\def\wacvPaperID{1030} % Enter the WACV Paper ID here

\wacvfinalcopy % *** Uncomment this line for the final submission

\ifwacvfinal
\fi
\ifwacvfinal
\usepackage[breaklinks=true,bookmarks=false]{hyperref}
\else
\usepackage[pagebackref=true,breaklinks=true,colorlinks,bookmarks=false]{hyperref}
\fi

\begin{document}

%%%%%%%%% TITLE
\title{Geometrically Adaptive Dictionary Attack on Face Recognition}

\author{Junyoung Byun, Hyojun Go, Changick Kim\\
Korea Advanced Institute of Science and Technology (KAIST)\\
{\tt\small \{bjyoung, gohyojun15, changick\}@kaist.ac.kr}
}

\maketitle

\ifwacvfinal
\thispagestyle{empty}
\fi

%%%%%%%%% ABSTRACT
\newcommand{\norm}[1]{\left\lVert#1\right\rVert}
%%%%%%%%% ABSTRACT
\begin{abstract}
CNN-based face recognition models have brought remarkable performance improvement, but they are vulnerable to adversarial perturbations. Recent studies have shown that adversaries can fool the models even if they can only access the models' hard-label output. However, since many queries are needed to find imperceptible adversarial noise, reducing the number of queries is crucial for these attacks. In this paper, we point out two limitations of existing decision-based black-box attacks. We observe that they waste queries for background noise optimization, and they do not take advantage of adversarial perturbations generated for other images. We exploit 3D face alignment to overcome these limitations and propose a general strategy for query-efficient black-box attacks on face recognition named Geometrically Adaptive Dictionary Attack (GADA). Our core idea is to create an adversarial perturbation in the UV texture map and project it onto the face in the image. It greatly improves query efficiency by limiting the perturbation search space to the facial area and effectively recycling previous perturbations. We apply the GADA strategy to two existing attack methods and show overwhelming performance improvement in the experiments on the LFW and CPLFW datasets. Furthermore, we also present a novel attack strategy that can circumvent query similarity-based stateful detection that identifies the process of query-based black-box attacks.
\end{abstract}
%%%%%%%%% BODY TEXT
\begin{figure}[t]
     \centering
     \begin{subfigure}[b]{0.45\textwidth}
         \centering
         \includegraphics[width=\textwidth,trim={0cm 0cm 0cm 0cm},clip]{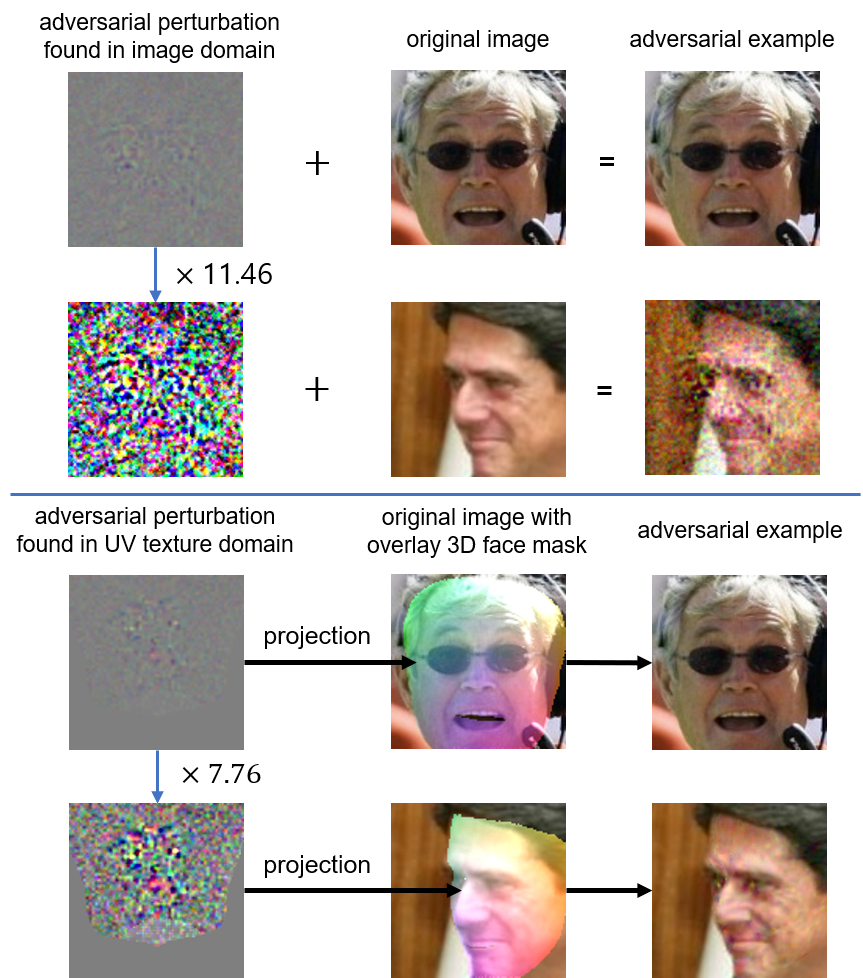}
     \end{subfigure}
        %       \caption{}
        %  \label{fig:fig1a}
        \caption{Illustrations of our intuitions. Finding adversarial perturbations in the UV texture map with 3D face alignment is beneficial for both search space reduction and effective utilization of previous perturbations. We repetitively multiply the previous perturbations by 1.05 until the model misclassifies the image. For visualization, we add 0.5 to the adversarial perturbations and multiply them by 5. The sample images are from the LFW dataset \cite{LFWTech}.}
        \label{fig:fig1}
\end{figure}

\section{Introduction}
Convolutional neural networks have brought remarkable performance improvement in face recognition, but maliciously crafted inputs can fool them with small noise called adversarial perturbations. Since facial recognition can be used in a wide range of areas, such as payment,  finance, and criminal identification, adversarial attacks pose a great threat to their security. Recent studies \cite{chen2020hopskipjumpattack,dong2019efficient,chen2020boosting} have shown that this attack is feasible even when adversaries cannot access the target models' interiors but can only obtain the hard-label predictions. However, decision-based black-box attacks require many queries to find an imperceptible adversarial perturbation for an image. Since making a lot of queries can cause a financial burden, time consumption, and even system administrators' suspicion, reducing the number of queries is crucial for these decision-based attacks.

In line with this research direction,
Dong \textit{et al.} \cite{dong2019efficient} propose EA, an evolutionary algorithm-based decision-based attack on face recognition. It accelerates the convergence of perturbations by updating the sampling distribution of random noise and shows the superior query efficiency over previous attack methods.

However, most decision-based attacks on face recognition \cite{dong2019efficient, chen2020boosting}, including EA, do not take advantage of the characteristics of face recognition in that it optimizes the noise in the whole image area. Even with the same person, the background can change drastically, so we can naturally assume that most of the key features for face recognition locate in facial areas. If we optimize perturbations only for facial areas, we can effectively reduce query waste for optimizing background noise. 

Furthermore, to the best of our knowledge, existing query-based black-box attacks do not utilize previous adversarial perturbations created for other images.  
Prior works on query-efficient black-box attacks reduce the number of required queries by exploiting the gradients of substitute networks \cite{guo2019subspace}, the gradients of previous iterations or the local correlation in an image as prior information \cite{ilyas2018prior}. However, any attempts to recycle adversarial perturbations of previous images are hardly found in literature.

It may be because directly applying a previous adversarial perturbation may not work in most cases, as it is not optimized for the current image. Even if we magnify the previous adversarial perturbation to make it effective against the image, it may harm the convergence rate because of the misalignment of the perturbation to the face. Nevertheless, if the previous adversarial perturbation properly aligns with the face, it may provide a better initial point with a smaller magnitude.

The above two limitations are difficult to overcome with naive solutions. For the background noise optimization, since it is difficult to estimate the gradient towards a target class in a decision-based black-box setting, an adversary needs to initialize an image with the target person's face and gradually reduce the norm of perturbation afterward. However, it is hard to replace the face with the target identity's face via simple copy and paste due to differences in facial pose and size. On the other hand, for the second limitation, one may try traditional face-warping based on Delaunay triangulation \cite{365004} of face landmarks to align the adversarial perturbation, but the warping may not work properly for large poses where some face landmarks are invisible.

In this paper, we exploit 3D face alignment to deal with the above limitations more flexibly and propose \textit{a general attack strategy} for query-efficient black-box attacks on face recognition. Our strategy is to create adversarial perturbations in the UV texture map and project them onto the 3D face in the image. This separation of \textit{the generation of adversarial perturbations} and \textit{the alignment process} enables efficient recycling of previous adversarial perturbations as shown in Fig. \ref{fig:fig1}. Moreover, it reduces unnecessary queries for background noise optimization by limiting the perturbation search space to the facial area that is important for face recognition.

Our major contributions can be listed as follows:

1) We propose Geometrically Adaptive Dictionary Attack (GADA), \textit{a general strategy} for query-efficient black-box attacks on face recognition using 3D face alignment. It remarkably improves query efficiency by limiting the perturbation search space to facial areas of images.

2) GADA can efficiently utilize previous adversarial perturbations created in the common UV texture map since it can align the perturbations to the faces. To the best of our knowledge, this is the first query-based black-box attack that takes advantage of previous adversarial perturbations created for other images.

3) To evaluate memory-based black-box attacks, we also propose a novel evaluation protocol that measures and compares the average number of queries used for finding norm-bounded adversarial perturbations for a predetermined sequence of images. This evaluates how effectively a black-box attack method utilizes previous attack experiences against a target model.

4) We present an attack strategy to evade stateful detection that computes the perceptual similarity of query images to detect the process of query-based black-box attacks. GADA can confuse similarity-based detectors by injecting noise into the background area while gradually reducing perturbations in the facial area. To our knowledge, this is the first effective way to bypass these detection techniques.

5) We demonstrate that GADA brings overwhelming performance improvement when applied to two attack methods, EA \cite{dong2019efficient} and SFA \cite{chen2020boosting}, through experiments on the LFW \cite{LFWTech} and CPLFW \cite{CPLFWTech} datasets. Specifically, for dodging attacks on the LFW dataset, compared to EA, our proposed strategy almost halves the perturbation norm when the query budget is 1K. It also drops the average number of queries required to find an adversarial perturbation whose $\ell_2$ norm is two by more than 2,100.

\section{Background}
\subsection{Face recognition}
Face recognition involves two sub-tasks: face verification and face identification. Face verification is a task of comparing a candidate face to another, and verifying whether the two face images are of the same identity, and face identification is a task to classify an image into one of the gallery's identities. In this paper, we deal with face verification task, but the proposed method can be adapted to face identification task as it is a problem of finding the face with the closest distance in the gallery.

In face verification, a neural network $f$ encodes an input image $\vx$ into a feature vector $f(\vx)\in\R^{D}$, where $D$ indicates the feature dimension. For a pair of images, $\vx_1$ and $\vx_2$, we can compute their $\ell_2$-normalized Euclidean distance (a proxy for cosine distance) as follows.

\begin{equation}
    Dist(\vx_1,\vx_2) =\norm{\frac{f(\vx_1)}{\norm{f(\vx_1)}_2} - \frac{f(\vx_2)}{\norm{f(\vx_2)}_2}}^2_2.
\end{equation}
If $Dist(\vx_1,\vx_2)$ is less than a threshold $\gamma$, the model recognizes that the people in two images represent the same identity. Otherwise, they are considered different. The performance of face verification models has been improved through various angular margin losses ranging from SphereFace\cite{liu2017sphereface}, CosFace \cite{liu2017sphereface}, and ArcFace\cite{deng2019arcface}. These losses are designed to increase the inter-class distance while shrinking the intra-class distance. Recently, Huang \textit{et al.} \cite{huang2020curricularface} further improve the accuracy by incorporating the idea of curriculum learning into their loss function to induce the models to treat easy samples in early stages and hard ones in later.

\subsection{Adversarial setting}
We now construct a black-box threat model that wraps a face verification model. First, let us denote $\vx_1$ as $\bm{x}_A$ and $\vx_2$ as $\bm{x}_S$ to clarify that each of them is owned by an adversary and a server, respectively. 
The target model performs face verification by comparing the query input $\bm{x}_A$ with $\bm{x}_S$ in the server. As a black-box threat model, $\bm{x}_S$ is inaccessible to the attacker, and thus, the adversary can only modify $\bm{x}_A$ and make queries to check the hard-label predictions of the model.
Based on the above setting, we can represent a black-box face verification model that returns a hard label $h\in \{1, 0\}$ for $\bm{x}_A$ as follows:
\begin{equation}
    h_{\bm{x}_S}(\bm{x}_A)=\begin{cases}
      1, & \text{ if $Dist(\bm{x}_A,\bm{x}_S)<\gamma$}\\
      0, & \text{ otherwise}
    \end{cases}.
\end{equation}
In the following, we will omit the subscript of $h$ for notational convenience.

An adversary has a clean image $\bm{x}_A$ and wants to generate an adversarial example that has minimal perturbation while successfully fooling the target model $h$. Then, we can represent the adversary’s objective as follows: 
\begin{equation}
    \arg\min_{\bm{\delta}_q}{\norm{\bm{\delta_q}}_p}
    \text{, s.t. } h(\bm{x}_A+\bm{\delta}_q)\neq h(\bm{x}_A)\text{ and $q \leq Q$},
    \label{eqn:adv}
\end{equation}
where $Q$ is the query budget of the adversary, and $q$ is the number of queries used to make $\bm{\delta}_q$ and $p\geq0$. Unless otherwise noted, we use $p=2$ (i.e., $\ell_2$ norm) for the perturbation norm objective. In this paper, we also assume that pixel values of images are normalized into $[0, 1]$. 
Depending on the value of $h(\bm{x}_A)$ (i.e., whether the pair of images originally represent the same 
identity), we call the attacks differently either dodging attacks ($h(\bm{x}_A){=}1$) or impersonation attacks ($h(\bm{x}_A){=}0$). These two types of attacks differ in their initial values and objectives, and we will explain them in detail in Section \ref{section:proposed}.

\subsection{3D face alignment}
3D face alignment identifies the 3D geometric shape of faces in an image, and it helps to find the semantic meanings of facial pixels. 3D face alignment is widely used in face frontalization \cite{zhou2020rotate} and face presentation attack detection \cite{wang2020deep, qin2019learning} but has not yet been used for adversarial attacks to the best of our knowledge.

Among learning-based 3D face alignment methods, Guo \textit{et al.} \cite{3DDFA_V2} propose 3DDFA\_V2 that shows superior performance in terms of accuracy, speed, and stability of predictions. This method regresses the parameters of the 3D Morphable Model (3DMM) \cite{blanz1999morphable} through a light model. The 3DMM used in \cite{3DDFA_V2} describes the 3D face with PCA, and it can be represented as follows:
\begin{equation}
    \bm{S}=\bar{\bm{S}}+\bm{A}_{id}\bm{\alpha}_{id}+\bm{A}_{exp}\bm{\alpha}_{exp},
\end{equation}
where $\bm{S}$ is the 3D mesh of face model, $\bar{\bm{S}}$ is the mean 3D face shape, $\bm{\alpha}_{id}$ and $\bm{\alpha}_{exp}$ are the facial shape and expression parameters, respectively. The 3DDFA\_V2 model predicts the following parameters of the 3DMM, $\bm{p}=(\bm{R},\bm{\alpha}_{id},\bm{\alpha}_{exp},\bm{t}_{2d})$, where $\bm{R}$ and $\bm{t}_{2d}$ are the rotation matrix and the translation vector, respectively. From the above predicted parameters, the 3D face can be reconstructed as follows.
\begin{equation}
    \bm{V}_{3D}=\bm{R}(\bar{\bm{S}}+\bm{A}_{id}\bm{\alpha}_{id}+\bm{A}_{exp}\bm{\alpha}_{exp})+[\bm{t}_{2d},0]^T.
\end{equation}
Furthermore, the reconstructed 3D Face can be rendered on a 2D image using rasterization with z-buffer as follows. 
\begin{equation}
    \bm{x}^*=Render(\bm{x},\bm{V}_{3D},\bm{C}_V,\bm{Z}),
\end{equation}
where $Render$ function displays the 3D triangular meshes on the image $\bm{x}$ with the vertex coordinates $\bm{V}_{3D}$ and z-Buffer $\bm{Z}$ with vertex colors $\bm{C}_{V}$. $\bm{Z}$ contains the depth of reconstructed 3D face and it helps to avoid rendering the occluded area.

\section{Proposed method}
\label{section:proposed}
As an overview, GADA creates adversarial perturbations in UV texture maps and projects them onto the 3D face obtained by 3D face alignment. A UV texture map is a planar representation of a 3D model's surface used to paint the surface of the 3D model. We can project face textures in a UV map onto a 3D face model's surface through UV mapping. Likewise, we can also extract the face texture from the 3D face mask as a UV texture map according to the relative coordinates of the face. Since GADA is a general attack strategy for query-efficient black-box attacks on face recognition, we explain its general scheme, but it can be adapted to various attacks for improving query efficiency.

\textbf{Initialization of 3D face alignment.} In face verification, it is assumed that each image has a face, and thus, a 3D face mask can be found via 3D face alignment. Since the face position in an image is fixed, there is no need to perform 3D face alignment for each query, so GADA performs 3D face alignment only once at the initial stage to get the vertex coordinates of the 3D face. We also compute $\bm{Z}$ at the initial stage to prevent rendering of occluded areas in the future.

\textbf{UV mapping.}
The adversarial perturbations in the UV texture map contain noise values according to the relative coordinates of the face. By exploiting the UV mapping, adversarial perturbations can be properly projected onto the 3D face with $\bm{C}_{V}$ that retains the noise of RGB colors for the vertices of the 3D face model. It operates as projecting a translucent UV texture map on the 3D face. To project the perturbation of $\bm{C}_V$ onto the image, we use the modified $Render$ function that adds the corresponding vertex color of the point in the 3D face to the original pixel values in rasterization. Since GADA finds adversarial perturbations in the UV space, we also need to use UV $\rightarrow$ $\bm{C}_V$ conversion. For this conversion, we obtain $\bm{C}_V$ from the UV texture map using bilinear interpolation with the UV coordinates. 

\textbf{Rendering details of GADA.} 
When rendering each triangular face of the 3D mesh, the color of an inner point is calculated as the combination of colors of its three vertices weighted by its distances. However, this slows down the convergence of perturbations as the color of one point in an image depends on the three vertices of a triangle. So, for the inner points of each triangle, we use the color of the triangle's first vertex. It improves the convergence rate, especially in $\ell_\infty$ norm-based attacks such as SFA \cite{chen2020boosting}.

\textbf{The reason for the use of the UV texture map.}
Vertex colors are stored in a matrix $\bm{C}_V$ of $N_{V}\times3$, where $N_{V}$ is the number of vertices of the 3DMM. Since we use a dense 3D face model, $N_{V}$ can be larger than the dimension of an image (in our experimental setting, $N_{V}$ is 38,365, and the spatial dimension of an image is $112\times112{=}12,544$). If adversaries try to find the optimal adversarial perturbations in $\bm{C}_V$, such a large search space can degrade query efficiency because many queries are required for search. Meanwhile, existing query-efficient attacks that operate in the image space tend to find perturbations in the down-scaled image to reduce the search space. However, in our case, the adjacency of projected vertices of $\bm{V}_{3D}$ in the image varies from image to image, so it is not easy to utilize common reduced space like the image-space attacks. In addition, if we find perturbations in $\bm{C}_V$, recycling efficiency becomes poor because only a relatively small percentage of the vertex colors are used in an image as our rendering function maps each image pixel to one vertex.
To effectively reduce the search space and facilitate recycling, we find perturbations in the UV texture map and convert it to $\bm{C}_V$ rather than directly searching for them in $\bm{C}_V$. We set the size of the UV texture map as the same size as the image, but this can be set arbitrarily regardless of the image size.

\subsection{Dodging attacks}
For dodging attacks, GADA initializes the UV texture map as a uniformly random image (and subtract the original UV texture as it adds the texture values when rendering) for misclassification of the target model and gradually reduce the perturbation from it.

\textbf{Dictionary attacks.} The adversarial perturbation created in the common UV space can provide a better initial state when attacking other images. We can naturally expect that the perturbation for a person can be effective in different poses of the same identity. Furthermore, We also conjecture that the perturbation for a person can be more effective against similar-looking people than others. From this motivation, when an attack is complete, GADA saves the minimal adversarial perturbation and the face feature in its dictionary. Therefore, if we use dictionary attacks, we initialize an image with a previous perturbation if it is available. When GADA draws a previous adversarial perturbation from its dictionary, it fetches the perturbation corresponding to the feature vector closest to that of the current image. This is because the closer the feature vectors of two images, the more likely they have similarities. After it fetches a perturbation, it repetitively multiplies the perturbation by 1.05 until the target model misclassifies the image pair. 

\newcommand{\figw}{0.45\textwidth}
\newcommand{\figsw}{0.19\textwidth}
\begin{figure}[t]
     \centering
     \begin{subfigure}[b]{\figw}
         \centering
         \includegraphics[width=\figsw,trim={0cm 0cm 0cm 0cm},clip]{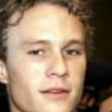}
         \includegraphics[width=\figsw,trim={0cm 0cm 0cm 0cm},clip]{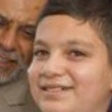}
         \includegraphics[width=\figsw,trim={0cm 0cm 0cm 0cm},clip]{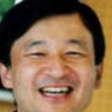}
         \includegraphics[width=\figsw,trim={0cm 0cm 0cm 0cm},clip]{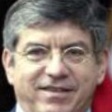}
         \includegraphics[width=\figsw,trim={0cm 0cm 0cm 0cm},clip]{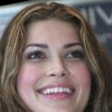}
     \end{subfigure}
          \begin{subfigure}[b]{\figw}
         \centering
         \includegraphics[width=\figsw,trim={0cm 0cm 0cm 0cm},clip]{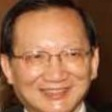}
         \includegraphics[width=\figsw,trim={0cm 0cm 0cm 0cm},clip]{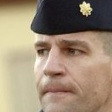}
         \includegraphics[width=\figsw,trim={0cm 0cm 0cm 0cm},clip]{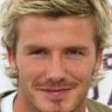}
         \includegraphics[width=\figsw,trim={0cm 0cm 0cm 0cm},clip]{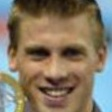}
         \includegraphics[width=\figsw,trim={0cm 0cm 0cm 0cm},clip]{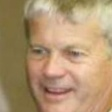}
     \end{subfigure}
          \begin{subfigure}[b]{\figw}
         \centering
         \includegraphics[width=\figsw,trim={0cm 0cm 0cm 0cm},clip]{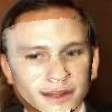}
         \includegraphics[width=\figsw,trim={0cm 0cm 0cm 0cm},clip]{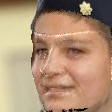}
         \includegraphics[width=\figsw,trim={0cm 0cm 0cm 0cm},clip]{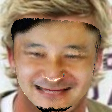}
         \includegraphics[width=\figsw,trim={0cm 0cm 0cm 0cm},clip]{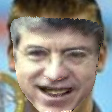}
         \includegraphics[width=\figsw,trim={0cm 0cm 0cm 0cm},clip]{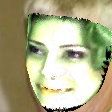}
     \end{subfigure}
        \caption{Examples of initial images of GADA for impersonation attacks on the LFW dataset \cite{LFWTech}.
The top and middle row images are the source and target images, respectively. As shown in the bottom row, the target images' faces are replaced with the source images' faces. We apply various data augmentation to the source images' textures to increase the chance of prediction to the target identity. An example is shown in the rightmost image.}
        \label{fig:fig2}
        %\vspace{-0.2cm}
\end{figure}

\subsection{Impersonation attacks}
In the decision-based black-box setting, an impersonation attack starts from a target identity's image (source image) and gradually update the image to look like the target image.
Note that the target image differs from $\bm{x}_S$, which is inaccessible to attackers. Except for the difference in the initialization of the image, the attack process is similar to that of dodging attacks. Since the proposed GADA strategy applies perturbations only to facial areas, it is necessary to extract the UV face texture map from the target identity's source image to make an initial image. To do this, it extracts $\bm{C}_V$ from the image and converts it into a UV texture map. We first obtain $\bm{C}_V$ from an image using bilinear interpolation with the image coordinates. To convert $\bm{C}_V$ to a UV texture map, we can use the $Render$ function as follows.
\begin{equation}
    \bm{x}_{uv}=Render(\bm{0},\bm{V}_{UV},\bm{C}_V,-10^8\times\bm{1}),
\end{equation}
where $\bm{V}_{UV}$ is the vertex coordinates in the UV space. With the obtained UV texture map, we can replace the target image's face with the source image's face. Since the $Render$ function adds the vertex colors to the corresponding original pixel values, we use the UV texture of the target image subtracted by that of the source image.
We illustrate some initial images of impersonation attacks in Fig. \ref{fig:fig2}.

In impersonation attacks, we do not use the dictionary for the following reasons: (1) if the dictionary does not have the adversarial perturbation for the target identity, it is impossible to make the image to be recognized as the target identity; (2) even if the dictionary has the perturbation for the target identity, the projected perturbation may not work due to changes in the facial pose and scale. In score-based attacks where adversaries can obtain the distance between two facial features, they may exploit the dictionary for impersonation attacks.
%  the image to be recognized as the target identity 

\textbf{Data augmentation on the UV texture map.}
Even if the source image's face is projected on the target image, there may be some cases in which the target model does not recognize it as the target identity due to errors in texture mapping or missing textures. To increase the chance to find an image recognized as the target identity, GADA applies data augmentation on the UV texture map of the source image with random horizontal flip and random color jittering (brightness, contrast, saturation, hue). If it fails to find an image recognized as the target identity even after 200 attempts, it applies the original image-space attack for the image. 

 \newcolumntype{P}[1]{>{\centering\arraybackslash}p{#1}}

\setlength{\tabcolsep}{4pt} % Default value: 6pt
\begin{table*}[t]
\centering
\begin{tabular}{lcccc P{1.5cm}P{1.5cm} cccc P{1.5cm}P{1.5cm}}
\specialrule{.1em}{.05em}{.05em} 
\multicolumn{1}{c}{} & \multicolumn{12}{c}{Dodging attacks} \\
 & \multicolumn{6}{c}{LFW dataset \cite{LFWTech}} & \multicolumn{6}{c}{CPLFW dataset \cite{CPLFWTech}} \\ 
\specialrule{.1em}{.05em}{.05em} 
 & \multicolumn{4}{c}{\begin{tabular}[c]{@{}c@{}}Minimum perturbation\\ norm with query budget\end{tabular}} & \multicolumn{2}{c}{\begin{tabular}[c]{@{}c@{}}Avg. \# queries for\\  perturbation with norm\end{tabular}} & \multicolumn{4}{c}{\begin{tabular}[c]{@{}c@{}}Minimum perturbation\\ norm with query budget \end{tabular}} & \multicolumn{2}{c}{\begin{tabular}[c]{@{}c@{}}Avg. \# queries for\\ perturbation with norm \end{tabular}} \\
Attack method & 1K & 2K & 5K & 10K & 4 & 2 & 1K & 2K & 5K & 10K & 4 & 2 \\ \hline
Sign-OPT \cite{cheng2019sign} & 17.54 & 11.19 & 4.66 & 2.57 & 5738 & 8460 & 16.87 & 10.52 & 4.07 & 2.18 & 5016 & 7498 \\
HSJA \cite{chen2020hopskipjumpattack}& 11.4 & 7.62 & 3.8 & 2.32 & 4692 & 7769 & 11.31 & 7.27 & 3.41 & 2.01 & 4191 & 6833 \\
EA \cite{dong2019efficient}& 13.73 & 7.17 & 2.74 & 1.44 & 3561 & 6445 & 13.30 & 6.63 & 2.42 & 1.27 & 3231 & 5683 \\
EAD & 10.20 & 5.50 & 2.27 & 1.27 & 2807 & 5705 & 10.20 & 5.50& 2.27 &1.27 &2807& 5705  \\
EAG & 7.99 & 4.20 & 1.81 & 1.20 & 2155 & 4697 & 7.83 & 4.00 & 1.68 & 1.09 & 2086 & 4263 \\
EAGD & \textbf{6.39} & \textbf{3.58} & \textbf{1.71} & \textbf{1.19} &\textbf{1778} & \textbf{4275} & \textbf{6.42} & \textbf{3.45} & \textbf{1.55} & \textbf{1.04} & \textbf{1757} & \textbf{3845} \\
\specialrule{.1em}{.05em}{.05em} 

\multicolumn{1}{c}{} & \multicolumn{12}{c}{Impersonation attacks} \\
 & \multicolumn{6}{c}{LFW dataset \cite{LFWTech}} & \multicolumn{6}{c}{CPLFW dataset \cite{CPLFWTech}} \\
\specialrule{.1em}{.05em}{.05em} 
 & \multicolumn{4}{c}{\begin{tabular}[c]{@{}c@{}}Minimum perturbation\\ norm with query budget\end{tabular}} & \multicolumn{2}{c}{\begin{tabular}[c]{@{}c@{}}Avg. \# queries for\\  perturbation with norm\end{tabular}} & \multicolumn{4}{c}{\begin{tabular}[c]{@{}c@{}}Minimum perturbation\\ norm with query budget \end{tabular}} & \multicolumn{2}{c}{\begin{tabular}[c]{@{}c@{}}Avg. \# queries for\\ perturbation with norm \end{tabular}} \\

Attack method & 1K & 2K & 5K & 10K & 4 & 2 & 1K & 2K & 5K & 10K & 4 & 2 \\ \hline
Sign-OPT \cite{cheng2019sign}& 22.19 & 16.47 & 8.29 & 4.20 & 7574 & 9076 & 20.45 & 13.67 & 5.16 & 2.09 & 5163 & 7083 \\
HSJA \cite{chen2020hopskipjumpattack}& 18.72 & 13.53 & 6.40 & 3.58 & 6566 & 8538 & 15.48 & 9.47 & 3.47 & 1.71 & 3844 & 5963 \\
EA \cite{dong2019efficient}& 14.84 & 8.50 & 3.43 & 1.82 & 4363 & 7323 & 11.58 & 5.61 & 1.92 & 0.96 & 2652 & 4537 \\
EAG & \textbf{10.85} & \textbf{6.22} & \textbf{2.63} & \textbf{1.57} & \textbf{3219} & \textbf{6277} & \textbf{8.40} & \textbf{4.04} & \textbf{1.43} & \textbf{0.83} & \textbf{1925} & \textbf{3505} \\
\specialrule{.1em}{.05em}{.05em} 
\end{tabular}
\caption{Evaluation of decision-based advesrarial attacks on the two datasets.}
\label{table:main_results}
\end{table*}

\setlength{\tabcolsep}{6pt} % Default value: 6pt
\section{Experiments}
\subsection{Experimental settings}
We evaluated the performance improvement of GADA with Labeled Faces in the Wild (LFW) \cite{LFWTech} and Cross-Pose LFW (CPLFW) \cite{CPLFWTech} datasets. We used the datasets for the following reasons: (1) the LFW dataset is one of the most representative datasets for face recognition; (2) Since the CPLFW dataset has more diverse facial pose changes than the LFW dataset, it is appropriate to show that the GADA strategy works well for general cases. We found the best threshold with the highest accuracy using 10-fold splits for each dataset and made the model classify images based on that threshold.

For creating test image sequences for dodging attacks, we randomly extracted 500 pairs from each dataset, each pair of which represents an identity. When composing test image sequences for impersonation attacks, we permuted each pair's left image (i.e., $\bm{x}_A$) so that all pairs are perceived as different identities and used them for target images for impersonation attacks. Meanwhile, we used the original images as the source images for the target identities. The maximum number of queries available in each image is set to 10K for both types of attacks. For measuring the query efficiency of attacks, we used two types of metrics: (1) the smallest norm of adversarial perturbations found within a specific query budget; (2) the average number of queries used to generate an adversarial example whose norm is less than or equal to a threshold. This metric is useful for comparing the average number of queries spent to find a sufficiently small adversarial perturbation.

In our experiments, we used the ArcFace ResNet-50 \cite{deng2019arcface,he2016deep} model\footnote{We use the pre-trained ArcFace model provided from \url{https://github.com/ZhaoJ9014/face.evoLVe.PyTorch}} trained on the $112{\times}112$ aligned MS-Celeb-1M dataset \cite{guo2016ms} as the target model for the black-box attacks. For GADA's dictionary attacks, a feature embedding for $\bm{x}_A$ needs to be computed. Since we assumed a realistic black-box setting, we used a network which is different 
from the target model. Specifically, we used FaceNet\footnote{We use the pre-trained FaceNet model provided from \url{https://github.com/timesler/facenet-pytorch}} \cite{schroff2015facenet} trained on the VGGFace2 dataset \cite{cao2018vggface2}.
For 3D face alignment in GADA, we used a pre-trained model\footnote{We use the default model whose backbone is MobileNet\_V1 \cite{howard2017mobilenets} from \url{https://github.com/cleardusk/3DDFA_V2}} of 3DDFA\_V2 provided by the authors.

The proposed method can be applied to various existing query-based black-box attacks for improving their query efficiency. In this paper, we applied GADA to two different decision-based attacks, EA \cite{dong2019efficient} and SFA \cite{chen2020boosting}, and show their performance improvement. 

\textbf{Notations.} We used diverse variants of attack methods depending on their additional functions in our experiments. To refer to them, we attach `G' to the name of an attack when using geometrically adaptive attacks and add `D' when using dictionary attacks. For example, EAGD is the improved version of the EA attack \cite{dong2019efficient} with the proposed geometrically adaptive attacks and dictionary attacks. 

\subsection{Quantitative results}
We evaluated the query efficiency of the variants of EA. For comprehensive comparisons, we also evaluated other state-of-the-art decision-based attack methods, HSJA \cite{chen2020hopskipjumpattack} and Sign-OPT \cite{cheng2019sign}. 
GADA can also be applied to the above two attacks, but we applied GADA to EA because EA shows the best query efficiency in our experiments. We described hyperparameter settings for each attack method in supplementary material. In EAD and EAGD, to prevent the evolutionary algorithm from being stuck at the boundary of the pixel value range, we clip the image of perturbation-applied areas into the range between 0.2 and 0.8 when they utilize a previous perturbation.

Table \ref{table:main_results} shows the evaluation results of decision-based adversarial attacks on the two datasets. Note that for all methods, we measured the norm of perturbations in the image space, not in the UV space. For dodging attacks on the LFW dataset, EAGD reduces the smallest perturbation norm by 1.03 compared to EA. The results of EAGD clearly show that using a dictionary for utilizing previous perturbation helps to improve query efficiency. Besides, compared to EA, EAGD reduces the number of queries required to find an adversarial perturbation whose norm is less than or equal to two by 2,170 for dodging attacks and 1,046 for impersonation attacks. This query efficiency can save considerable resources of attackers. We illustrated the perturbation norm curves that visually shows the query efficiency of each method in supplementary material.

\begin{figure}[t]
     \centering
     \begin{subfigure}[b]{0.48\textwidth}
         \centering
         Dodging attacks
         \includegraphics[width=\textwidth,trim={0cm 0cm 0cm 0cm},clip]{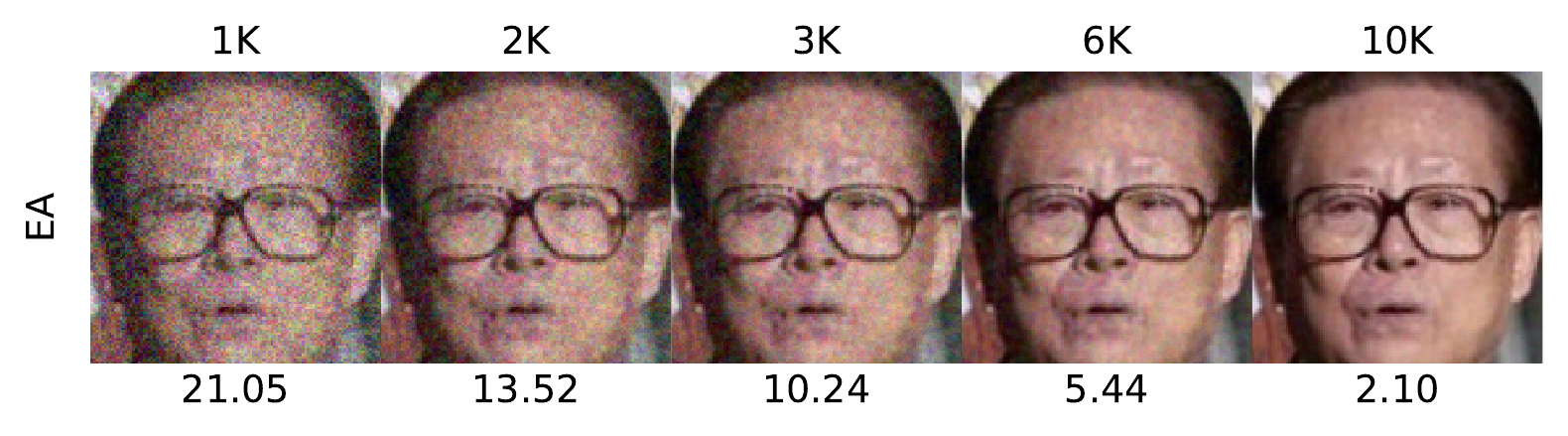}
         \includegraphics[width=\textwidth,trim={0cm 0cm 0cm 0.3cm},clip]{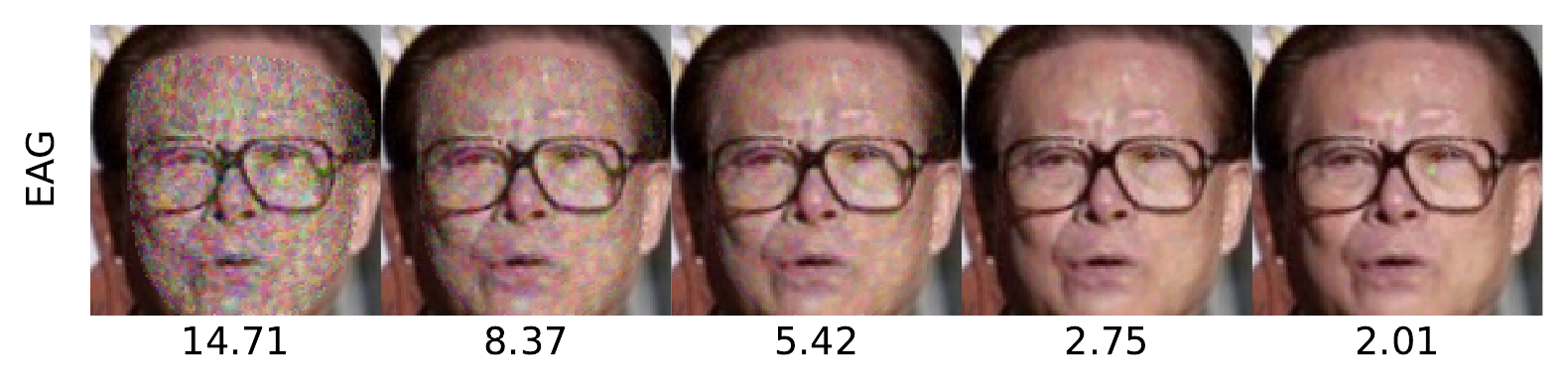}
         \includegraphics[width=\textwidth,trim={0cm 0cm 0cm 0.3cm},clip]{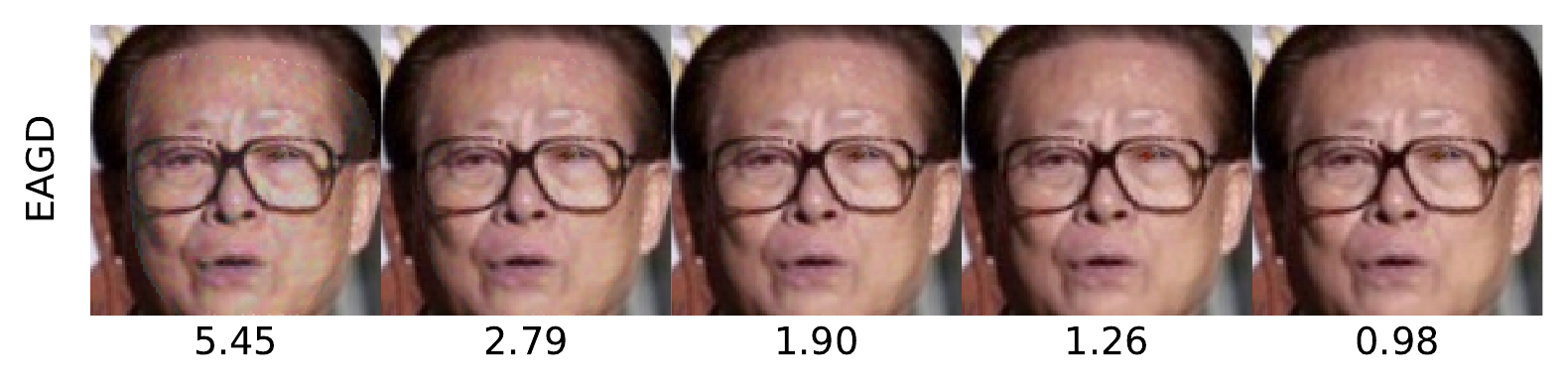}
     \end{subfigure}
     Impersonation attacks
          \begin{subfigure}[b]{0.48\textwidth}
         \centering
         \includegraphics[width=\textwidth,trim={0cm 0cm 0cm 0cm},clip]{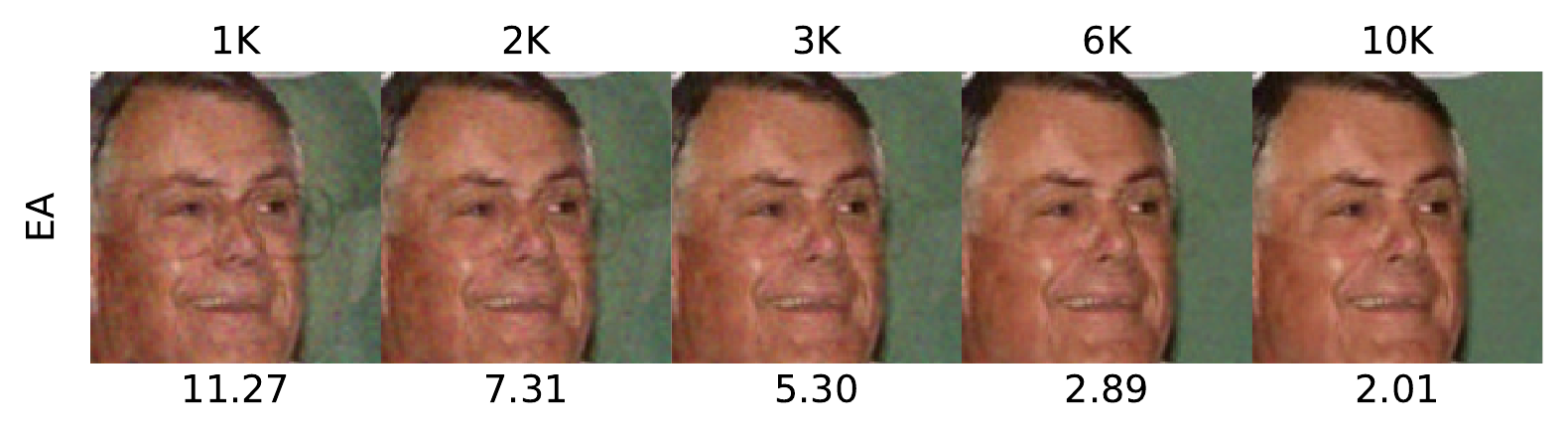}
         \includegraphics[width=\textwidth,trim={0cm 0cm 0cm 0.3cm},clip]{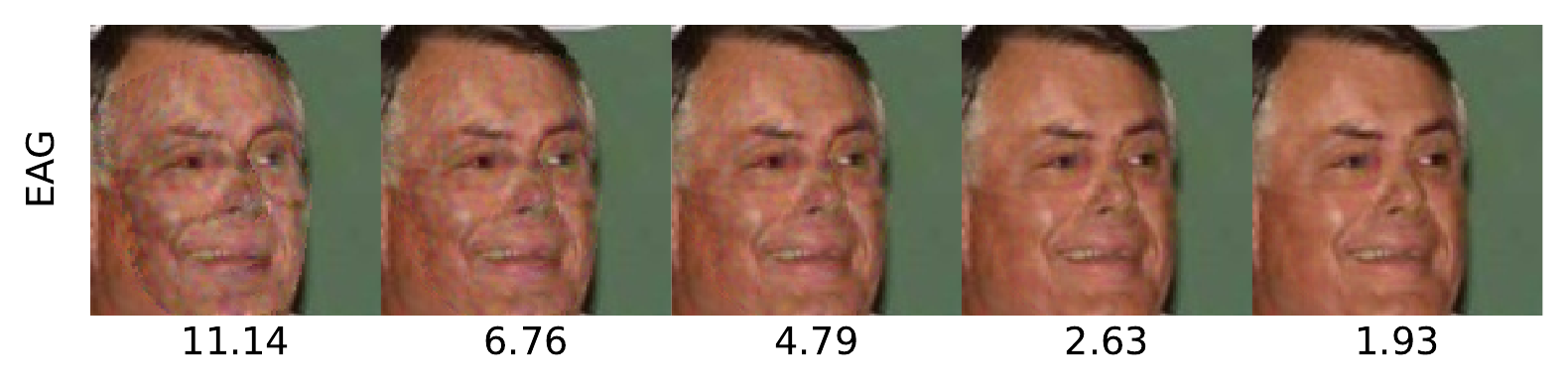}
     \end{subfigure}
        %\vspace{-0.2cm}
        \caption{Qualitative results of dodging and impersonation attacks on the LFW dataset \cite{LFWTech}. For each attack, we illustrate the minimum norm-adversarial examples in each query budget. The $\ell_2$ norm of perturbation is displayed under each image.}
        %\vspace{-0.2cm}
        \label{fig:fig3}
\end{figure}

\subsection{Qualitative results}
Figure \ref{fig:fig3} shows adversarial examples from EA and its variants. EAG generates perturbations only in the facial area, so unlike EA, background noise does not exist.
Since GADA reduces the search space, it can be seen that when $Q$=3K, the perturbation is significantly reduced compared to EA. Dictionary attacks of EAGD help to start at a smaller perturbation and the results show that when $Q$=1K, EAGD reduces the perturbation norm to $0.37\times$ compared to EAG. 
%Tnd the target model is fooled with much less perturbation than other methods.

For impersonation attacks, existing methods start from a source image of the target identity and iteratively update the adversarial example to look like the target image. However, as shown in Fig. \ref{fig:fig3}, if the target image has a flat background, the adversarial perturbation becomes more noticeable. In contrast, as GADA initially replace the target image's face with the source image's face and optimizes the perturbation in the facial area, the noise becomes far more imperceptible when the query budget is only 2K. For more qualitative comparisons with other attacks, we include more exemplary results in supplementary material. We also conducted the above experiments with a deeper target model, Curricular Face ResNet-100 \cite{huang2020curricularface, he2016deep}, and the experimental results are listed in supplementary material.

\begin{figure}[t]
     \centering
         \includegraphics[width=0.45\textwidth,trim={0cm 0.1cm 0cm 0cm},clip]{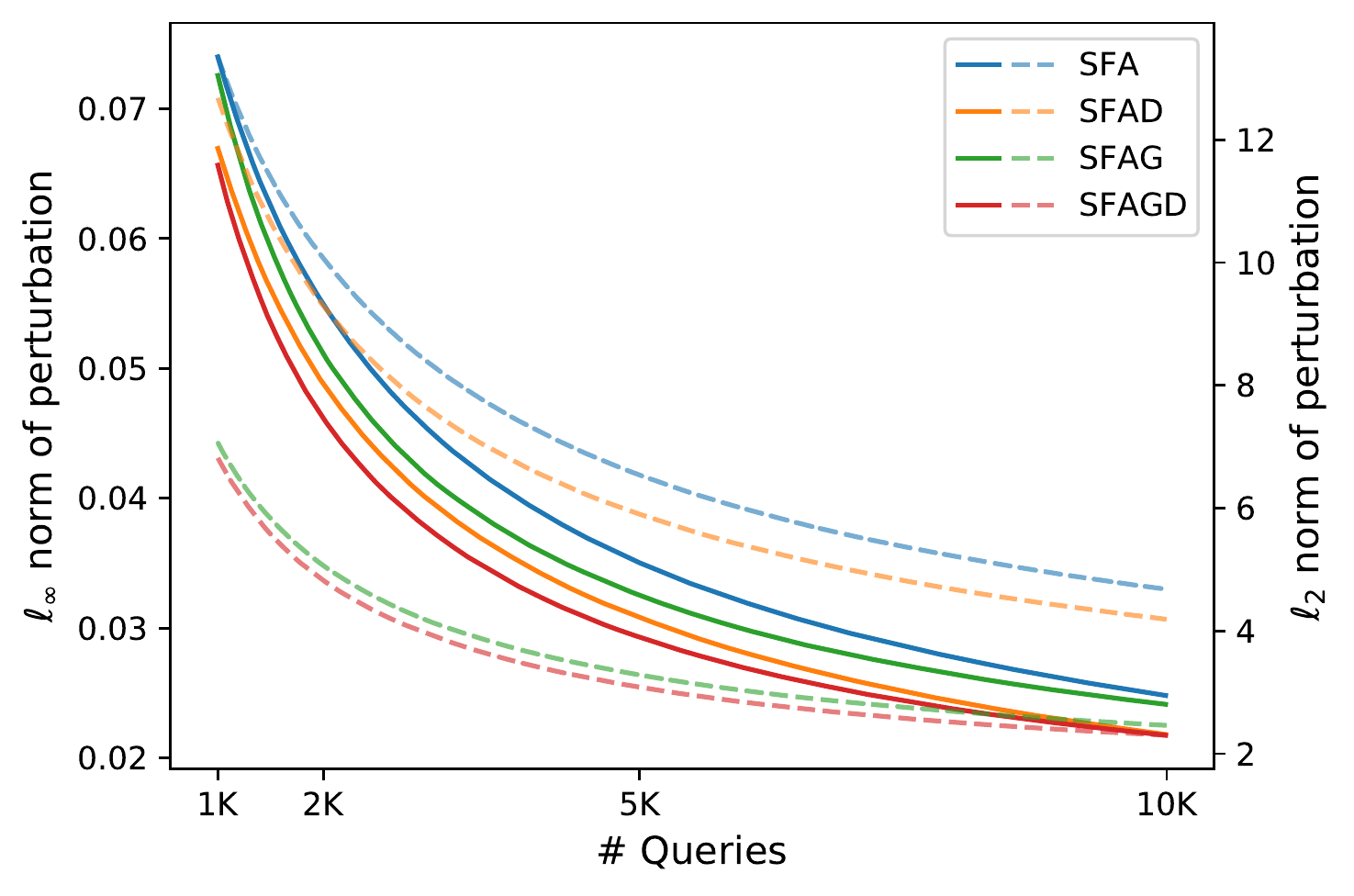}
        \caption{Perturbation norm curves of SFA and its variants for dodging attacks on the LFW dataset. The solid lines represent $\ell_\infty$ norm of perturbations, and the dotted lines represent $l_2$ norm of perturbations.
        }
        %\vspace{-0.2cm}
        \label{fig:sfa}
\end{figure}

\subsection{Adaptation to $\ell_\infty$ norm-based attacks}
Since GADA is a general strategy applicable to various query-based black-box attacks, we applied it to SFA \cite{chen2020boosting} that is particularly effective for $\ell_\infty$ norm constraint. Figure \ref{fig:sfa} shows the results of SFA and its variants for dodging attacks on the LFW dataset. SFAGD clearly reduces the perturbation norm faster than SFA. When $Q$=10K, the difference of $\ell_\infty$ norm between SFAD and SFAGD is small, but SFAGD almost halves $\ell_2$ norm of the perturbation since SFAGD perturbs the facial area only.

\begin{figure}[t]
     \centering
         \includegraphics[width=0.4\textwidth,trim={0cm 0.1cm 0cm 0cm},clip]{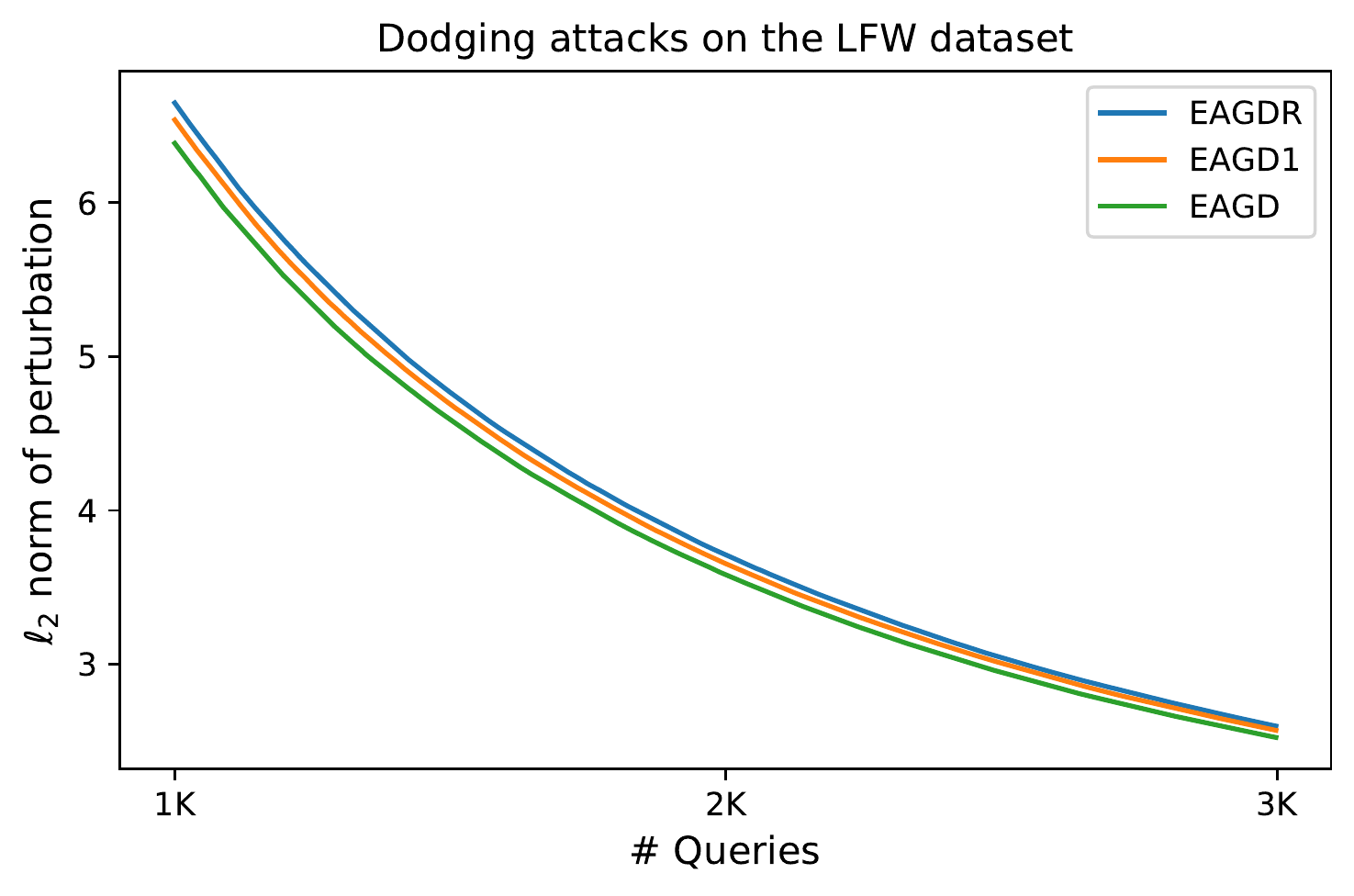}
        \caption{Perturbation norm curves for different ways of fetching perturbations.}
        \label{fig:fetch}
        %\vspace{-0.2cm}
\end{figure}
\subsection{Different ways to fetch a perturbation}
Originally, GADA stores the adversarial perturbations of many identities in a dictionary to fetch the perturbation of the closest identity in the feature space for better initialization in the future. In this ablation study, we find out the effectiveness of this way of fetching. We evaluated two different ways of fetching: EAGD1 and EAGDR. EAGD1 has a dictionary with only one memory slot so that it fetches and updates the perturbation for each attack. EAGDR stores adversarial perturbations of many identities like EAGD, but it fetches a  perturbation in the dictionary randomly unless the same face feature exists in the dictionary. Figure \ref{fig:fetch} shows the results of EAGD along with the above variants. The results show that EAGD indeed has superior query efficiency than the other variants in the early stages. One may take the closest top-k perturbations and multiply them by a large number and gradually scale them down to find the smallest adversarial perturbation. Devising an efficient way to find a more useful perturbation in the dictionary can be an interesting research topic in the future.

\subsection{Evading stateful detection}
Query-based black-box attacks inevitably need to send a large number of perceptually similar images for queries in their processes. By targeting this commonality, a memory-based detection technique \cite{chen2019stateful} has been proposed recently. It stores perceptual similarity embeddings of recent queries and detects the generation of adversarial examples if too many similarity embeddings are close. In detail, it encodes each image into a feature with a similarity encoder and stores it in its memory. If the $k$-nearest neighbor ($k$-NN) distance of the current image's embedding is smaller than a threshold, the detector judges the query as an adversarial attack. Existing decision-based black-box attacks are nearly impossible to avoid this detection technique due to their nature.
\begin{figure}[t]
     \centering
         \includegraphics[width=0.45\textwidth,trim={0cm 0.1cm 0cm 0cm},clip]{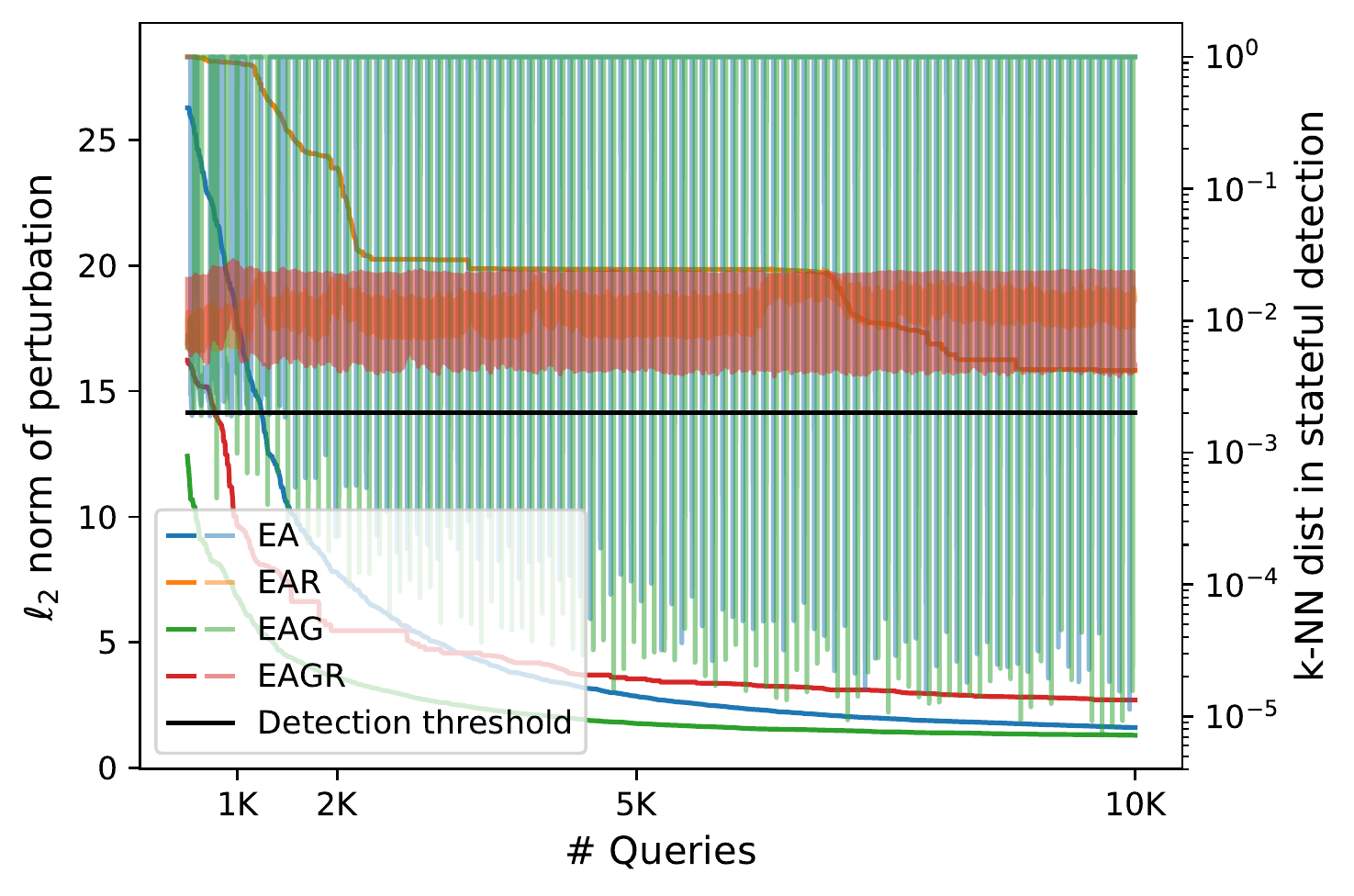}
        \caption{Perturbation norm and $k$-NN distance curves of EA and its variants. If the $k$-NN distance falls below the threshold, the query is detected as an adversarial attack.}
        \label{fig:sd_graph}
        %\vspace{-0.2cm}
\end{figure}

Despite the difficulties, we present a novel attack strategy that avoids this detection by injecting noise differently into the foreground and the background. Specifically, we disturb the similarity detector by consistently adding small random noise to the background while optimizing the foreground perturbation. An essential premise for this is that background noise should not affect the prediction of the target model. Since most face recognition models focus on inner part of faces, we assume that this premise is established for general cases.

To evaluate this approach in our experimental setting, we need a new similarity encoder for larger-sized images as Chen \textit{et al.} \cite{chen2019stateful} use perceptual similarity encoders for the 32x32 sized images of the CIFAR-10 dataset. Instead, we used Learned Perceptual Image Patch Similarity (LPIPS) \cite{zhang2018perceptual} as a similarity encoder, which can measure general perceptual similarity. We used a pre-trained SqueezeNet model for LPIPS v0.1 and used $k$=50 for $k$-NN distance and 2e-3 for the threshold of detection. For its memory, we used a circular buffer containing 100 recent queries. The detection operates only when more than $k{-}1$ queries are stored in the buffer. Following \cite{chen2019stateful}, we flush the buffer whenever an adversarial attack is detected. We evaluated this detection against dodging attacks on the first 100 images of the test sequence of the LFW dataset. For SO, HSJA, EA, and EAG, the average number of detection per image is 166, 191, 188, 192, respectively. Considering that the total query budget of 10K divided by $k$ (50) is 200, the above results show that most attacks are immediately detected when the detector starts to operate. 

Before we describe the results of the new attack strategy against the detection, we explain our attack strategy in detail. We add random Gaussian noise into the background with $\sigma=0.01$ divided by the background area ratio in the image. This makes larger noise when the background area is relatively small. We also applied random Gaussian noise with $\sigma=0.02$ to the entire image for EA to avoid detection. We name the two variants EAGR and EAR, respectively. Note that the ratio of the background is less than 30\% on average, so we used a smaller $\sigma$ for EAR in most cases. For both methods, we make a query without background noise every $i$ iterations to reduces the perturbation norm by excluding the background noise. We used $i=20$ in our experiments, but $i$ should be set in proportional to the detector's buffer size. With the same experimental settings as the other attacks, both methods are never detected for the 100 test images. However, when the query budget is 10K, EAR and EAGR have 11.69 and 3.47, respectively, in their smallest perturbation norm on average. It means that unlike EAR, EAGR can successfully reduce the perturbation norm while avoiding detection. Figure \ref{fig:sd_graph} shows the $k$-NN distance and perturbation norm curves for a test image.

\section{Related work}
In the following, we briefly introduce related studies and their difference to our strategy. Dabouei \textit{et al.} utilize spatial transformation of images with face landmarks to fool face recognition models. It creates an adversarial example by displacing the position of the facial landmarks in the white-box threat model. It is common with our strategy in that it creates perturbation based on facial geometry, but GADA significantly differs as GADA is an intensity-based query-efficient black-box attack strategy that finds perturbations in the UV map. On the other hand, there are several 3D model-based adversarial attacks \cite{athalye2018synthesizing,xiao2019meshadv,zeng2019adversarial} which render 3D models in 2D space and find adversarial shapes or textures. However, their aim is not to query-efficient black-box attacks but to robust attacks under diverse views.
%------------------------------------------------------------------------
\section{Conclusion}
In this paper, we propose a general strategy for query-efficient black-box attacks on face recognition. It creates an adversarial perturbation in a common UV texture map and projects it onto the face area through 3D face alignment. By separating the facial areas and the background, we also suggest that injecting noise into the background that hardly affects predictions can help to circumvent stateful detection. In this paper, we opened a new research avenue for memory-based black-box attacks that can efficiently utilize previously found perturbations. We will release GADA's code and the test image sequences' indices for fruitful exploration with other researchers. The generalized core ideas of our work are limiting the perturbation search space to the region of interest and recycling previously found perturbations with object-aware semantic correspondence. In this paper, we deal with face recognition, but the above ideas can be effective for attacks on other tasks. We leave them for future work.
{\small
\bibliographystyle{ieee_fullname}
\bibliography{egbib}
}

\end{document}

%%%%%%%%% TITLE
\title{Supplementary Material: Geometrically Adaptive Dictionary Attack on Face Recognition}

\maketitle

\ifwacvfinal
\thispagestyle{empty}
\fi

%%%%%%%%% ABSTRACT
\newcommand{\norm}[1]{\left\lVert#1\right\rVert}
%%%%%%%%% ABSTRACT

%%%%%%%%% BODY TEXT
\section{Overview}
In this supplementary material, we describe hyperparameter settings for attack methods used in our experiments. We also illustrate the perturbation norm curves that visually show each method's query efficiency and additional qualitative results for comprehensive comparisons. We also list the experimental results of the decision-based black-box attacks against a deeper target model, CurricularFace ResNet-100 \cite{huang2020curricularface, he2016deep}. 

\section{Additional details of experimental setting}
We use Pytorch framework \cite{NEURIPS2019_9015} for our experiments and borrow the code for face recognition from face.evoLVe library\footnote{\url{https://github.com/ZhaoJ9014/face.evoLVe.PyTorch}}. We use the $112\times112$ aligned datasets provided by face.evoLVe library for the LFW \cite{LFWTech} and CPLFW \cite{CPLFWTech} datasets.

\section{Implementation and hyperparameter settings of the attacks}
\textbf{Sign-OPT (SO) \cite{cheng2019sign}.} We adopt the code\footnote{\url{https://github.com/cmhcbb/attackbox}} of Sign-OPT provided by the authors without special tuning the hyperparameters of the attack.

\textbf{HSJA \cite{chen2020hopskipjumpattack}.} We implement HSJA by using Adversarial Robustness Toolbox (ART) library \cite{art2018}.
From its default setting, we increase the maximum iterations to 64 to follow the authors' experimental settings\footnote{\url{https://github.com/Jianbo-Lab/HSJA}}.

\textbf{EA \cite{dong2019efficient}.}
We implement EA based on the code\footnote{\url{https://github.com/thu-ml/realsafe/blob/master/realsafe/attack/evolutionary_worker.py}} of EA provided by the authors. We set the dimension of the perturbation search space as $60\times60\times3$ and the coefficient of the distance in calculation of $\sigma$ as $0.03$ instead of $0.01$ for faster convergence. These settings also apply to the EA's variants (EAD, EAG, EAGD).

\textbf{SFA \cite{chen2020boosting}.}
We adopt the code\footnote{\url{https://github.com/wubaoyuan/Sign-Flip-Attack}} of SFA provided by the authors without special tuning of the hyperparameters. We set the dimension reduction ratio as 2 for reducing the perturbation search space. These settings also apply to the SFA's variants (SFAD, SFAG, SFAGD).

\section{More experimental results on the ArcFace ResNet-50 model}
\begin{figure}[t]
    \centering
     \begin{subfigure}[b]{0.47\textwidth}
         \centering
         \includegraphics[width=\textwidth]{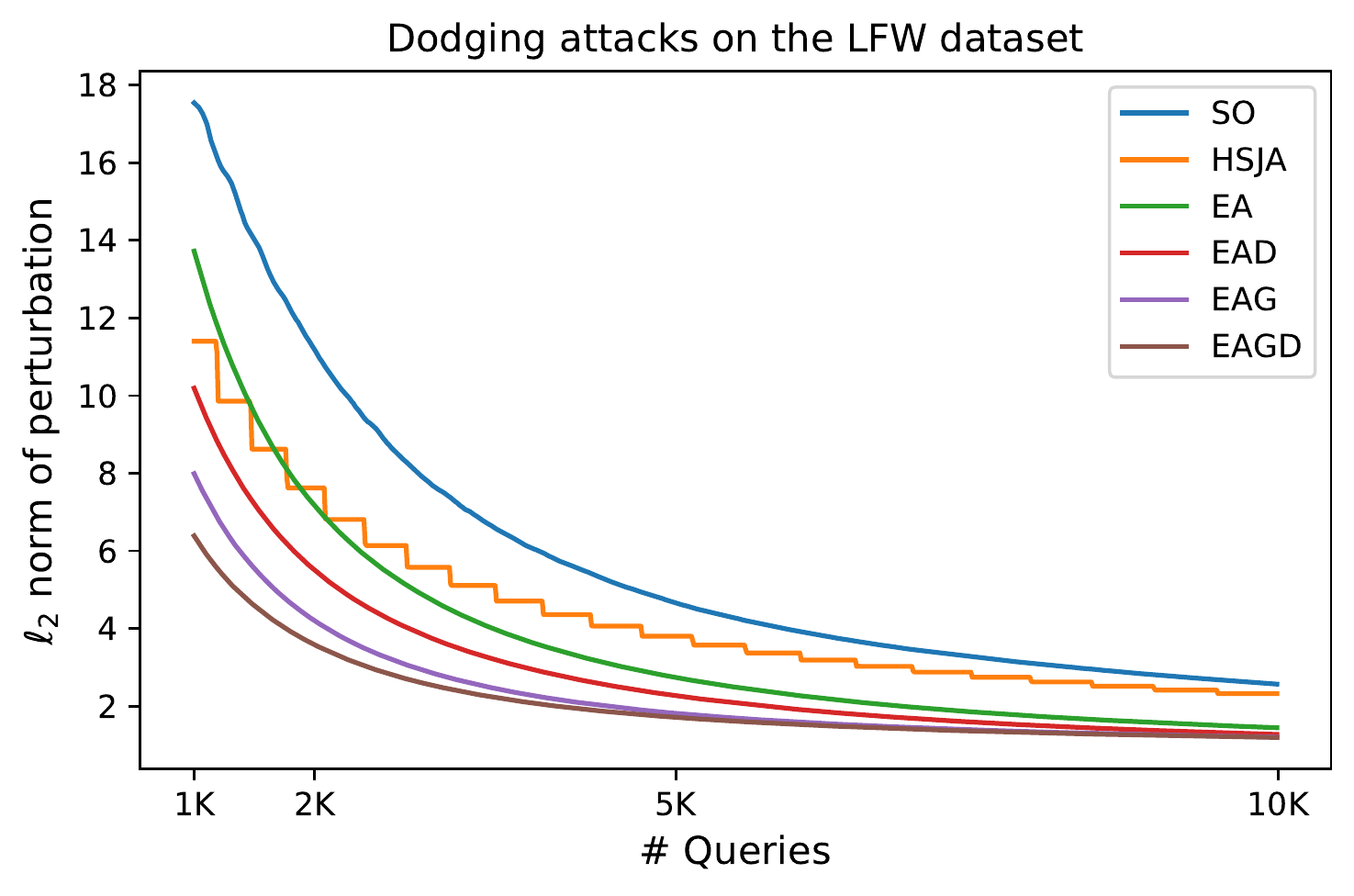}
     \end{subfigure}
          \begin{subfigure}[b]{0.47\textwidth}
         \centering
         \includegraphics[width=\textwidth]{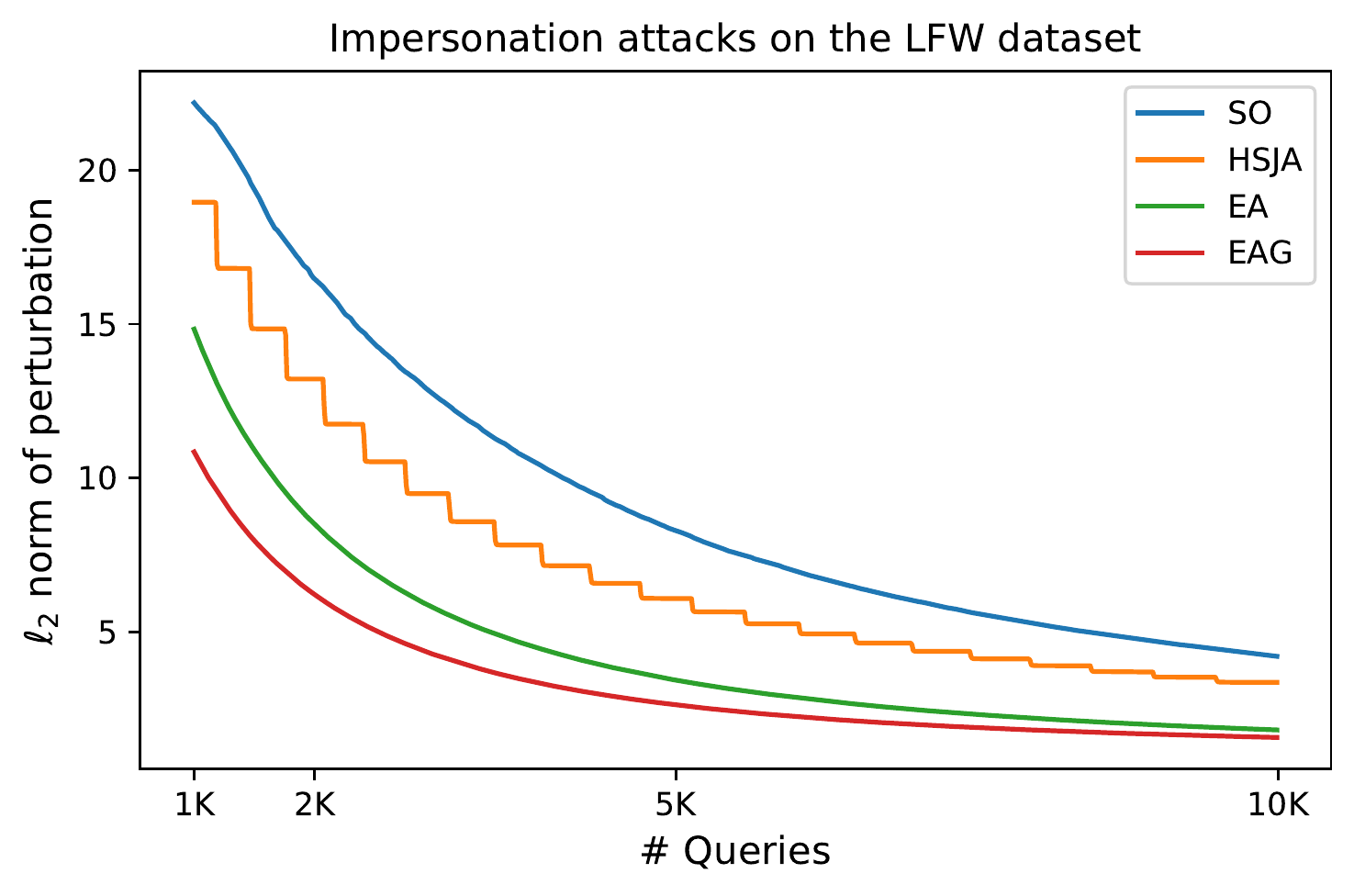}
     \end{subfigure}
        \caption{Perturbation norm curves of the decision-based attacks against ArcFace ResNet-50 model with the LFW dataset \cite{LFWTech}. 
        }%The results for EAF and EAFD are omitted to avoid complicating the comparison. 
   \label{fig:ir50_lfw}
\end{figure}

\begin{figure}[t]
    \centering
     \begin{subfigure}[b]{0.47\textwidth}
         \centering
         \includegraphics[width=\textwidth]{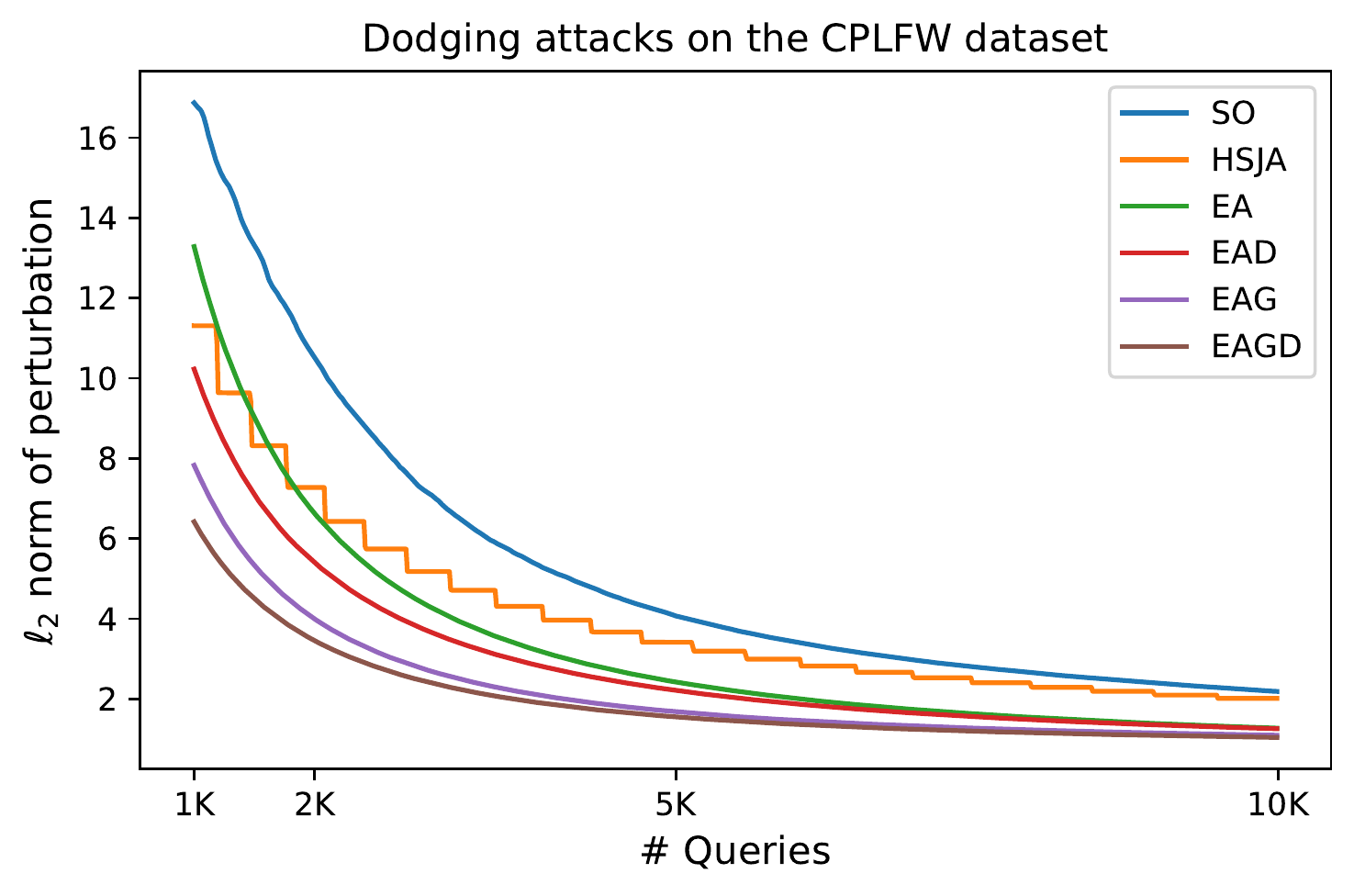}
     \end{subfigure}
         \begin{subfigure}[b]{0.47\textwidth}
         \centering
         \includegraphics[width=\textwidth]{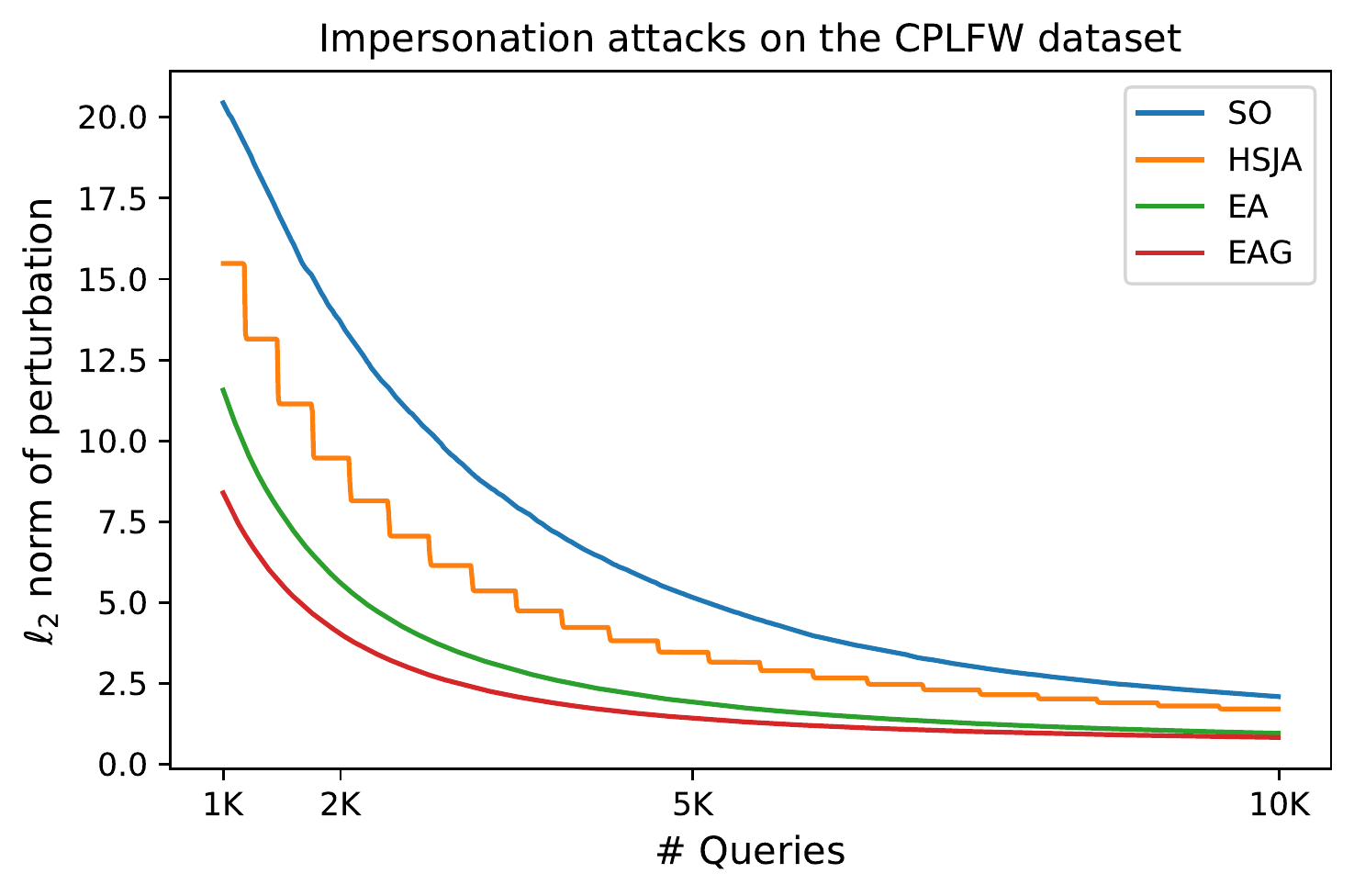}
     \end{subfigure}
        \caption{Perturbation norm curves of the decision-based attacks against ArcFace ResNet-50 model with the CPLFW dataset \cite{CPLFWTech}.
        }% The results for EAF and EAFD are omitted to avoid complicating the comparison. 
    \label{fig:ir50_cplfw}
\end{figure}

\begin{figure}[t]
     \centering
     \begin{subfigure}[b]{0.47\textwidth}
         \centering
         \includegraphics[width=\textwidth,trim={0cm 0.3cm 0cm 0cm},clip]{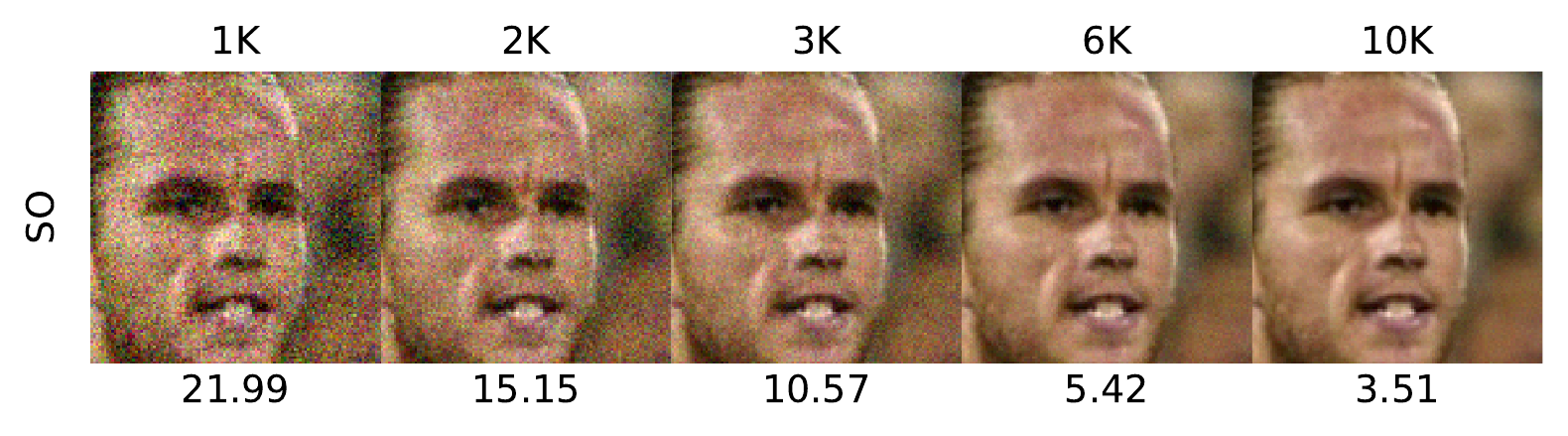}
         \includegraphics[width=\textwidth,trim={0cm 0.3cm 0cm 0cm},clip]{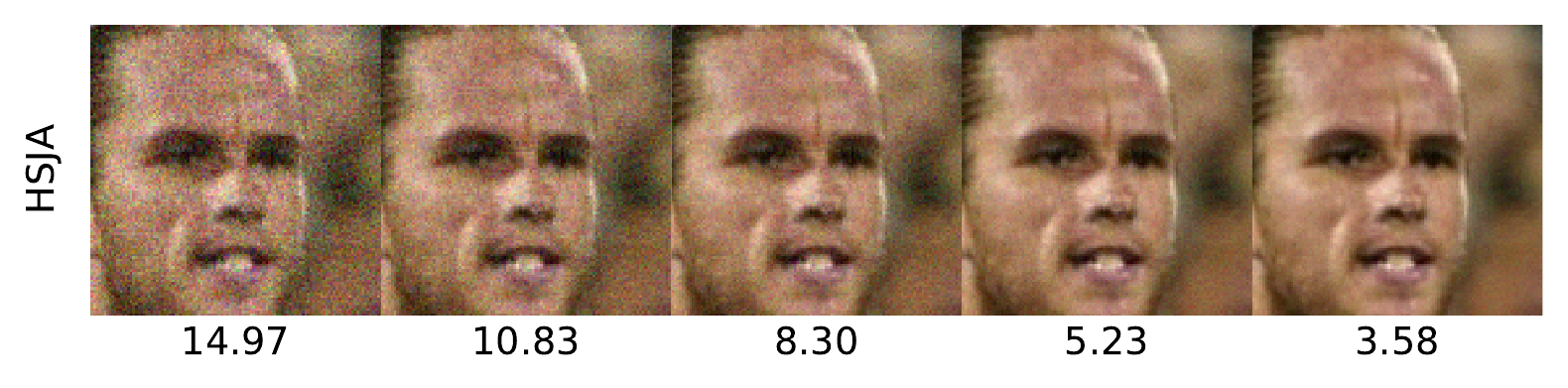}         \includegraphics[width=\textwidth,trim={0cm 0.3cm 0cm 0cm},clip]{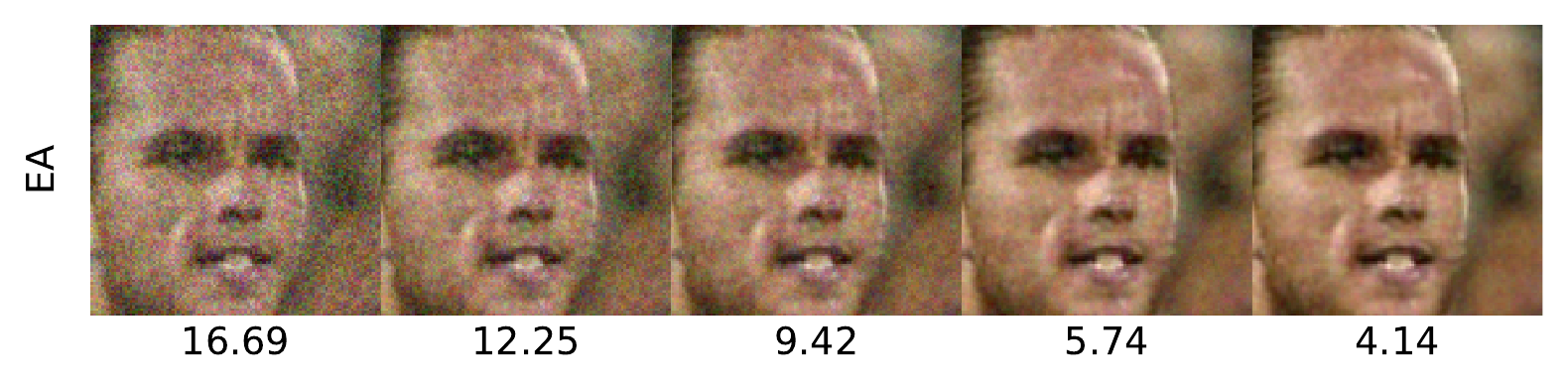}
         \includegraphics[width=\textwidth,trim={0cm 0.3cm 0cm 0cm},clip]{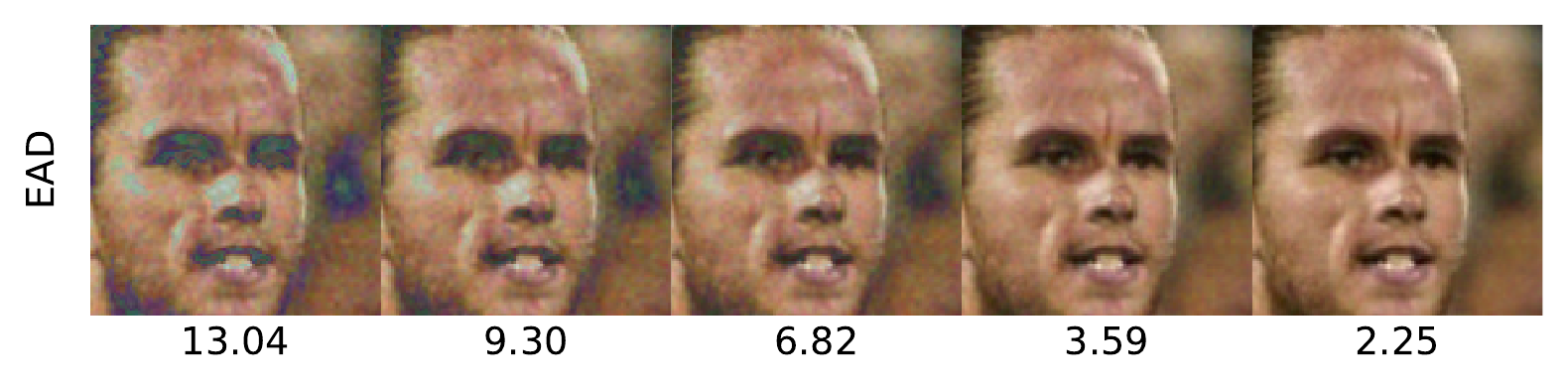}
          \includegraphics[width=\textwidth,trim={0cm 0.3cm 0cm 0cm},clip]{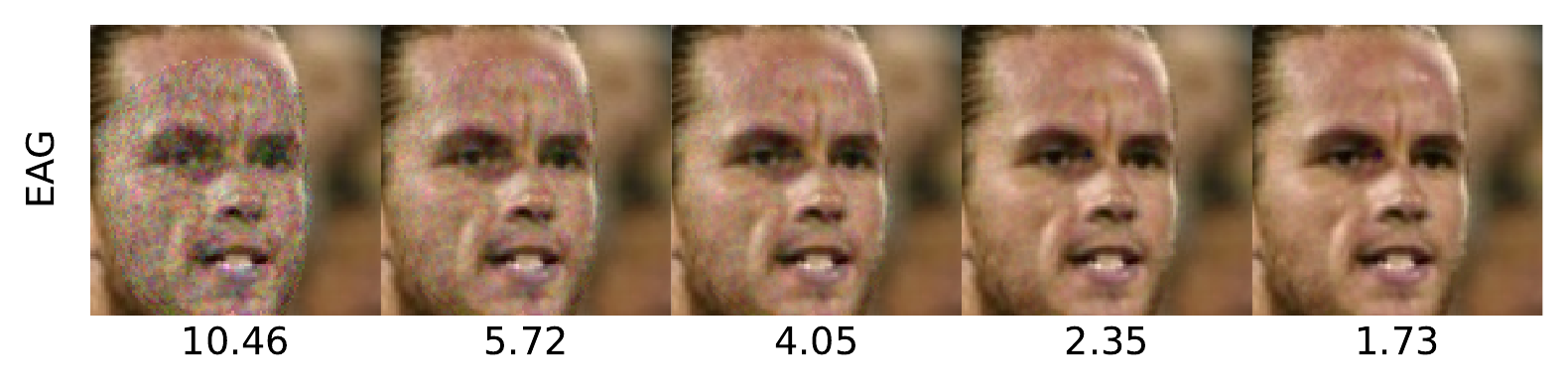}          \includegraphics[width=\textwidth,trim={0cm 0.3cm 0cm 0cm},clip]{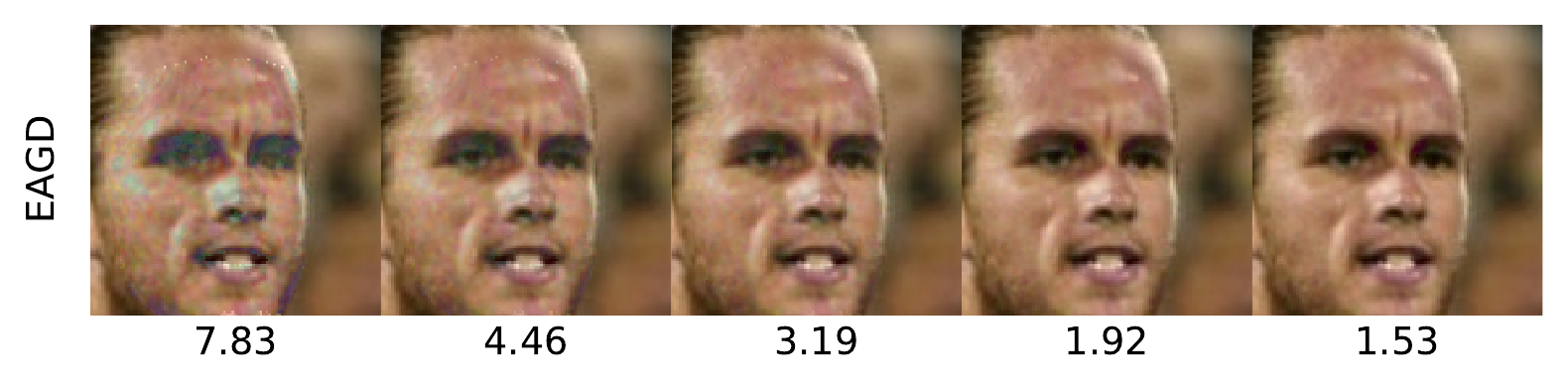} \includegraphics[width=\textwidth,trim={0cm 0.3cm 0cm 0cm},clip]{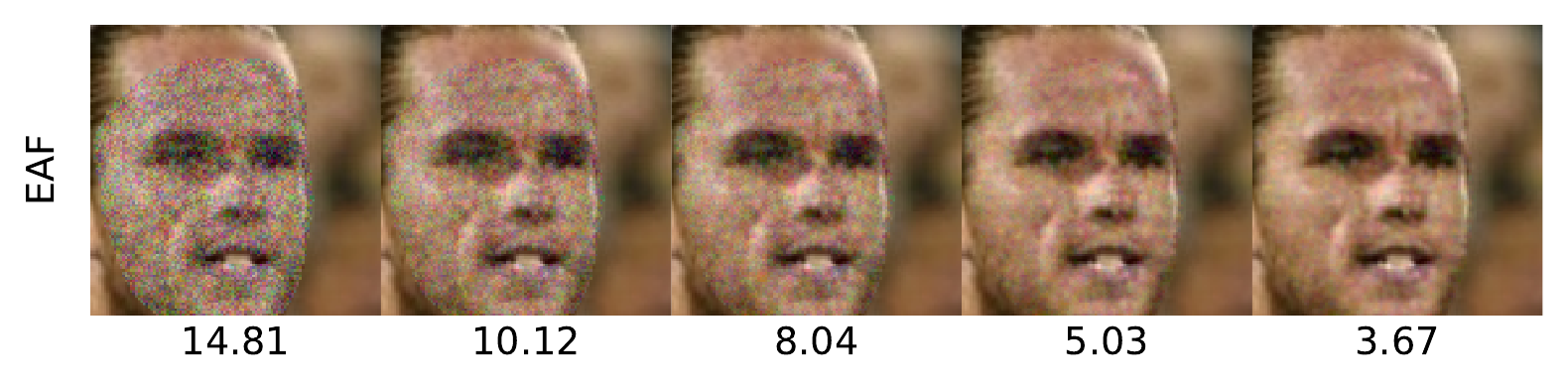}          \includegraphics[width=\textwidth,trim={0cm 0.3cm 0cm 0cm},clip]{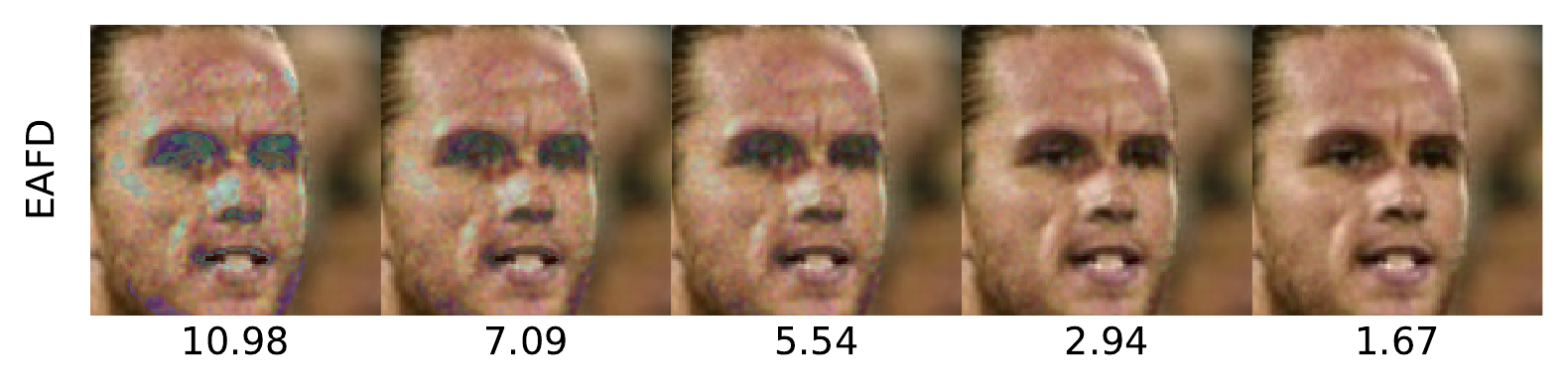}
     \end{subfigure}
 
        %\vspace{-0.2cm}
        \caption{Qualitative results of dodging attacks on the LFW dataset \cite{LFWTech}. For each attack, we illustrate the minimum norm-adversarial examples in each query budget. The $\ell_2$ norm of perturbation is displayed under each image.}
        \label{fig:dodging_lfw1}
\end{figure}

\begin{figure}[t]
     \centering
     \begin{subfigure}[b]{0.47\textwidth}
         \centering
         \includegraphics[width=\textwidth,trim={0cm 0.3cm 0cm 0cm},clip]{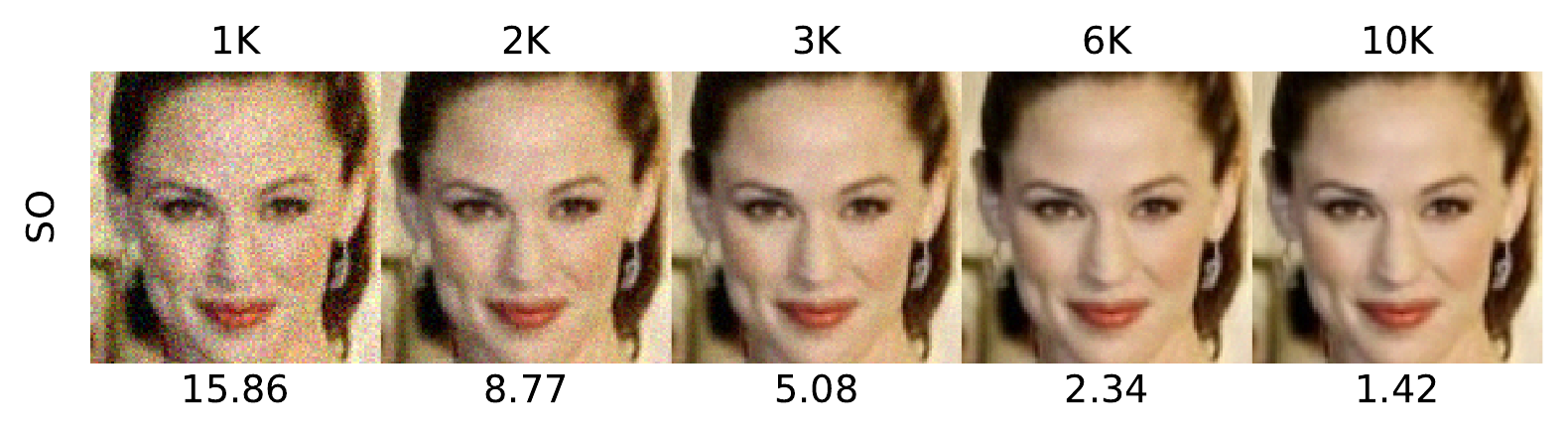}
         \includegraphics[width=\textwidth,trim={0cm 0.3cm 0cm 0cm},clip]{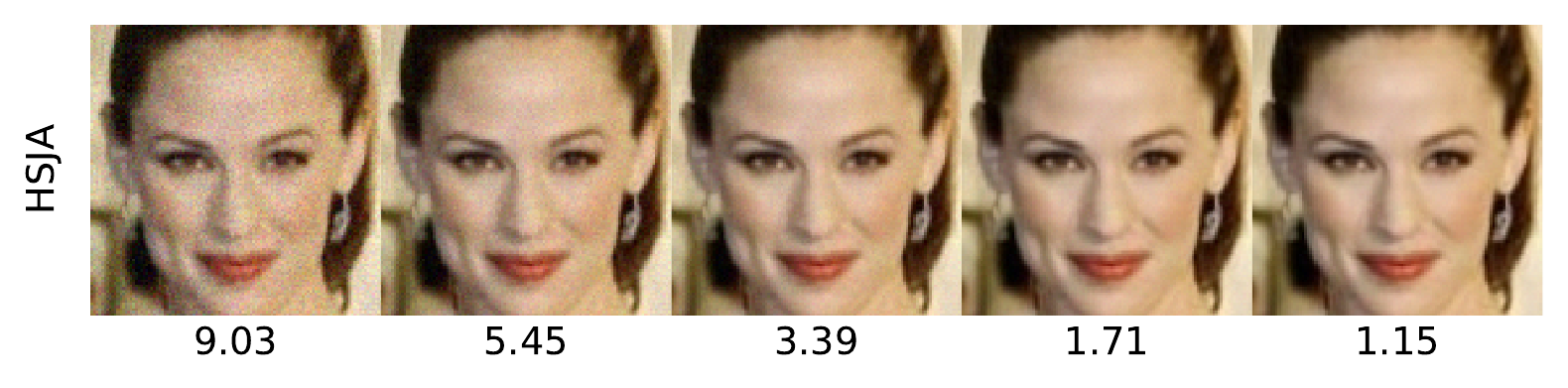}         \includegraphics[width=\textwidth,trim={0cm 0.3cm 0cm 0cm},clip]{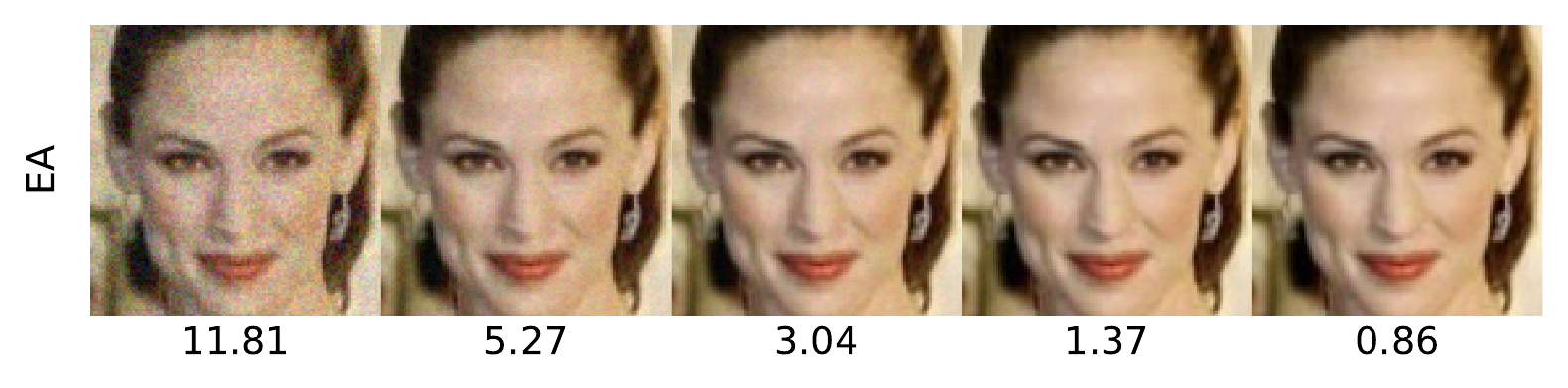}
         \includegraphics[width=\textwidth,trim={0cm 0.3cm 0cm 0cm},clip]{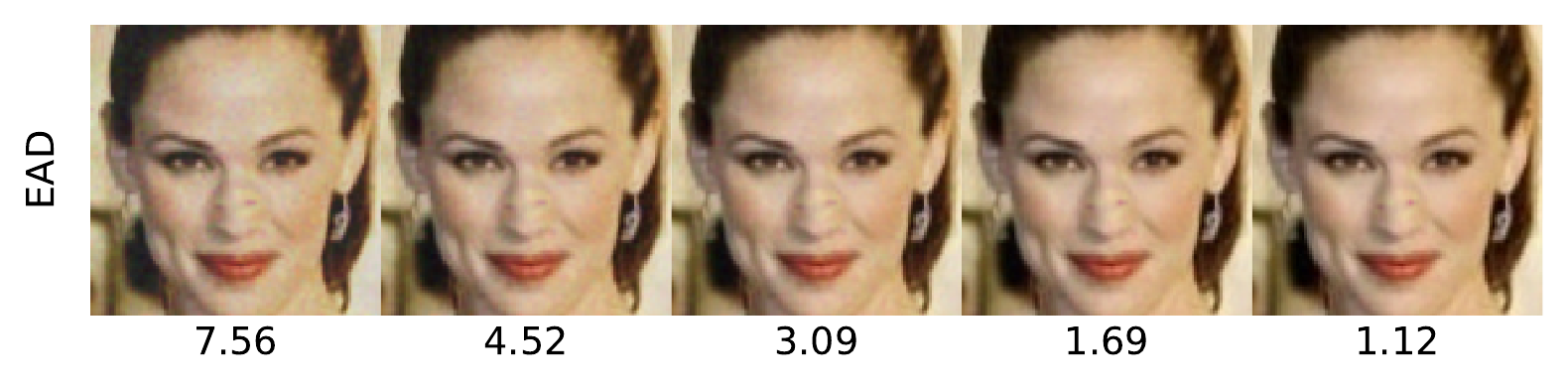}
          \includegraphics[width=\textwidth,trim={0cm 0.3cm 0cm 0cm},clip]{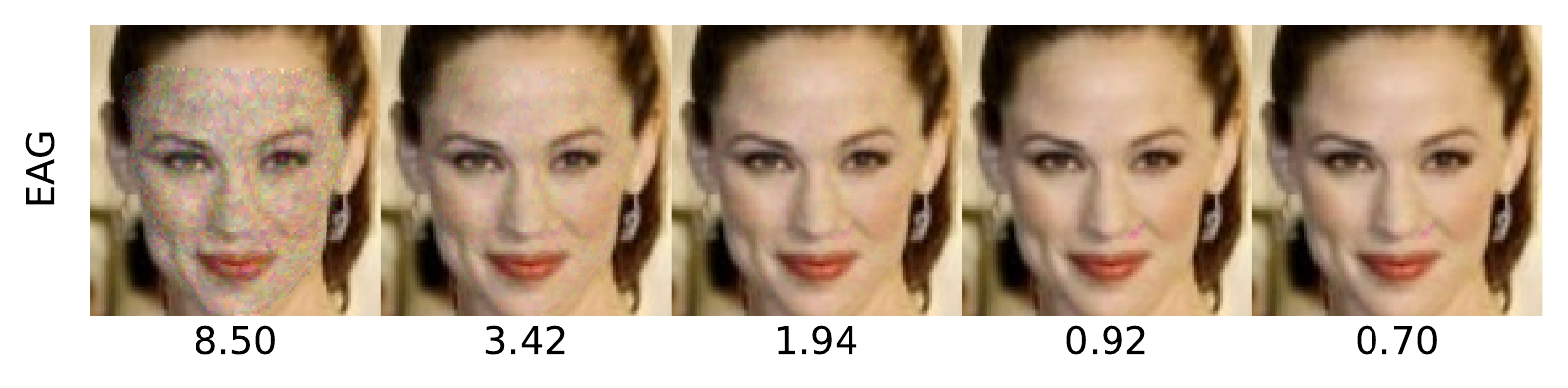}          \includegraphics[width=\textwidth,trim={0cm 0.3cm 0cm 0cm},clip]{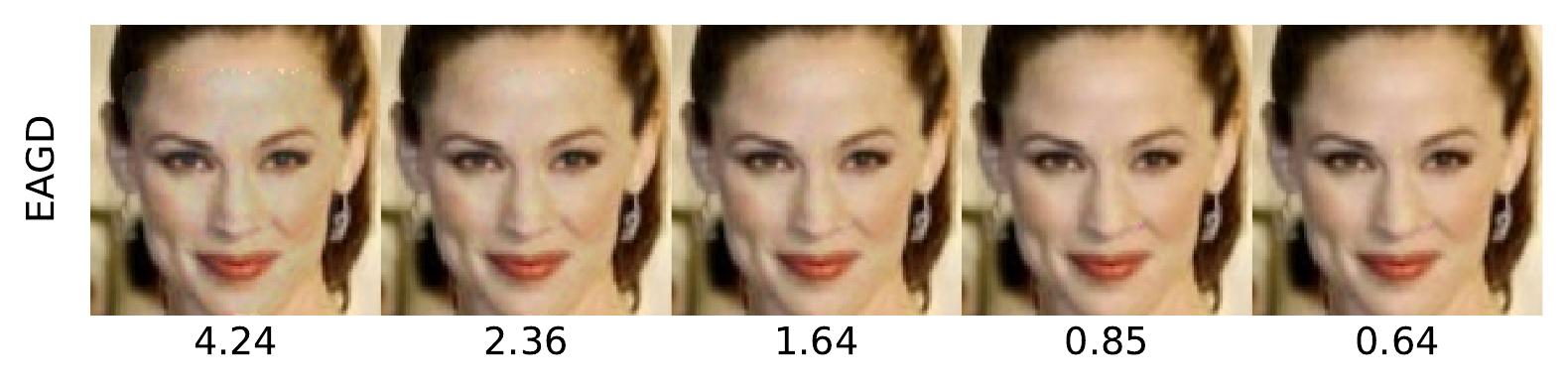}
          \includegraphics[width=\textwidth,trim={0cm 0.3cm 0cm 0cm},clip]{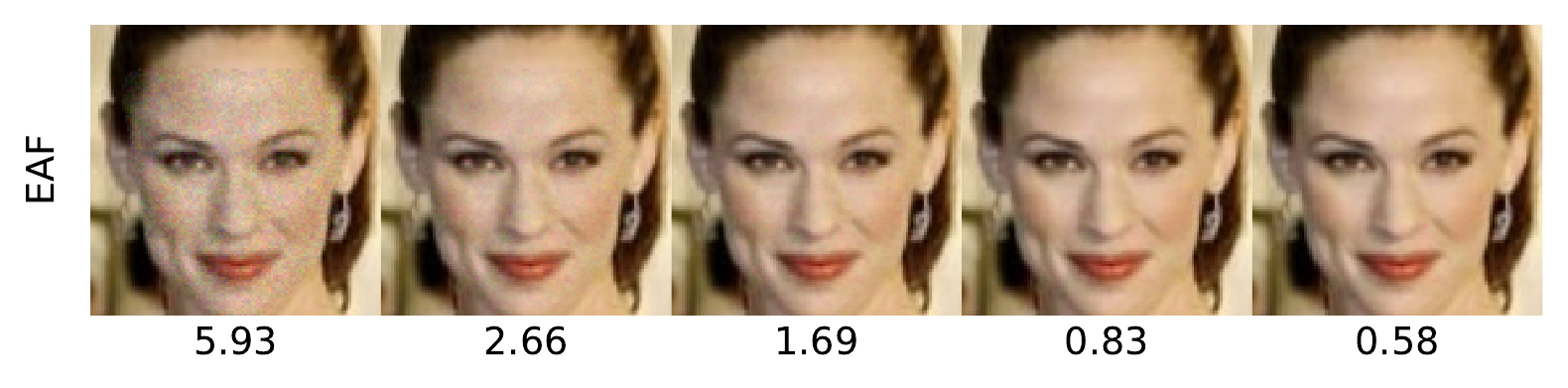}          \includegraphics[width=\textwidth,trim={0cm 0.3cm 0cm 0cm},clip]{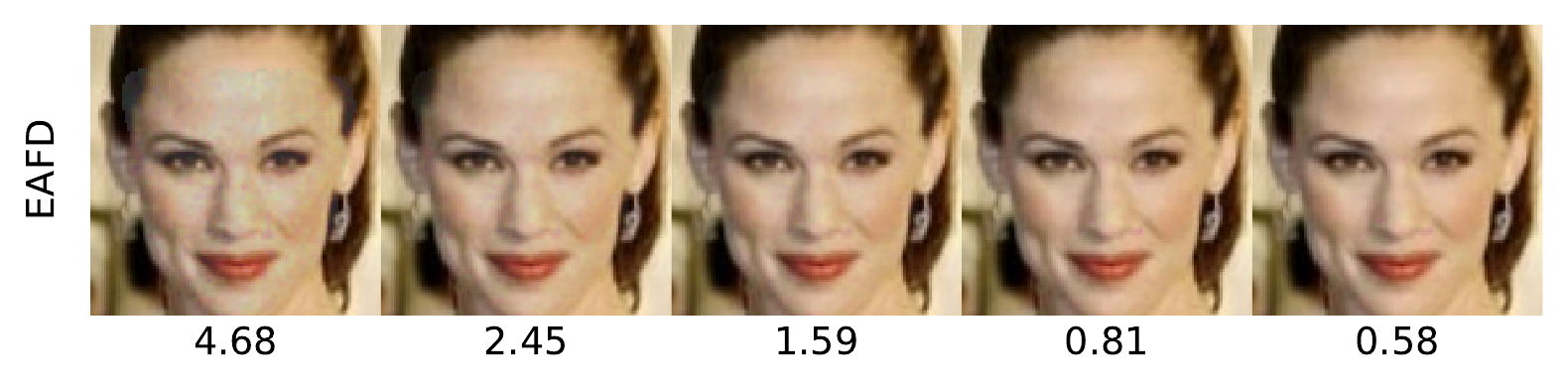}
     \end{subfigure}
 
        %\vspace{-0.2cm}
        \caption{Qualitative results of dodging attacks on the LFW dataset \cite{LFWTech}. For each attack, we illustrate the minimum norm-adversarial examples in each query budget. The $\ell_2$ norm of perturbation is displayed under each image.}
        \label{fig:dodging_lfw2}
\end{figure}

\begin{figure}[t]
     \centering
     \begin{subfigure}[b]{0.47\textwidth}
         \centering
         \includegraphics[width=\textwidth,trim={0cm 0.3cm 0cm 0cm},clip]{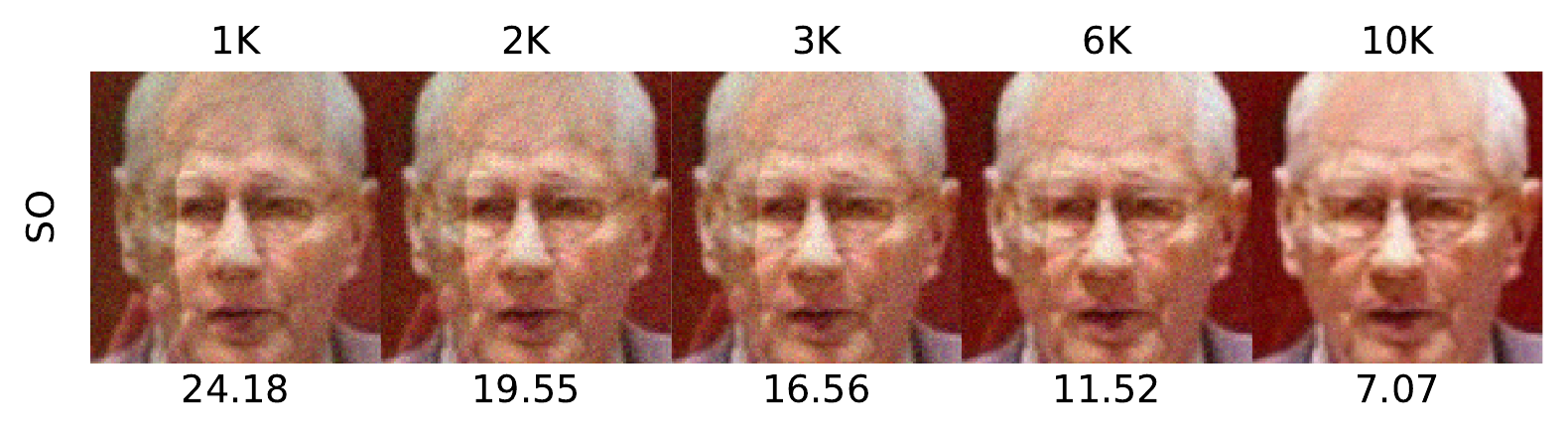}
         \includegraphics[width=\textwidth,trim={0cm 0.3cm 0cm 0cm},clip]{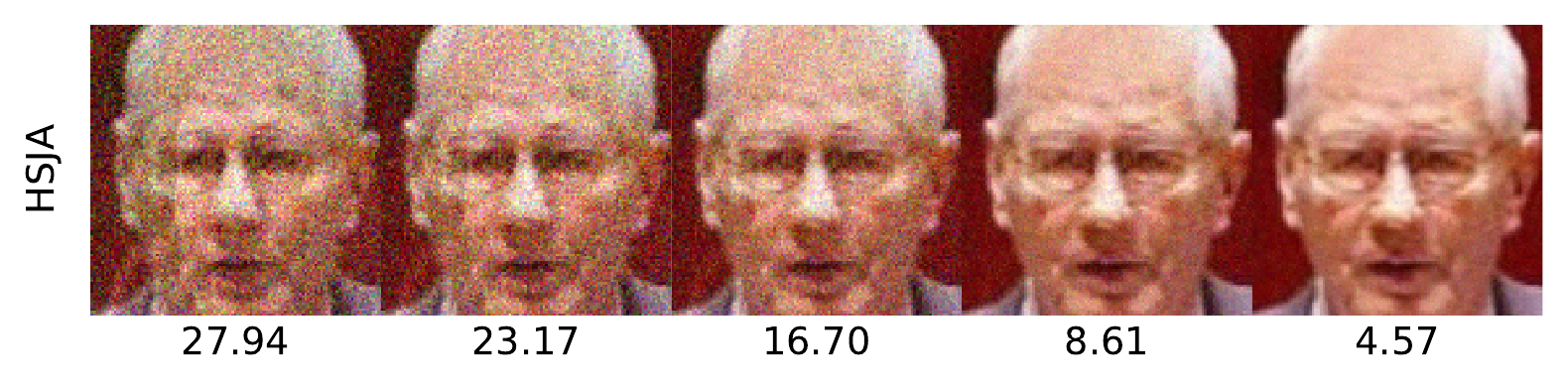}
         \includegraphics[width=\textwidth,trim={0cm 0.3cm 0cm 0cm},clip]{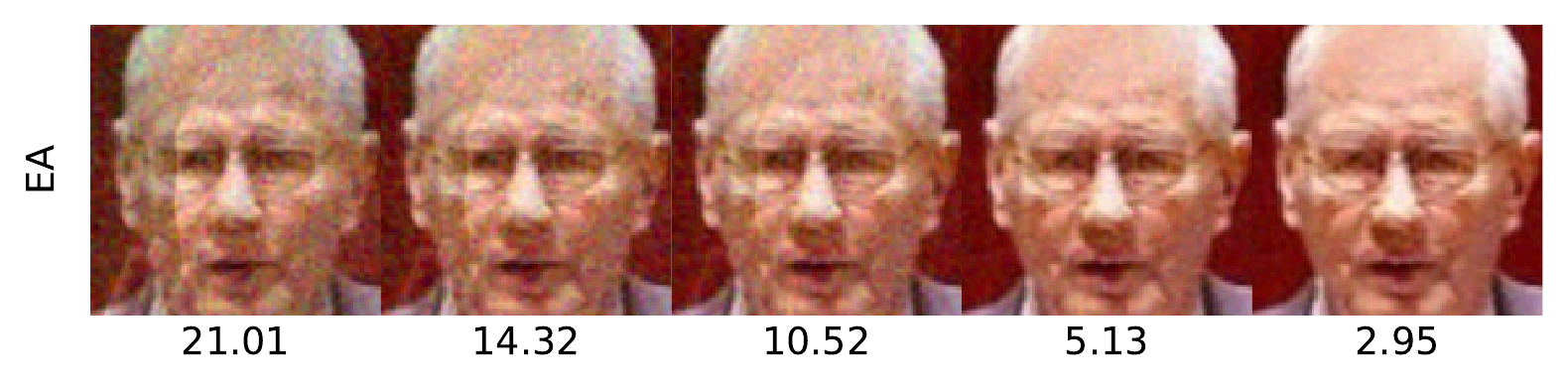}
         \includegraphics[width=\textwidth,trim={0cm 0.3cm 0cm 0cm},clip]{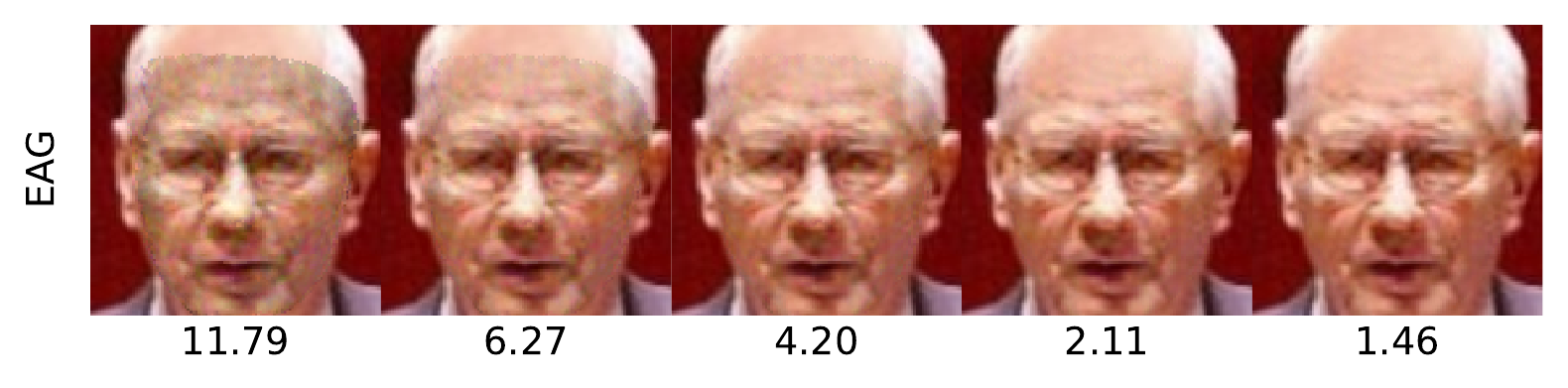}
           \includegraphics[width=\textwidth,trim={0cm 0.3cm 0cm 0cm},clip]{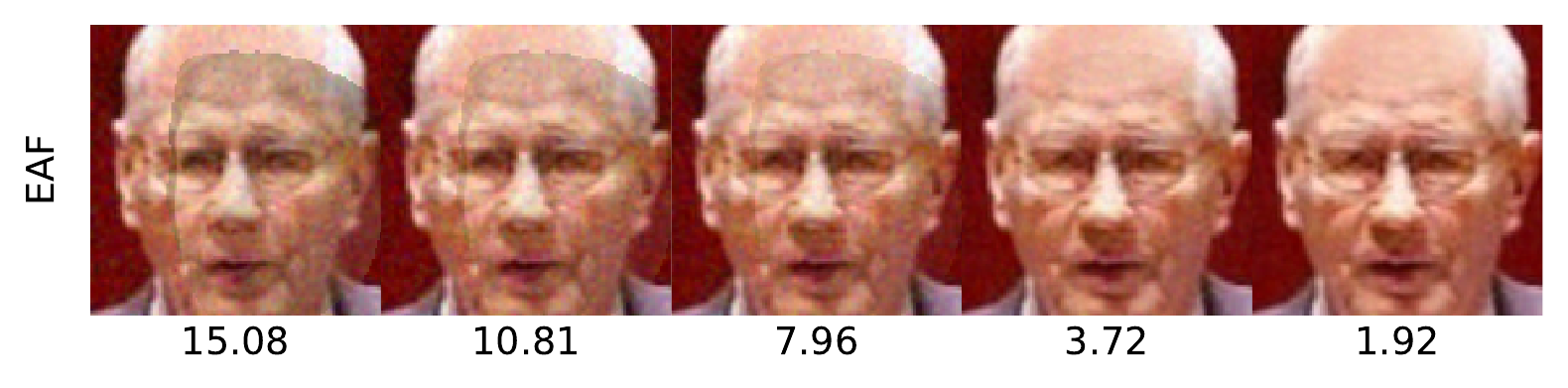}
     \end{subfigure}
        %\vspace{-0.2cm}
        \caption{Qualitative results of impersonation attacks on the LFW dataset \cite{LFWTech}. For each attack, we illustrate the minimum norm-adversarial examples in each query budget. The $\ell_2$ norm of perturbation is displayed under each image.}
        \label{fig:impersonation_lfw1}
\end{figure}

\begin{figure}[t]
     \centering
     \begin{subfigure}[b]{0.47\textwidth}
         \centering
         \includegraphics[width=\textwidth,trim={0cm 0.3cm 0cm 0cm},clip]{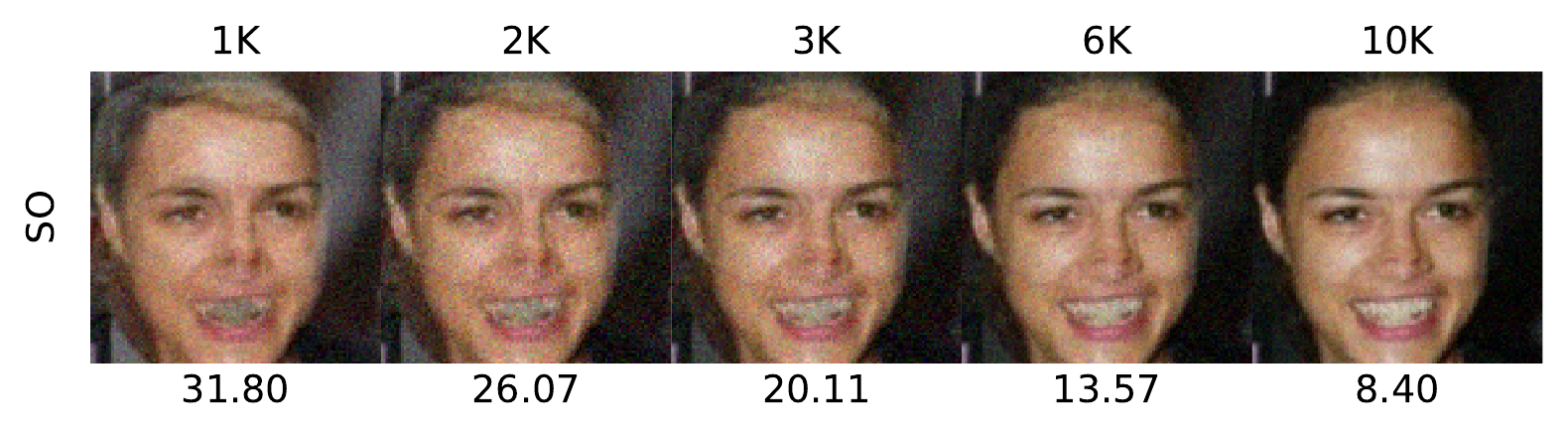}
         \includegraphics[width=\textwidth,trim={0cm 0.3cm 0cm 0cm},clip]{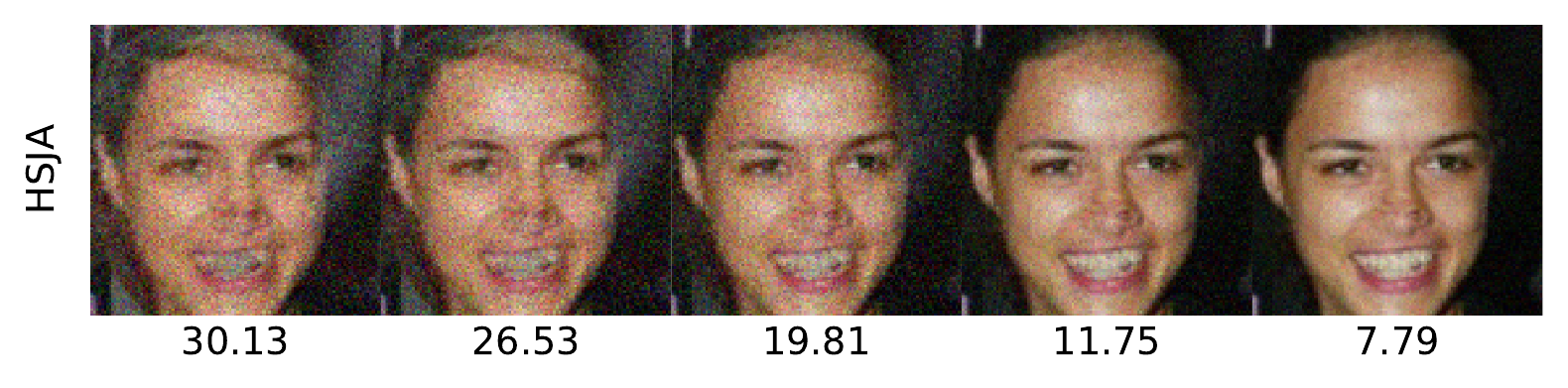}
         \includegraphics[width=\textwidth,trim={0cm 0.3cm 0cm 0cm},clip]{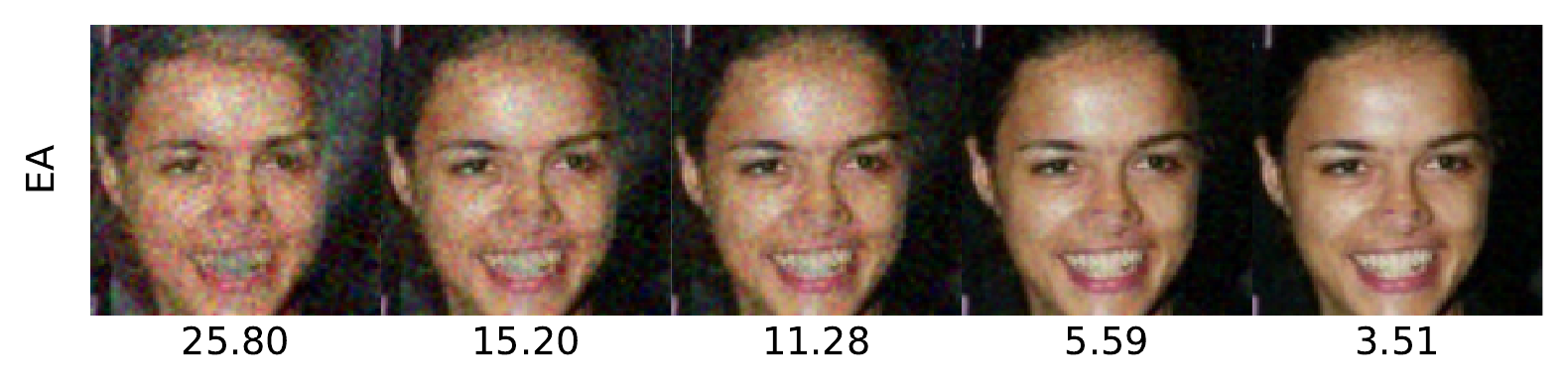}
         \includegraphics[width=\textwidth,trim={0cm 0.3cm 0cm 0cm},clip]{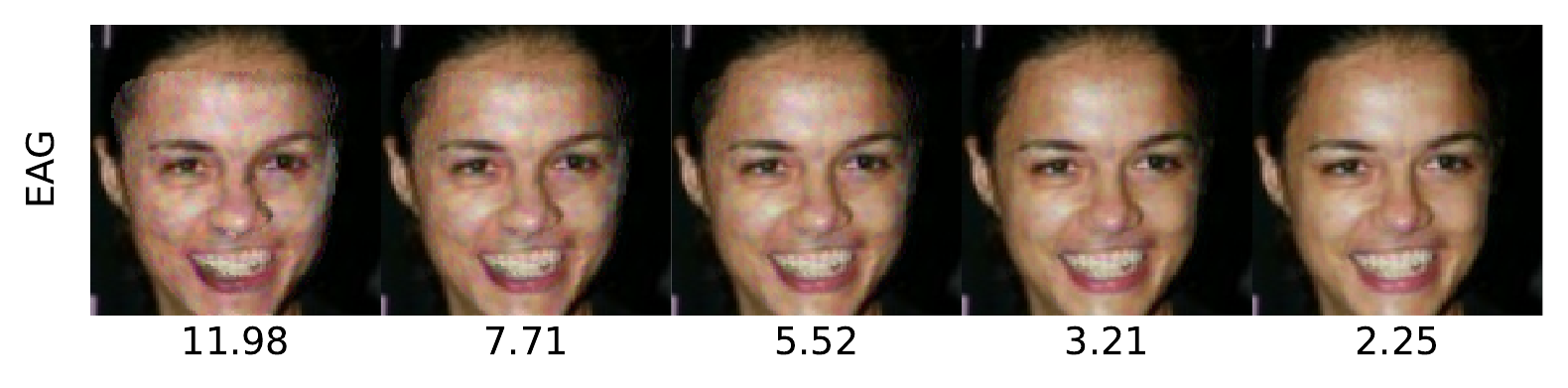}
          \includegraphics[width=\textwidth,trim={0cm 0.3cm 0cm 0cm},clip]{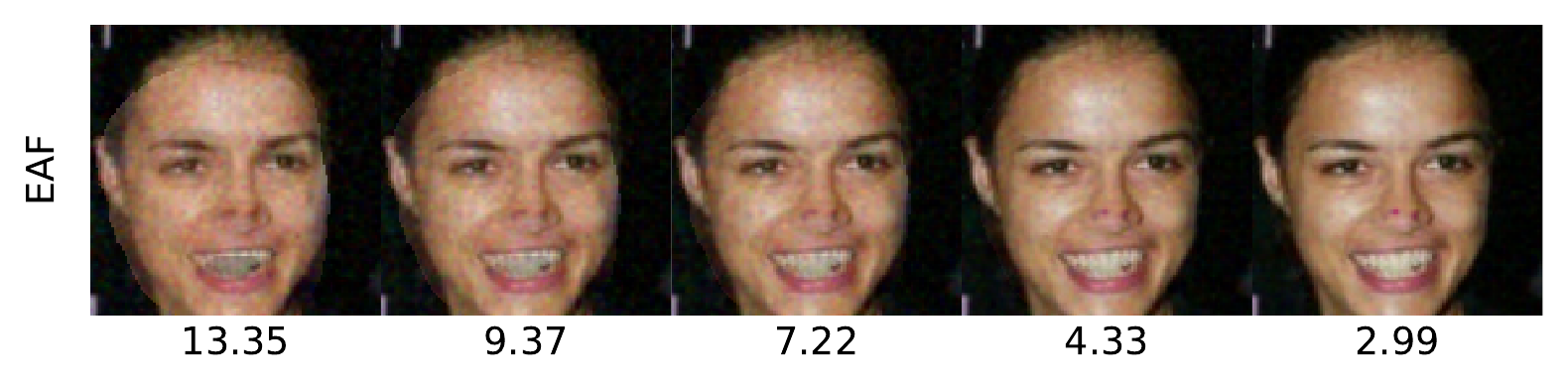}
     \end{subfigure}
        %\vspace{-0.2cm}
        \caption{Qualitative results of impersonation attacks on the LFW dataset \cite{LFWTech}. For each attack, we illustrate the minimum norm-adversarial examples in each query budget. The $\ell_2$ norm of perturbation is displayed under each image.}
        \label{fig:impersonation_lfw2}
\end{figure}

\begin{figure}[t]
     \centering
     \begin{subfigure}[b]{0.47\textwidth}
         \centering
         \includegraphics[width=\textwidth,trim={0cm 0.3cm 0cm 0cm},clip]{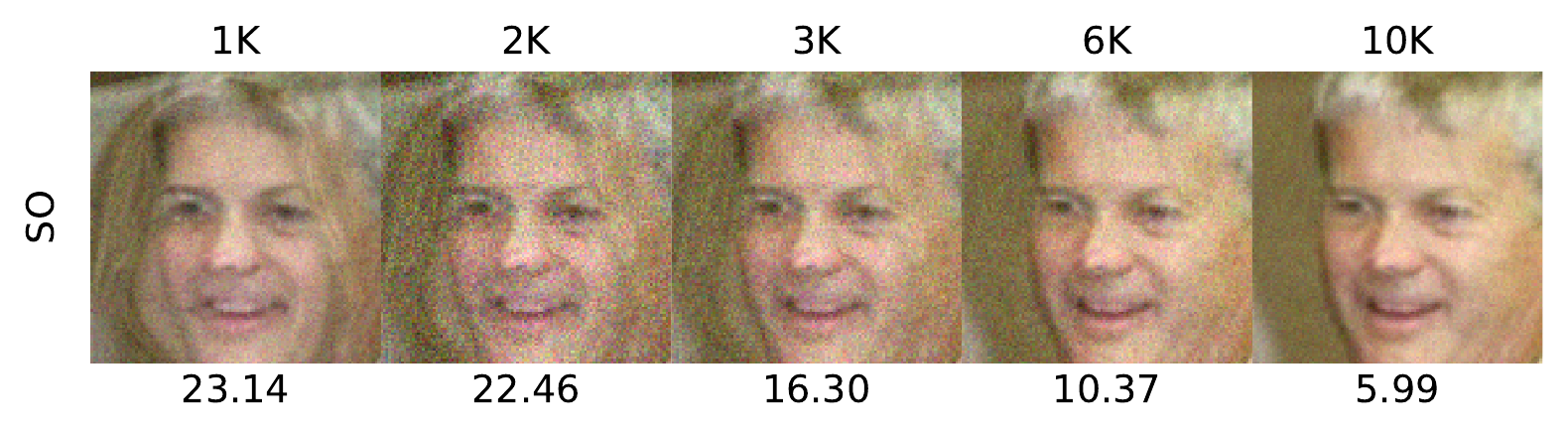}
         \includegraphics[width=\textwidth,trim={0cm 0.3cm 0cm 0cm},clip]{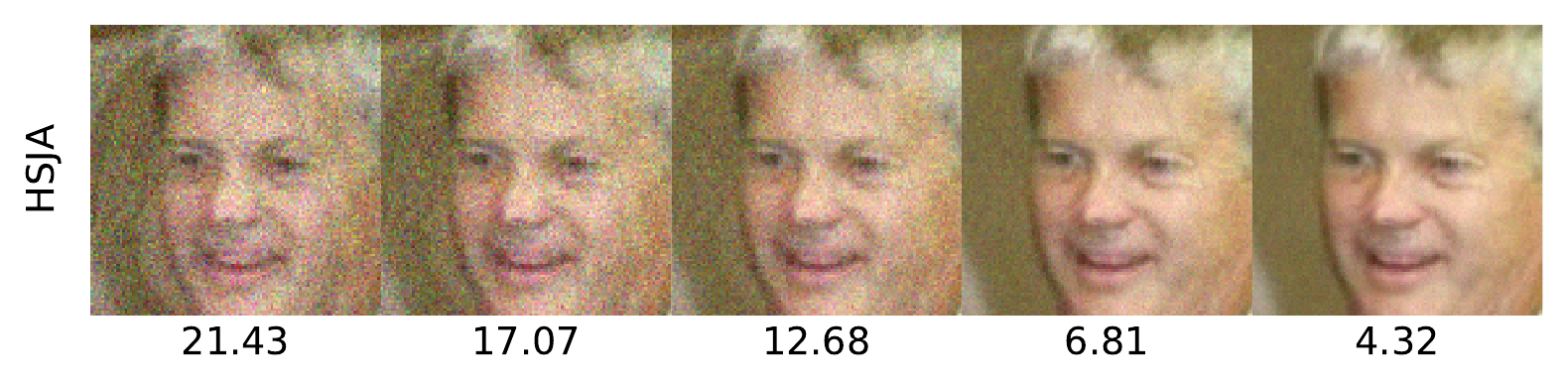}
         \includegraphics[width=\textwidth,trim={0cm 0.3cm 0cm 0cm},clip]{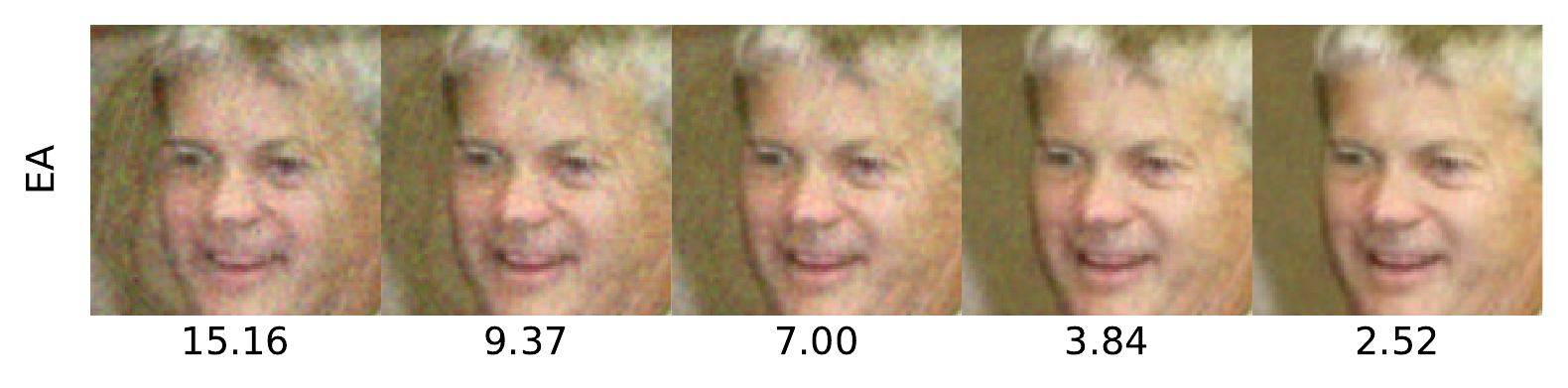}
         \includegraphics[width=\textwidth,trim={0cm 0.3cm 0cm 0cm},clip]{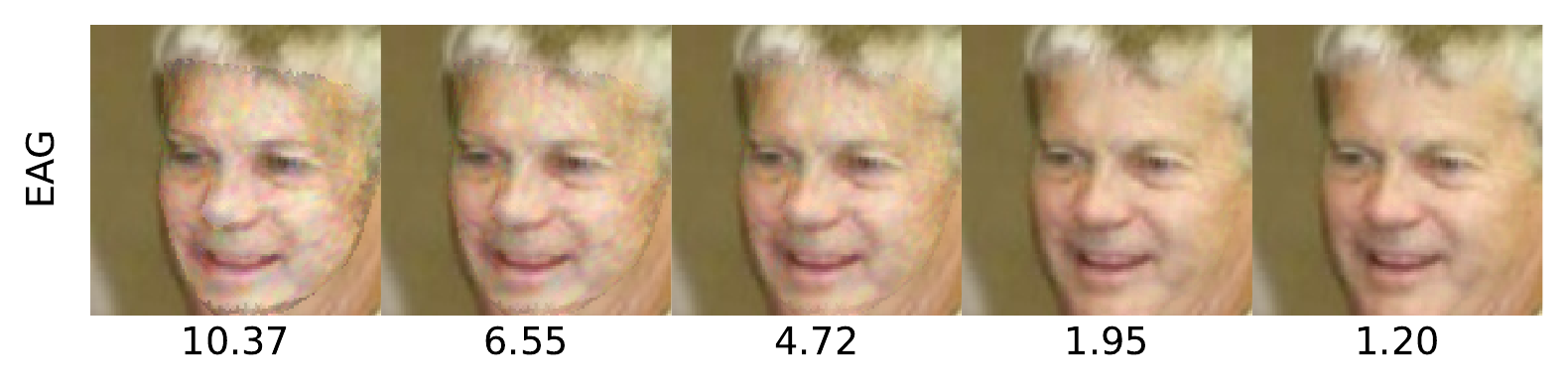}
            \includegraphics[width=\textwidth,trim={0cm 0.3cm 0cm 0cm},clip]{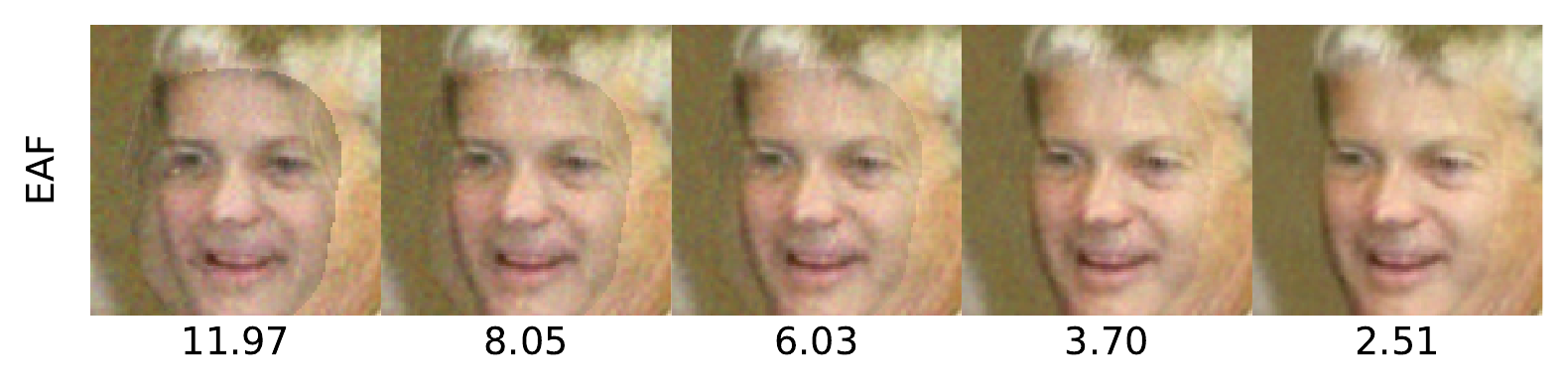}
     \end{subfigure}
        %\vspace{-0.2cm}
        \caption{Qualitative results of impersonation attacks on the LFW dataset \cite{LFWTech}. For each attack, we illustrate the minimum norm-adversarial examples in each query budget. The $\ell_2$ norm of perturbation is displayed under each image.}
        \label{fig:impersonation_lfw3}
\end{figure}

\begin{figure}[t]
     \centering
     \begin{subfigure}[b]{0.47\textwidth}
         \centering
         \includegraphics[width=\textwidth,trim={0cm 0.3cm 0cm 0cm},clip]{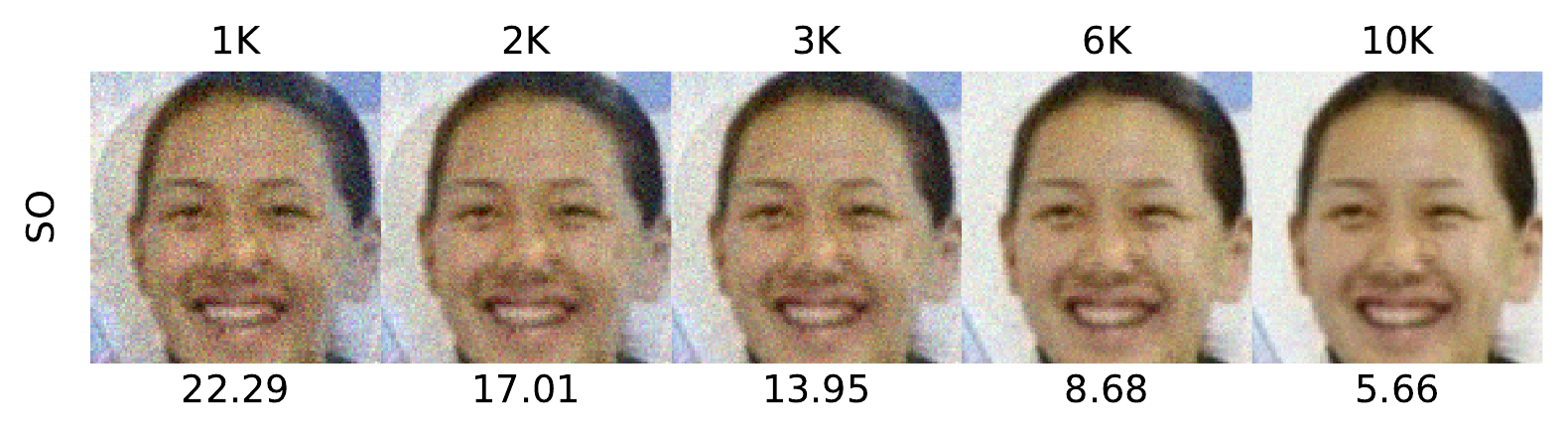}
         \includegraphics[width=\textwidth,trim={0cm 0.3cm 0cm 0cm},clip]{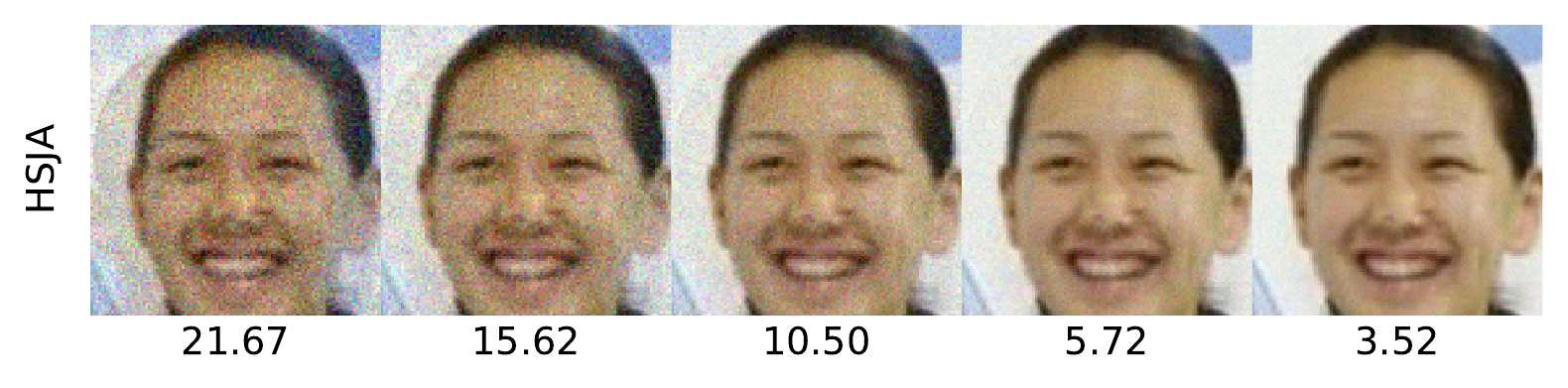}
         \includegraphics[width=\textwidth,trim={0cm 0.3cm 0cm 0cm},clip]{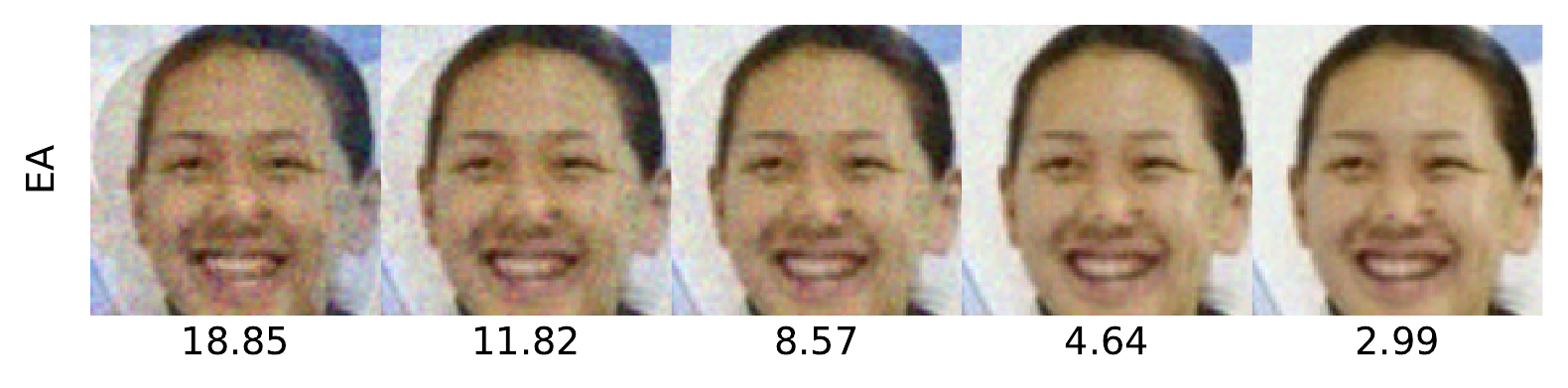}
         \includegraphics[width=\textwidth,trim={0cm 0.3cm 0cm 0cm},clip]{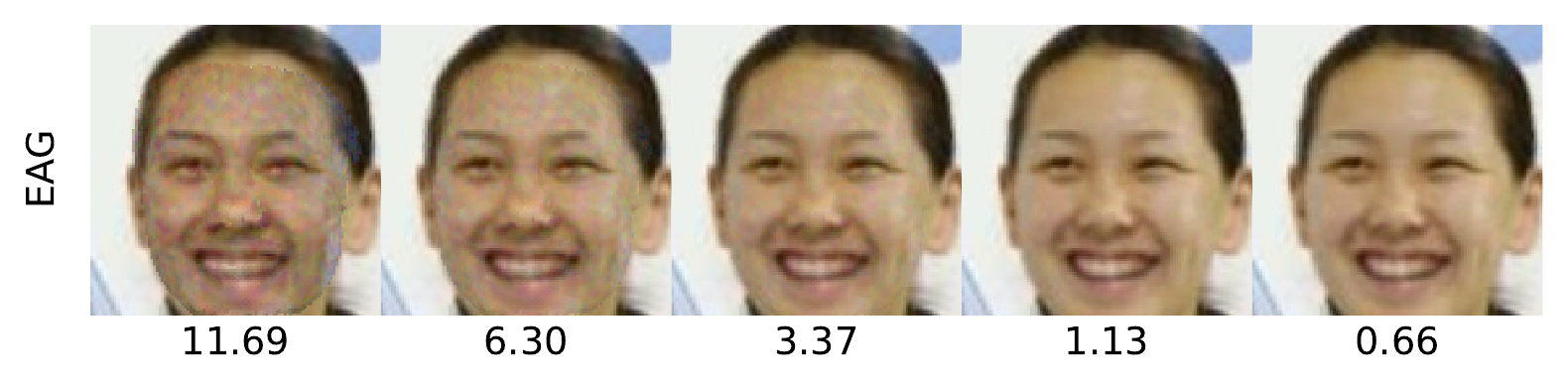}
             \includegraphics[width=\textwidth,trim={0cm 0.3cm 0cm 0cm},clip]{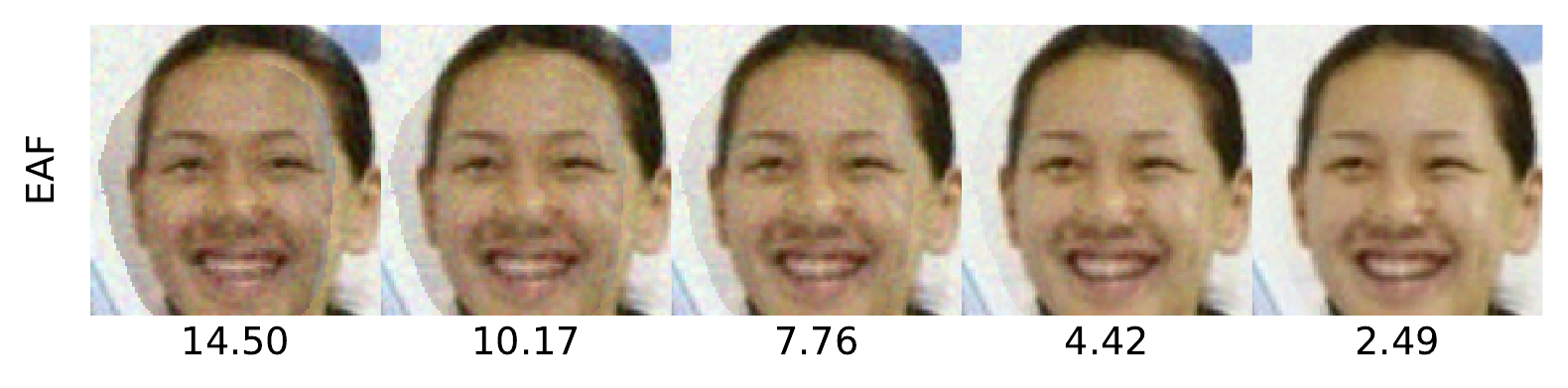}
     \end{subfigure}
        %\vspace{-0.2cm}
        \caption{Qualitative results of impersonation attacks on the LFW dataset \cite{LFWTech}. For each attack, we illustrate the minimum norm-adversarial examples in each query budget. The $\ell_2$ norm of perturbation is displayed under each image.}
        \label{fig:impersonation_lfw4}
\end{figure}

\begin{figure}[t]
     \centering
     \begin{subfigure}[b]{0.47\textwidth}
         \centering
         \includegraphics[width=\textwidth,trim={0cm 0.3cm 0cm 0cm},clip]{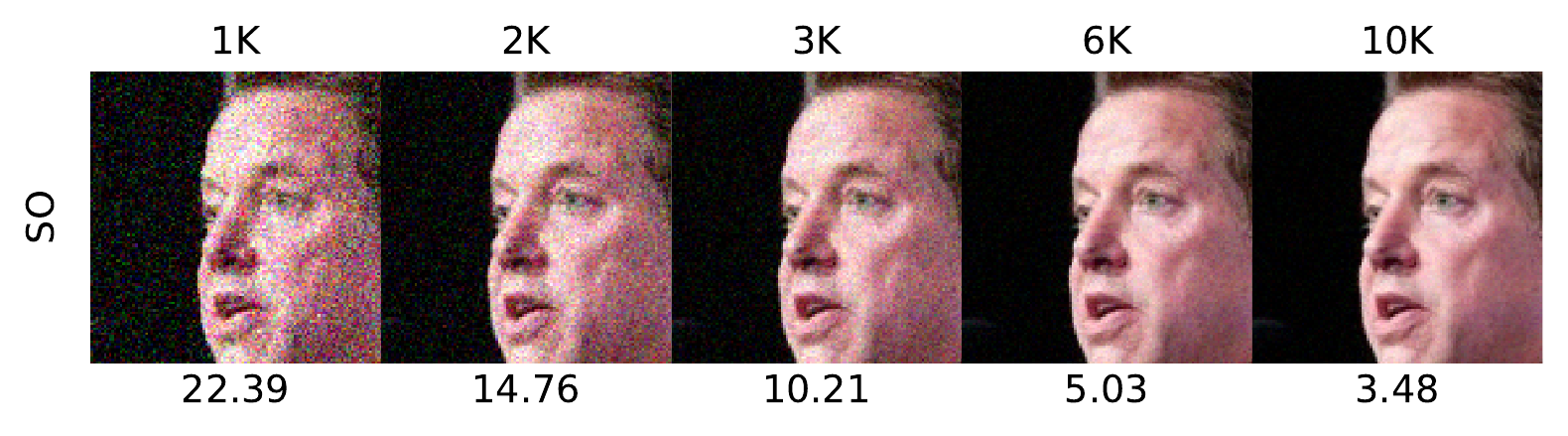}
         \includegraphics[width=\textwidth,trim={0cm 0.3cm 0cm 0cm},clip]{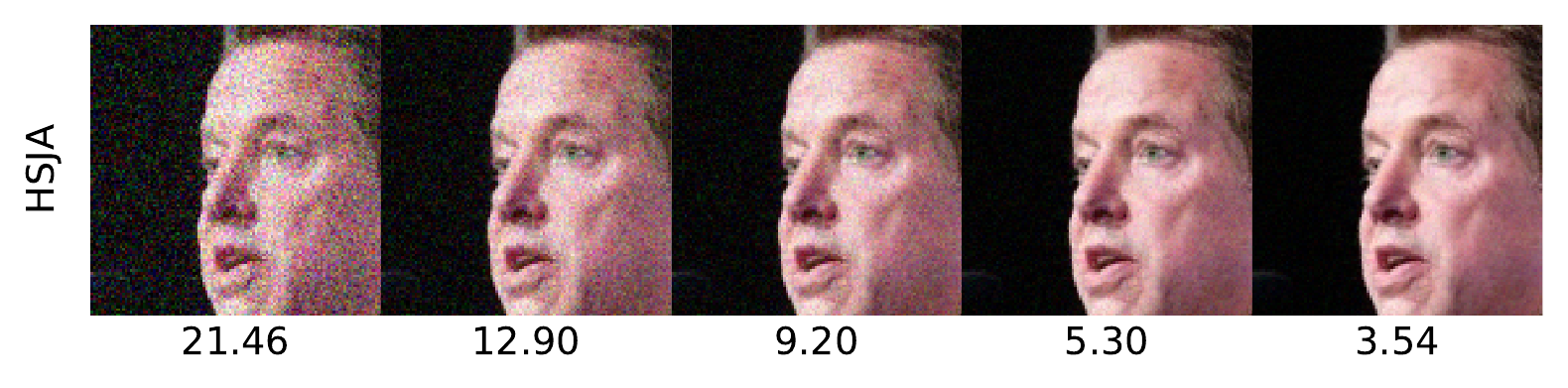}         \includegraphics[width=\textwidth,trim={0cm 0.3cm 0cm 0cm},clip]{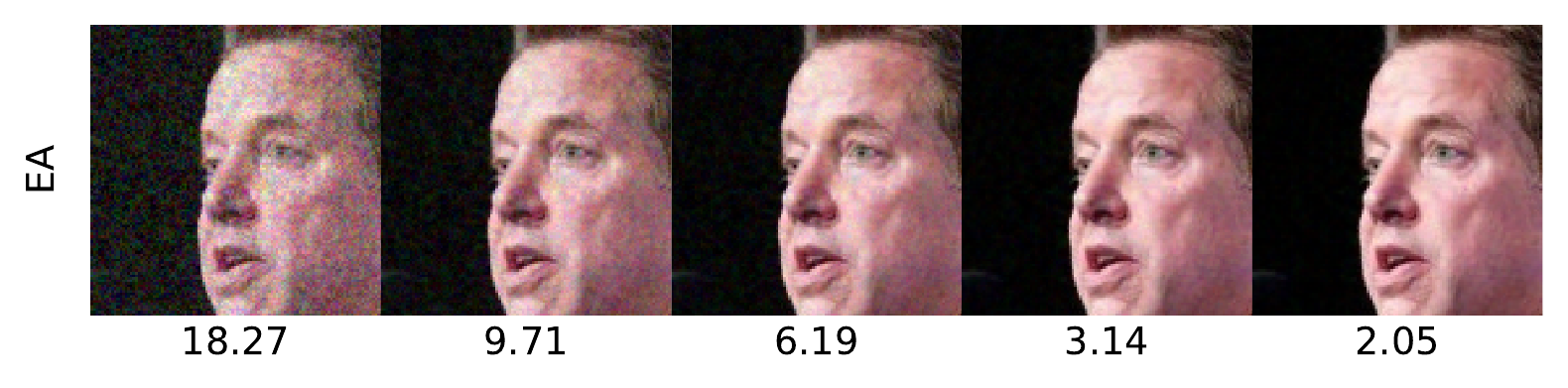}
         \includegraphics[width=\textwidth,trim={0cm 0.3cm 0cm 0cm},clip]{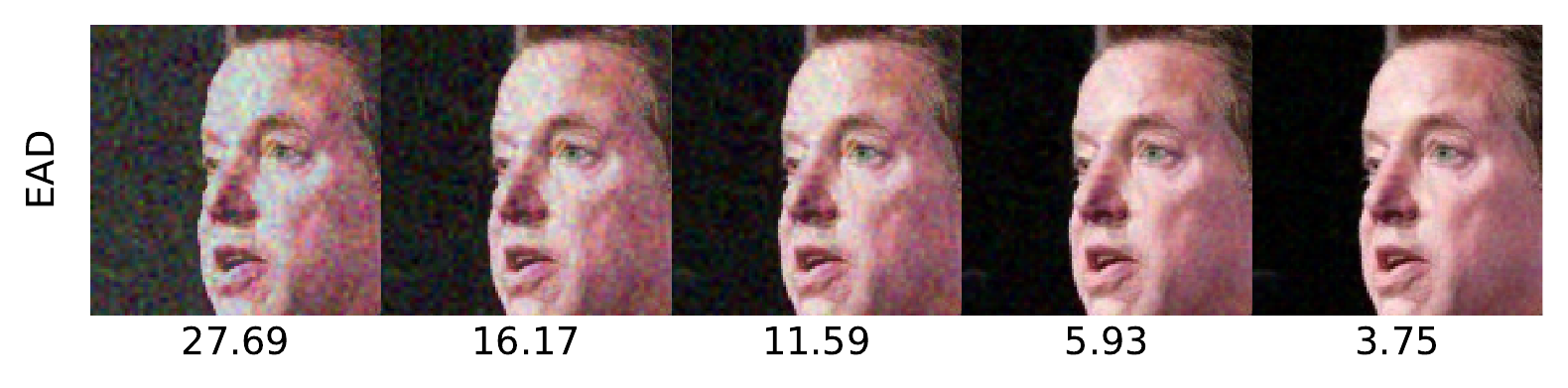}
          \includegraphics[width=\textwidth,trim={0cm 0.3cm 0cm 0cm},clip]{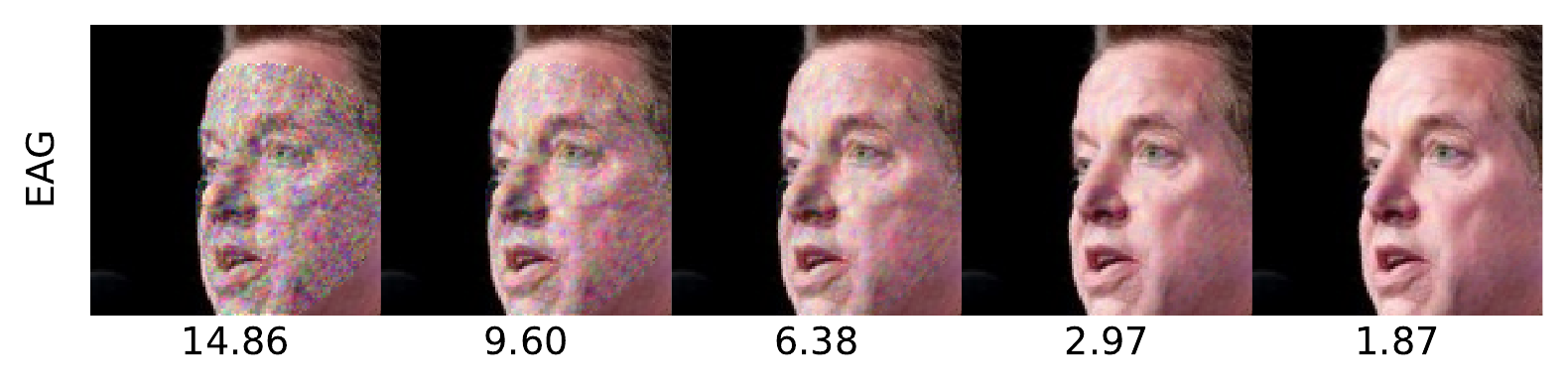}          \includegraphics[width=\textwidth,trim={0cm 0.3cm 0cm 0cm},clip]{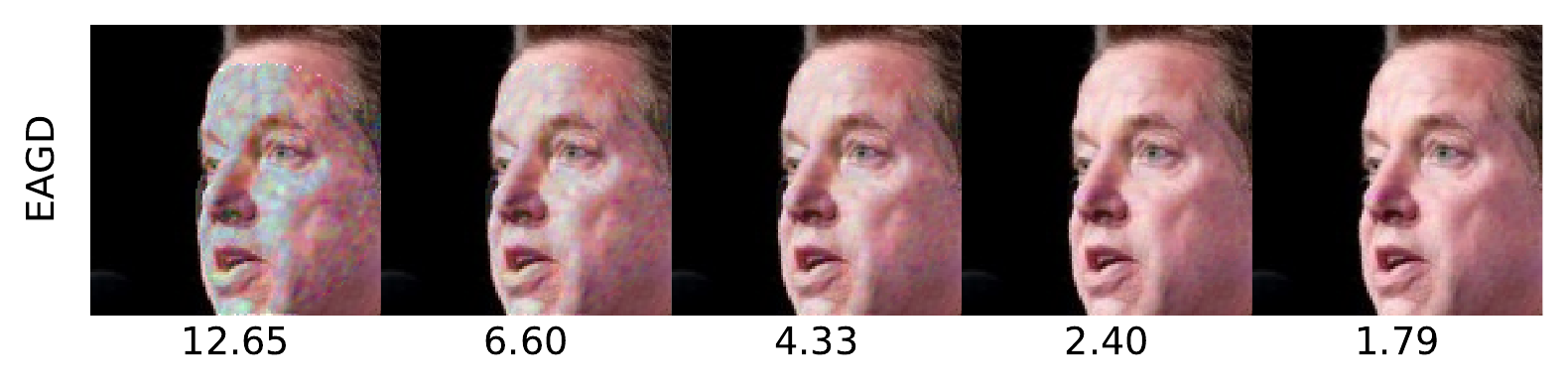}
            \includegraphics[width=\textwidth,trim={0cm 0.3cm 0cm 0cm},clip]{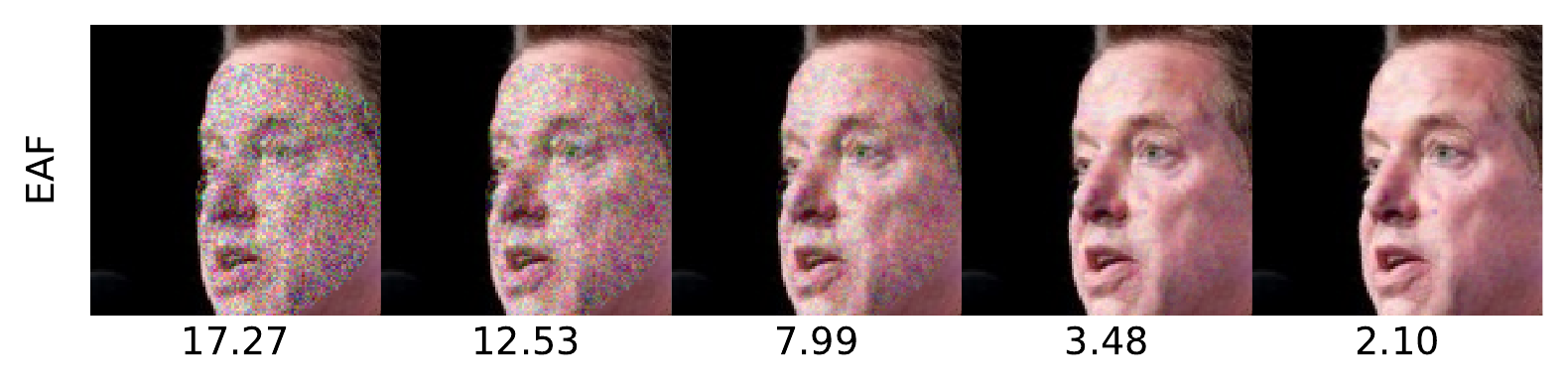}          \includegraphics[width=\textwidth,trim={0cm 0.3cm 0cm 0cm},clip]{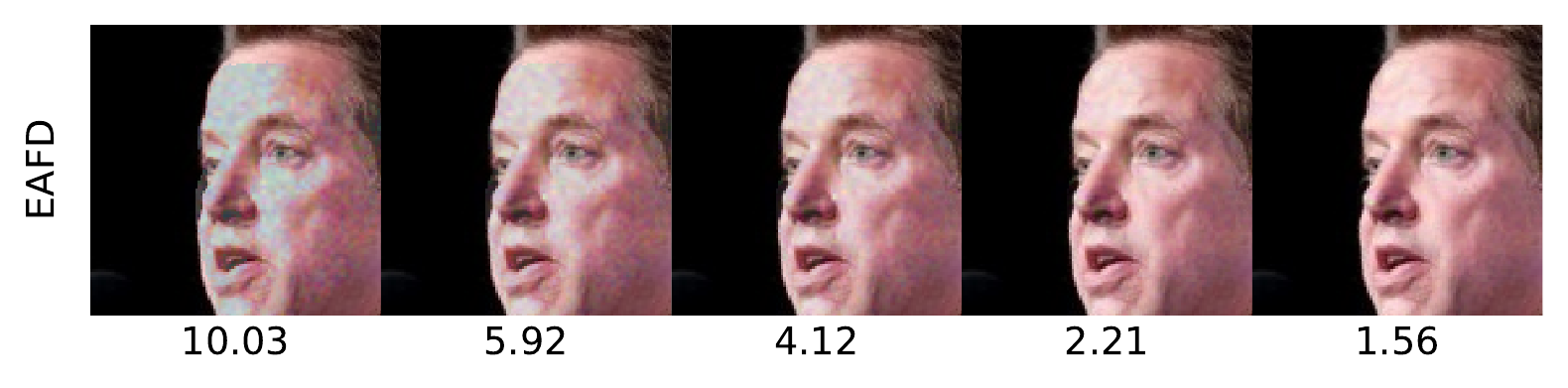}
     \end{subfigure}
 
        %\vspace{-0.2cm}
        \caption{Qualitative results of dodging attacks on the CPLFW dataset \cite{CPLFWTech}. For each attack, we illustrate the minimum norm-adversarial examples in each query budget. The $\ell_2$ norm of perturbation is displayed under each image.}
        \label{fig:dodging_cplfw1}
\end{figure}

\begin{figure}[t]
     \centering
     \begin{subfigure}[b]{0.47\textwidth}
         \centering
         \includegraphics[width=\textwidth,trim={0cm 0.3cm 0cm 0cm},clip]{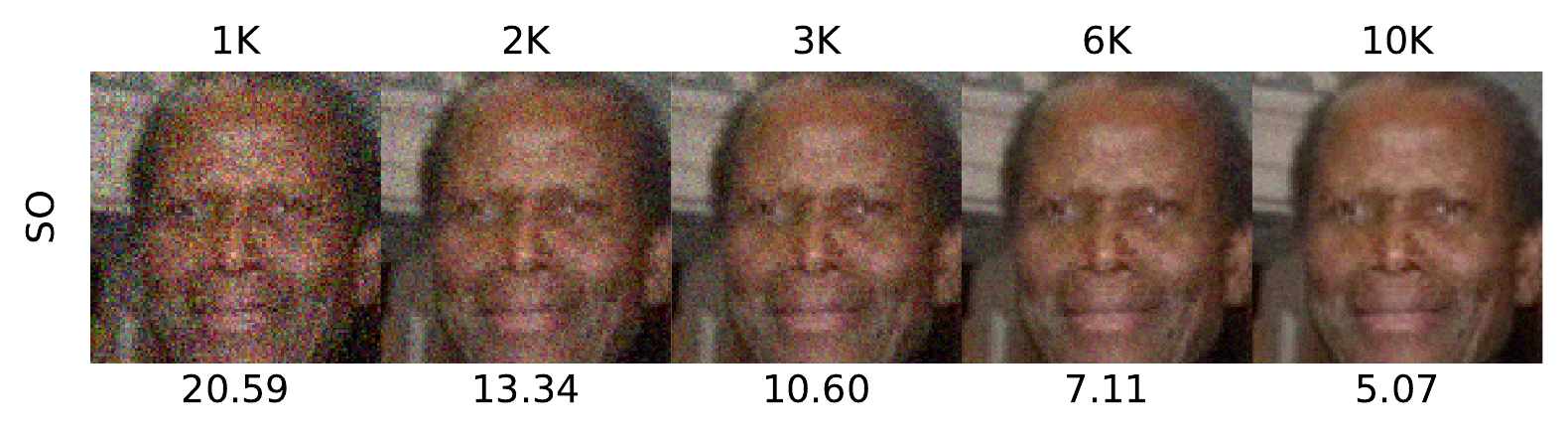}
         \includegraphics[width=\textwidth,trim={0cm 0.3cm 0cm 0cm},clip]{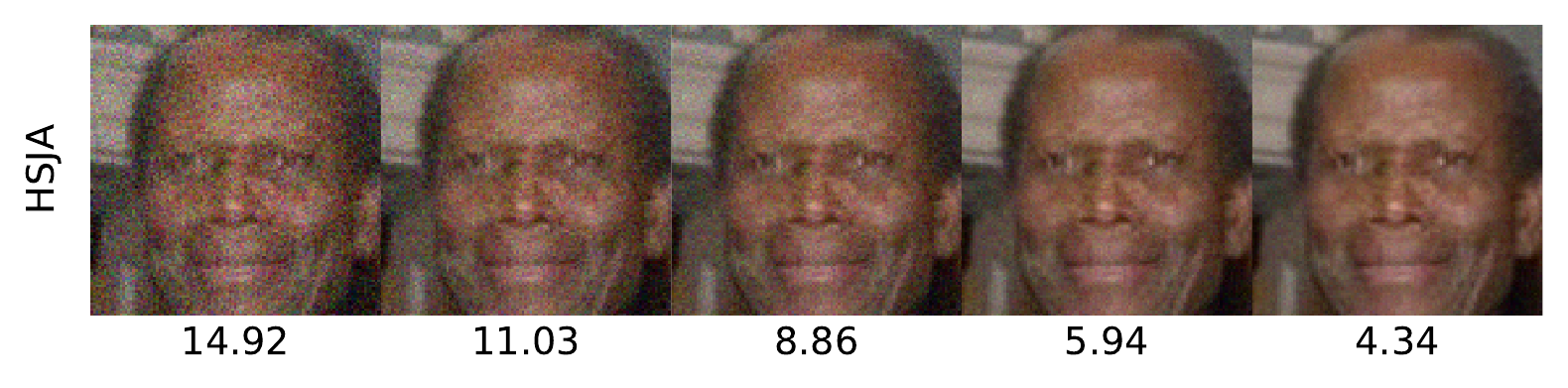}         \includegraphics[width=\textwidth,trim={0cm 0.3cm 0cm 0cm},clip]{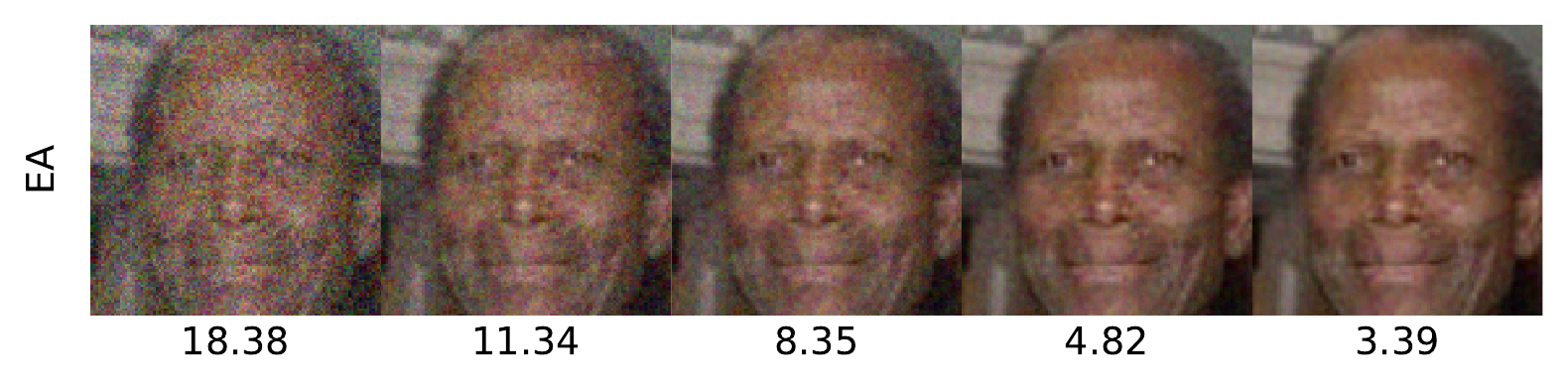}
         \includegraphics[width=\textwidth,trim={0cm 0.3cm 0cm 0cm},clip]{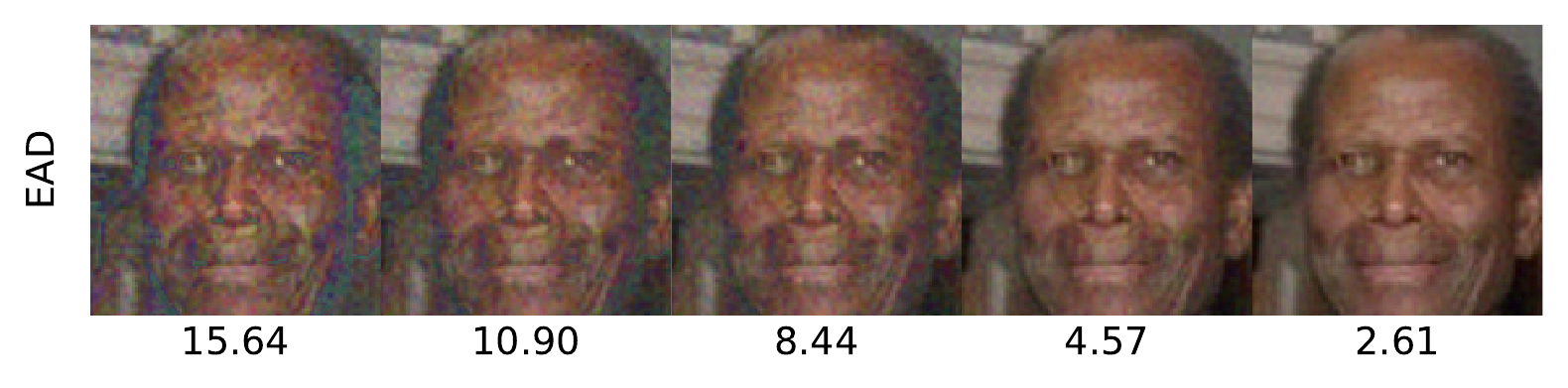}
          \includegraphics[width=\textwidth,trim={0cm 0.3cm 0cm 0cm},clip]{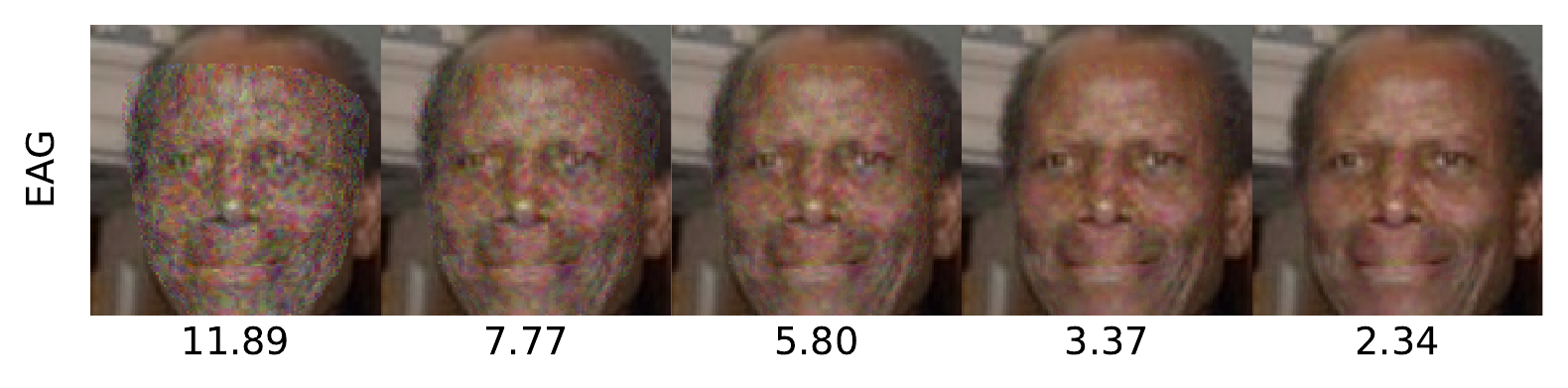}          \includegraphics[width=\textwidth,trim={0cm 0.3cm 0cm 0cm},clip]{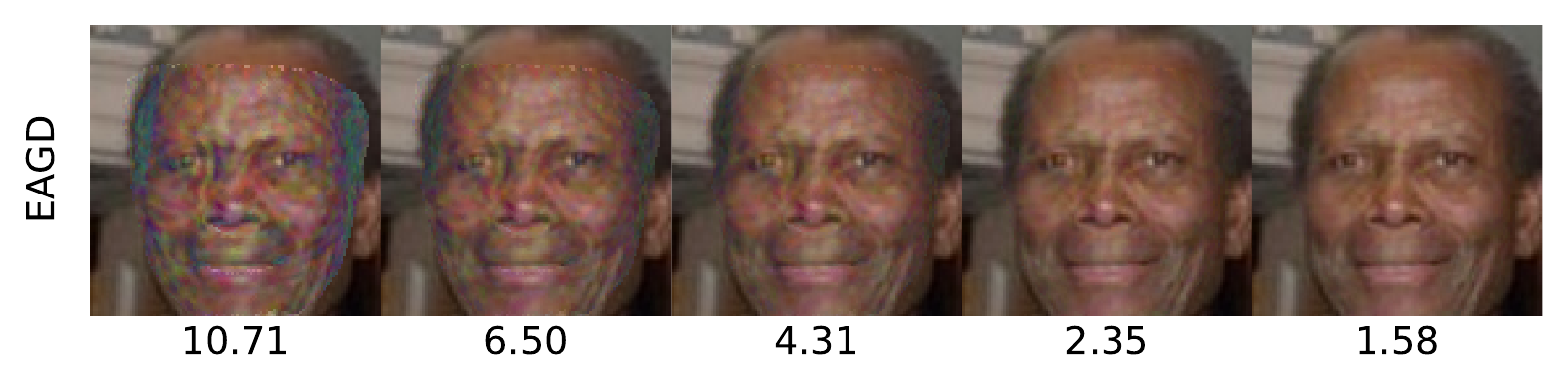}
              \includegraphics[width=\textwidth,trim={0cm 0.3cm 0cm 0cm},clip]{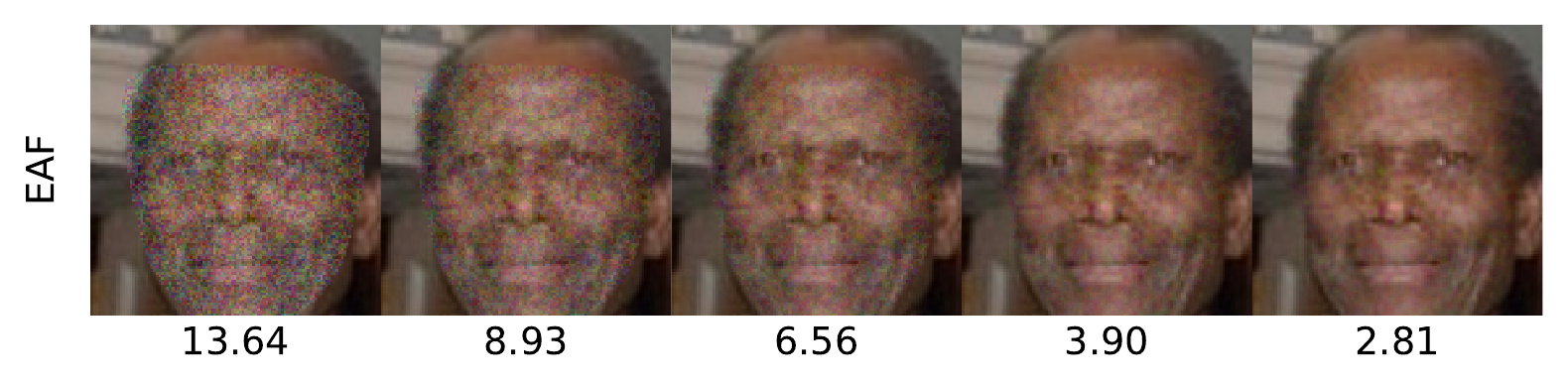}          \includegraphics[width=\textwidth,trim={0cm 0.3cm 0cm 0cm},clip]{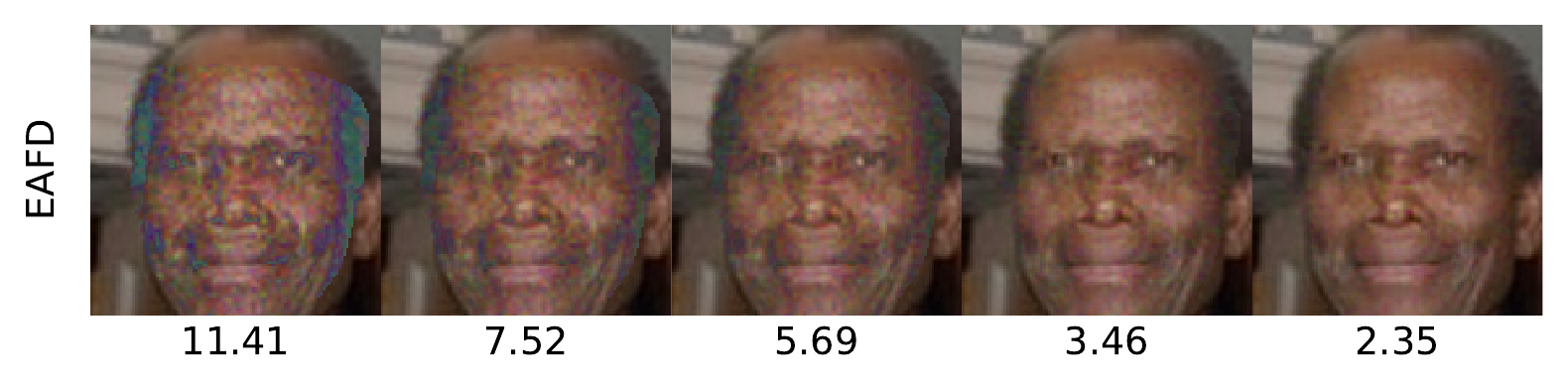}
     \end{subfigure}
 
        %\vspace{-0.2cm}
        \caption{Qualitative results of dodging attacks on the CPLFW dataset \cite{CPLFWTech}. For each attack, we illustrate the minimum norm-adversarial examples in each query budget. The $\ell_2$ norm of perturbation is displayed under each image.}
        \label{fig:dodging_cplfw2}
\end{figure}

\begin{figure}[t]
     \centering
     \begin{subfigure}[b]{0.47\textwidth}
         \centering
         \includegraphics[width=\textwidth,trim={0cm 0.3cm 0cm 0cm},clip]{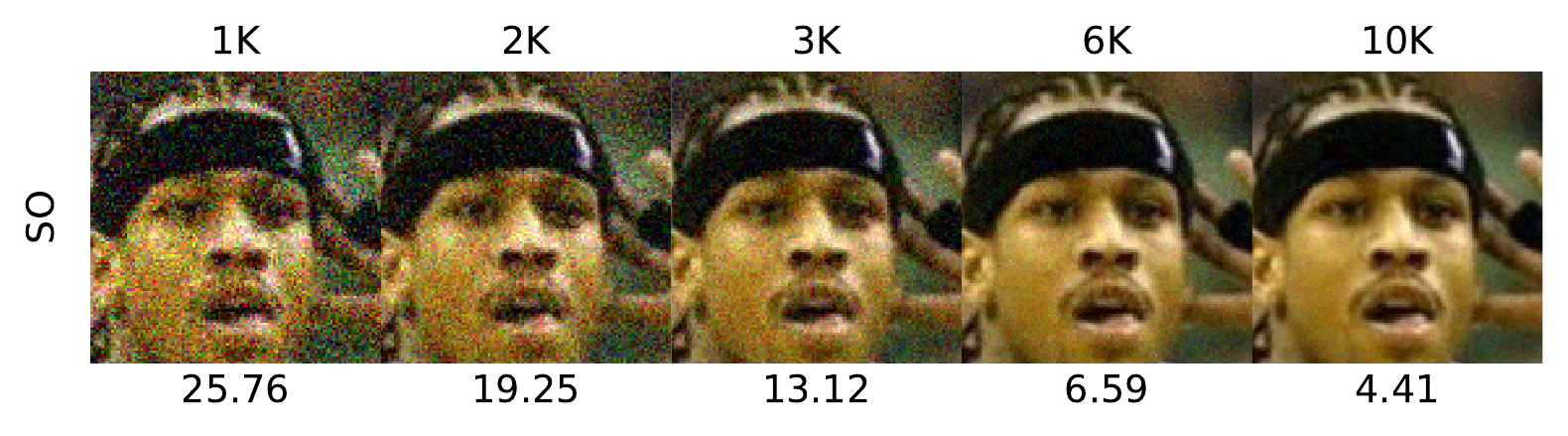}
         \includegraphics[width=\textwidth,trim={0cm 0.3cm 0cm 0cm},clip]{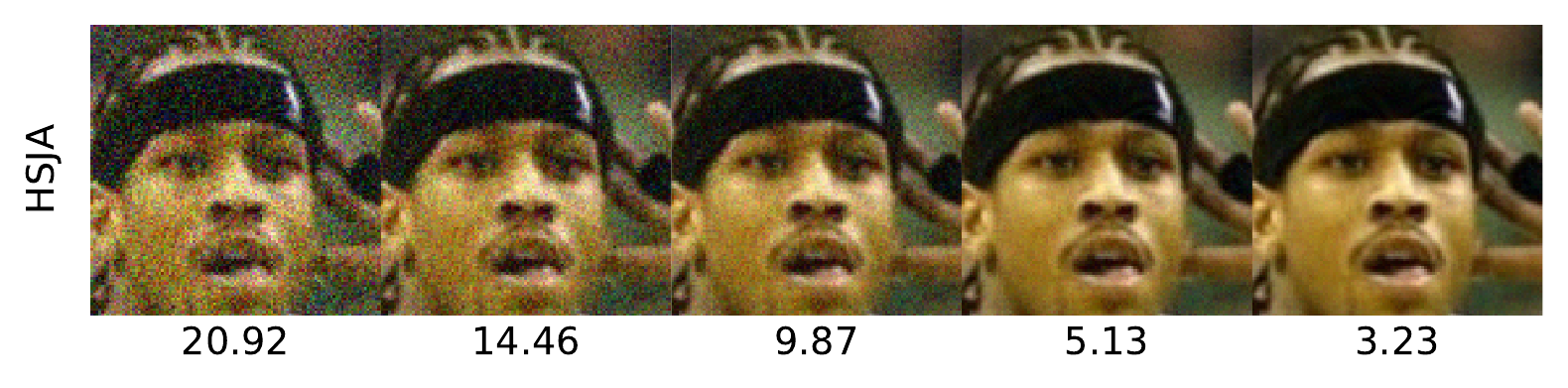}         \includegraphics[width=\textwidth,trim={0cm 0.3cm 0cm 0cm},clip]{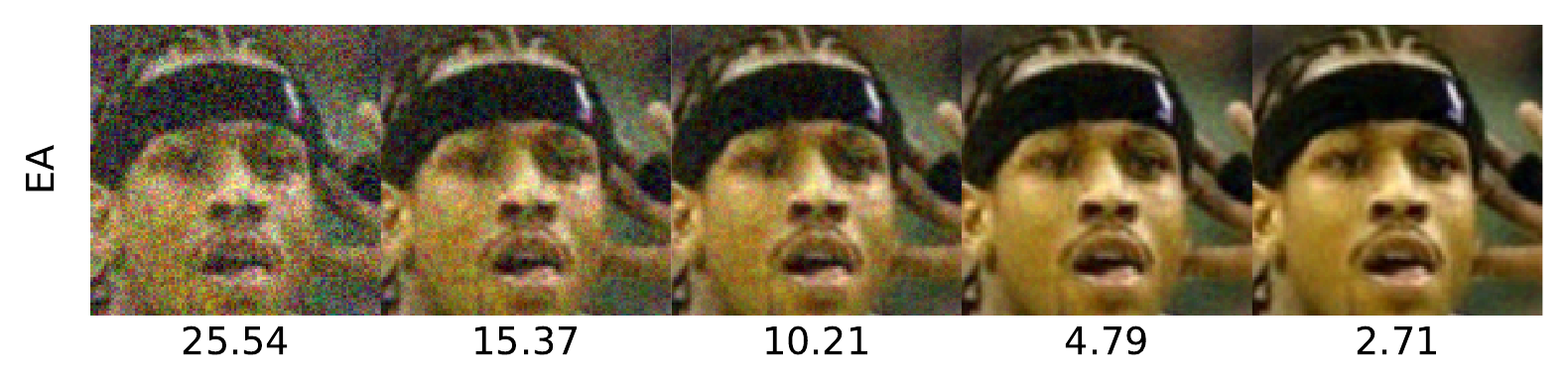}
         \includegraphics[width=\textwidth,trim={0cm 0.3cm 0cm 0cm},clip]{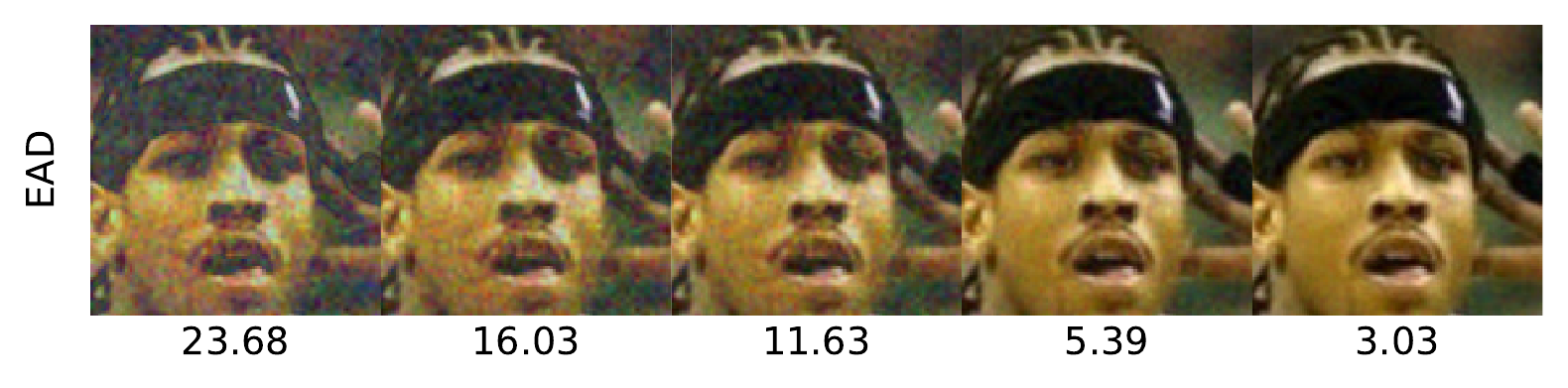}
          \includegraphics[width=\textwidth,trim={0cm 0.3cm 0cm 0cm},clip]{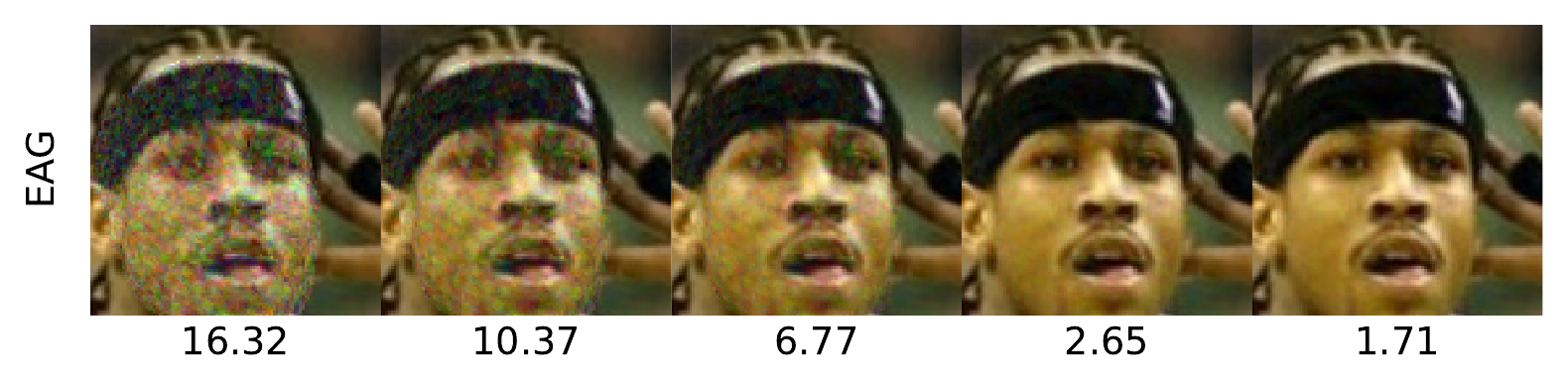}          \includegraphics[width=\textwidth,trim={0cm 0.3cm 0cm 0cm},clip]{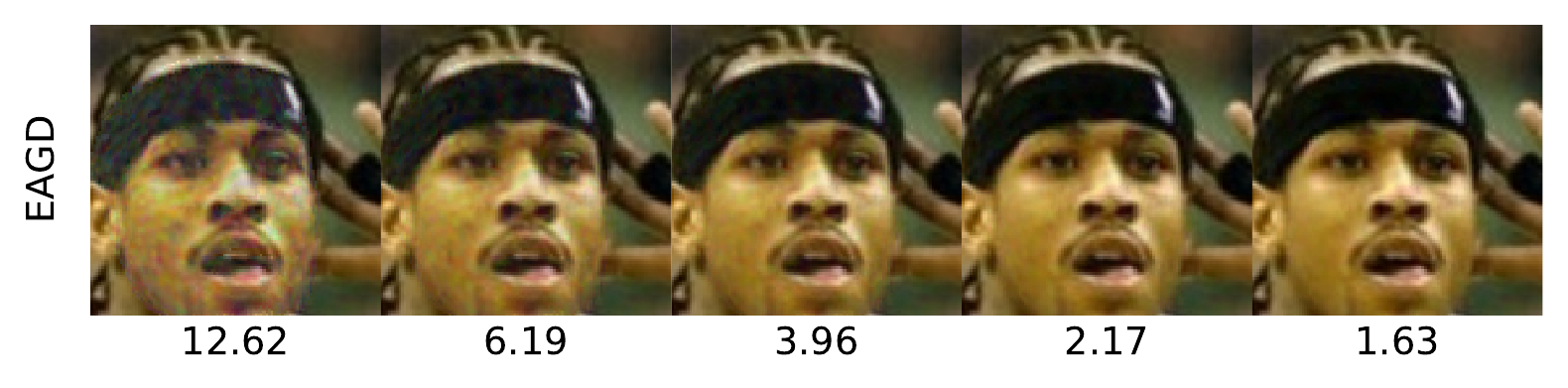}
              \includegraphics[width=\textwidth,trim={0cm 0.3cm 0cm 0cm},clip]{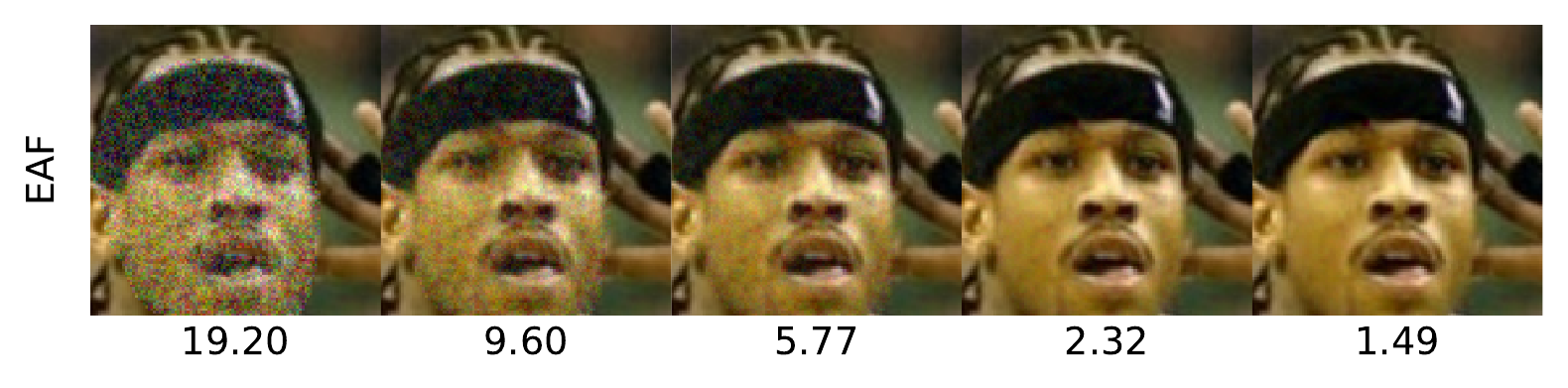}          \includegraphics[width=\textwidth,trim={0cm 0.3cm 0cm 0cm},clip]{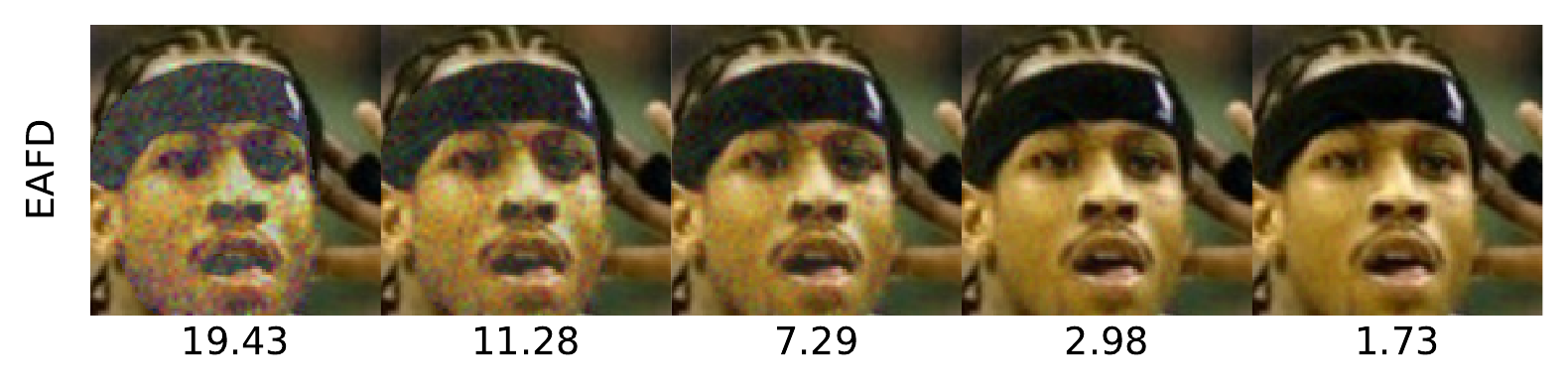}
     \end{subfigure}
 
        %\vspace{-0.2cm}
        \caption{Qualitative results of dodging attacks on the CPLFW dataset \cite{CPLFWTech}. For each attack, we illustrate the minimum norm-adversarial examples in each query budget. The $\ell_2$ norm of perturbation is displayed under each image.}
        \label{fig:dodging_cplfw3}
\end{figure}

\begin{figure}[t]
     \centering
     \begin{subfigure}[b]{0.47\textwidth}
         \centering
         \includegraphics[width=\textwidth,trim={0cm 0.3cm 0cm 0cm},clip]{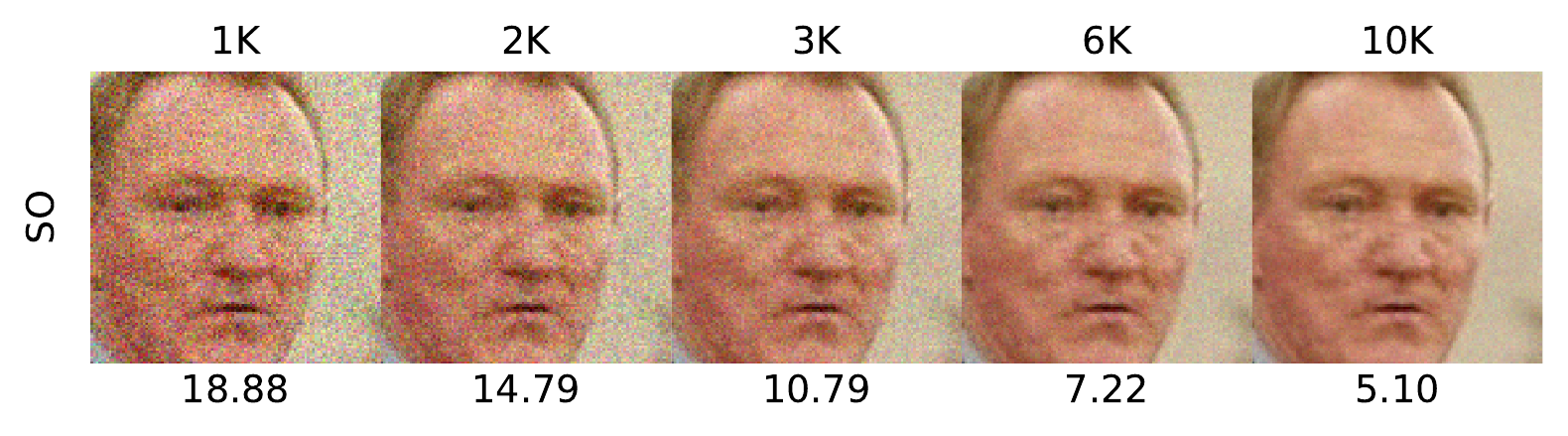}
         \includegraphics[width=\textwidth,trim={0cm 0.3cm 0cm 0cm},clip]{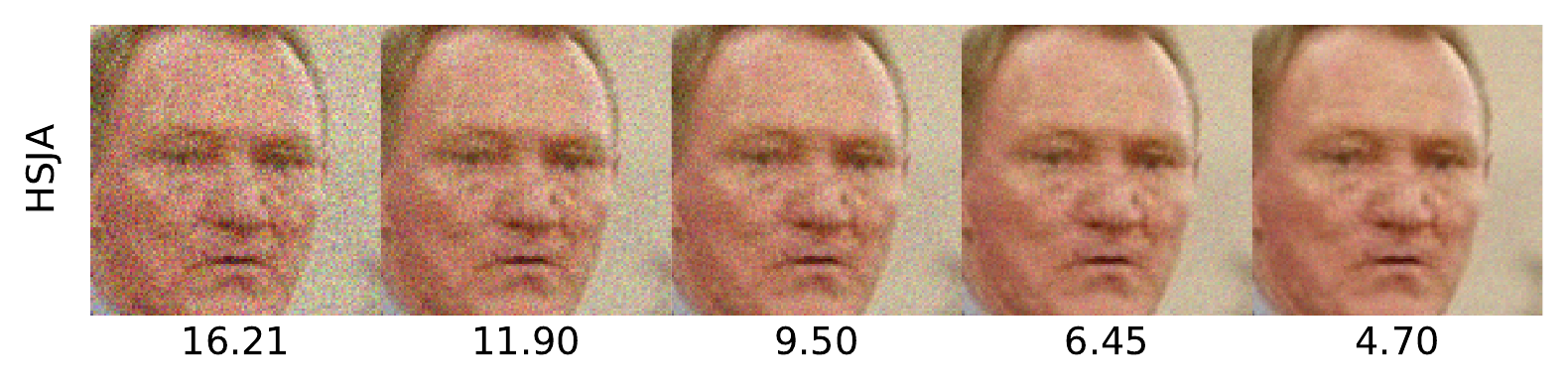}         \includegraphics[width=\textwidth,trim={0cm 0.3cm 0cm 0cm},clip]{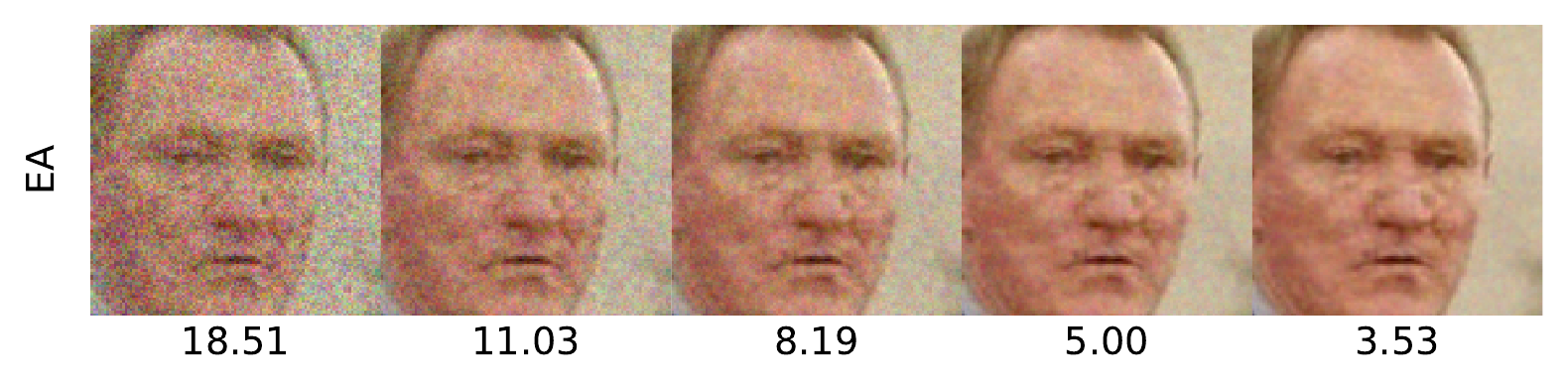}
         \includegraphics[width=\textwidth,trim={0cm 0.3cm 0cm 0cm},clip]{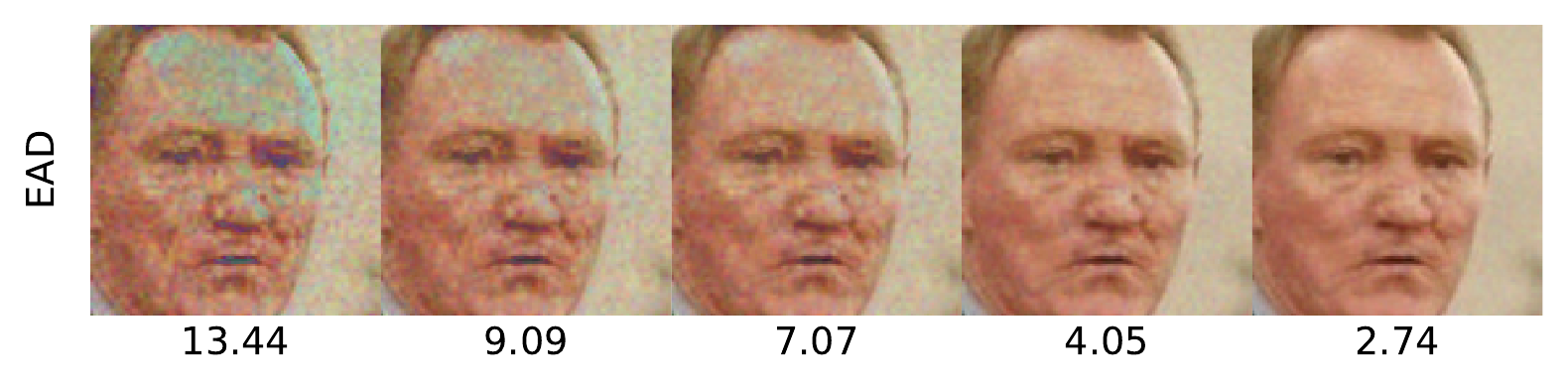}
          \includegraphics[width=\textwidth,trim={0cm 0.3cm 0cm 0cm},clip]{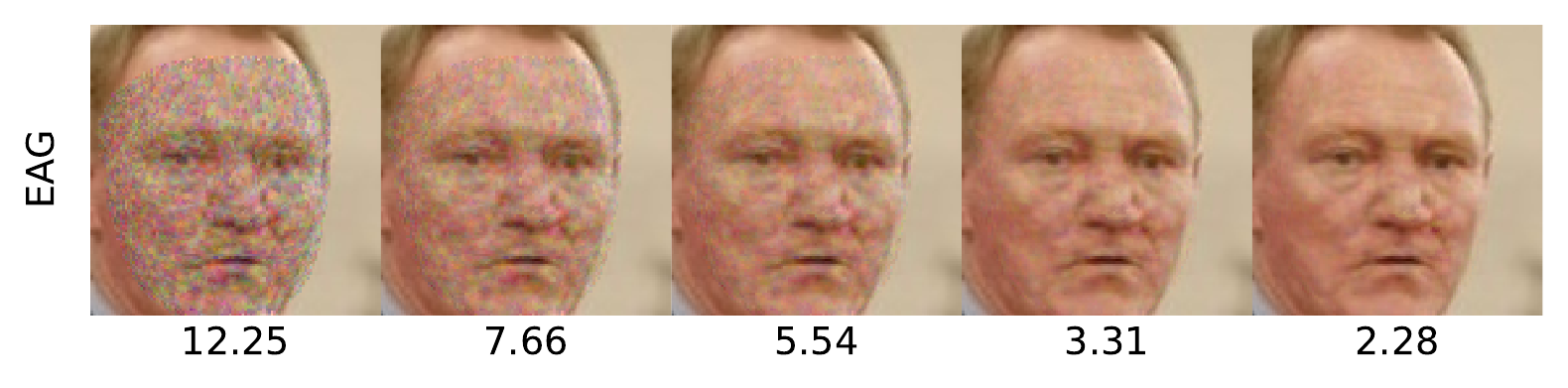}          \includegraphics[width=\textwidth,trim={0cm 0.3cm 0cm 0cm},clip]{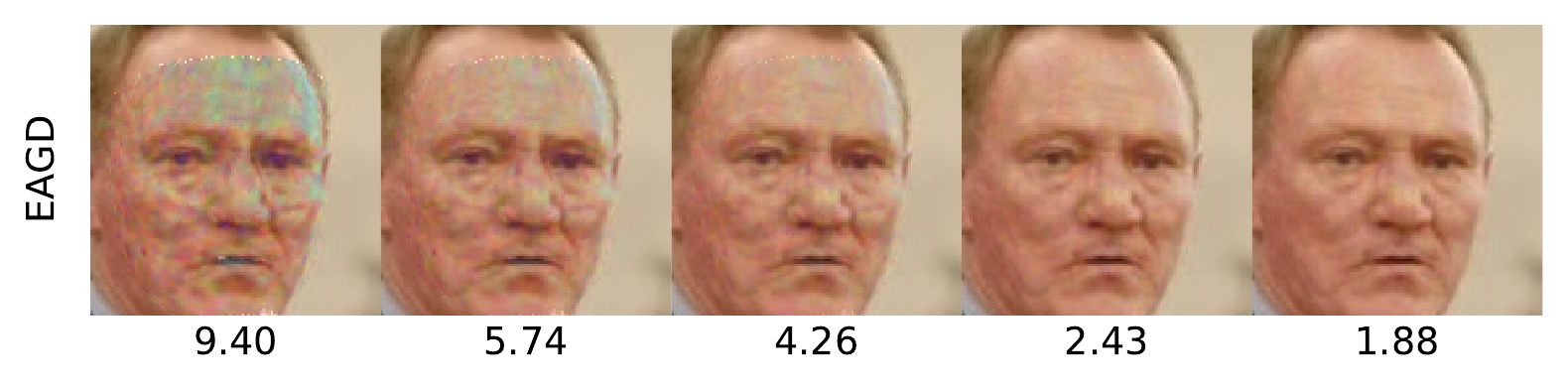}
              \includegraphics[width=\textwidth,trim={0cm 0.3cm 0cm 0cm},clip]{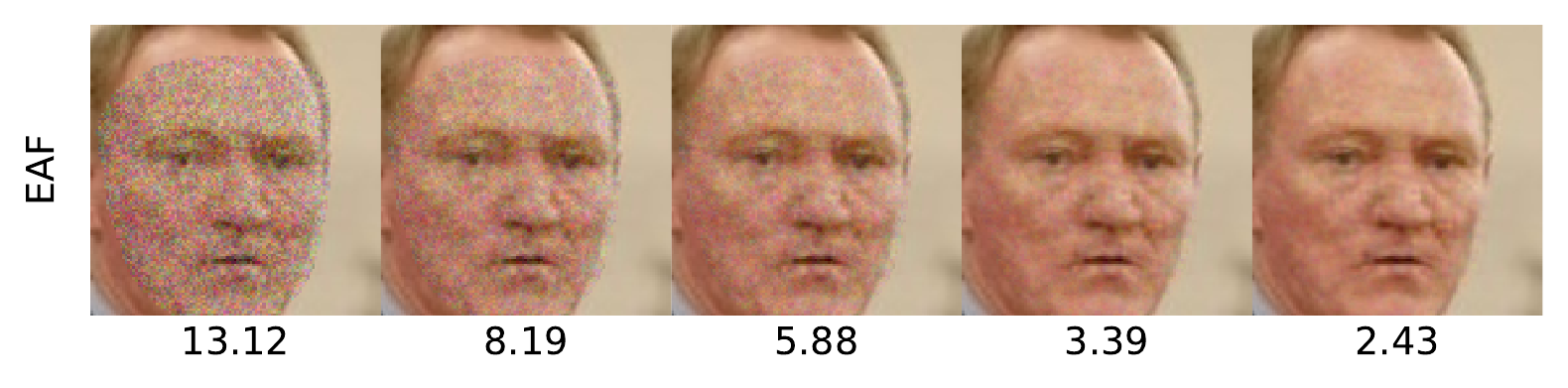}          \includegraphics[width=\textwidth,trim={0cm 0.3cm 0cm 0cm},clip]{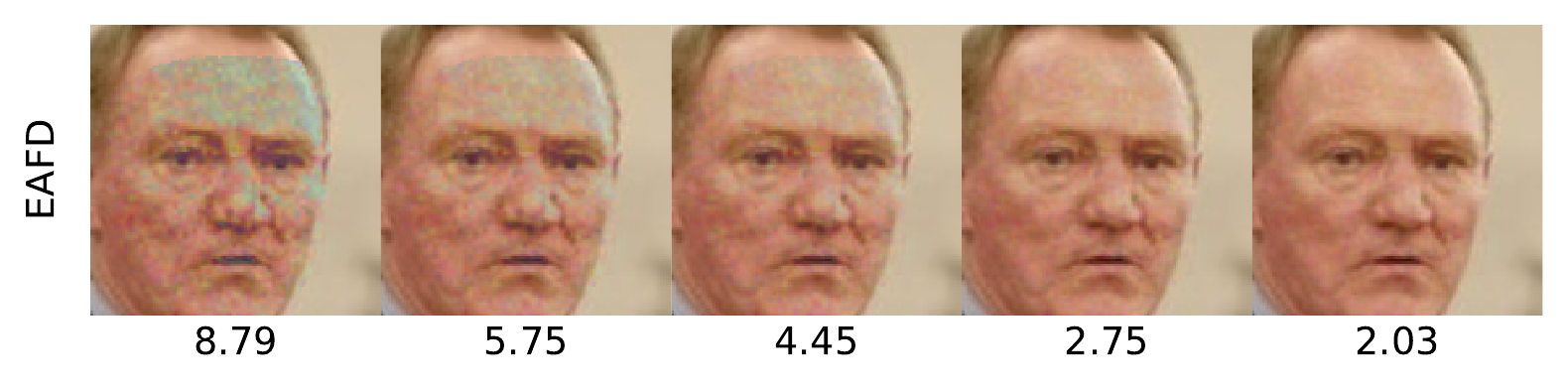}
     \end{subfigure}
 
        %\vspace{-0.2cm}
        \caption{Qualitative results of dodging attacks on the CPLFW dataset \cite{CPLFWTech}. For each attack, we illustrate the minimum norm-adversarial examples in each query budget. The $\ell_2$ norm of perturbation is displayed under each image.}
        \label{fig:dodging_cplfw4}
\end{figure}

\begin{figure}[t]
     \centering
     \begin{subfigure}[b]{0.47\textwidth}
         \centering
         \includegraphics[width=\textwidth,trim={0cm 0.3cm 0cm 0cm},clip]{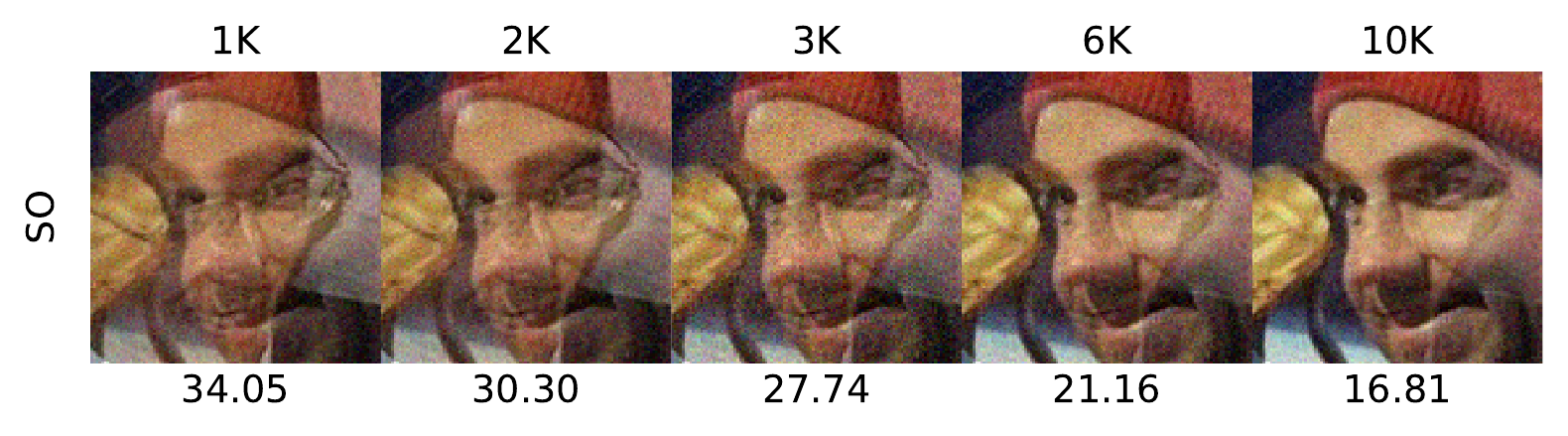}
         \includegraphics[width=\textwidth,trim={0cm 0.3cm 0cm 0cm},clip]{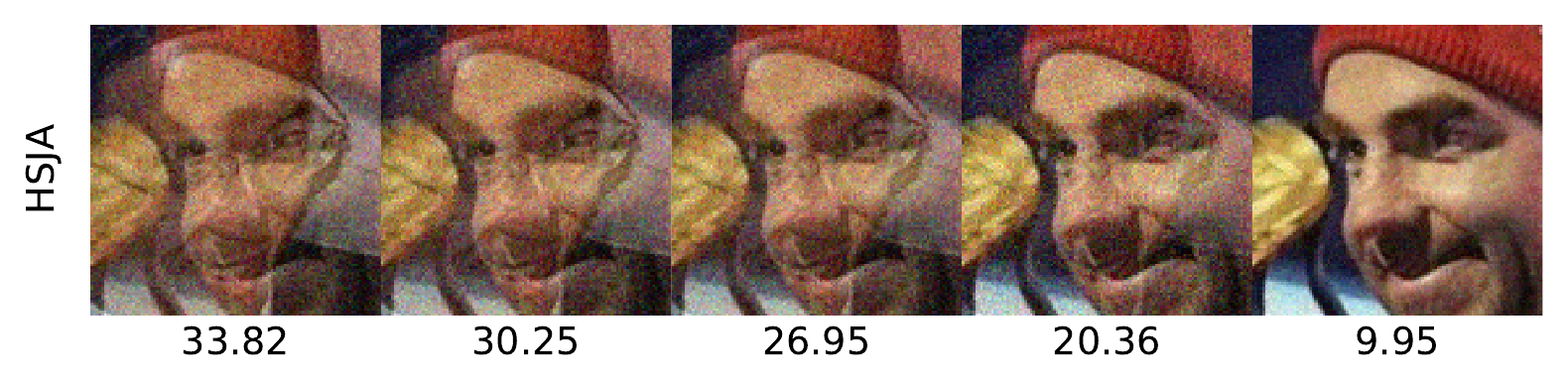}
         \includegraphics[width=\textwidth,trim={0cm 0.3cm 0cm 0cm},clip]{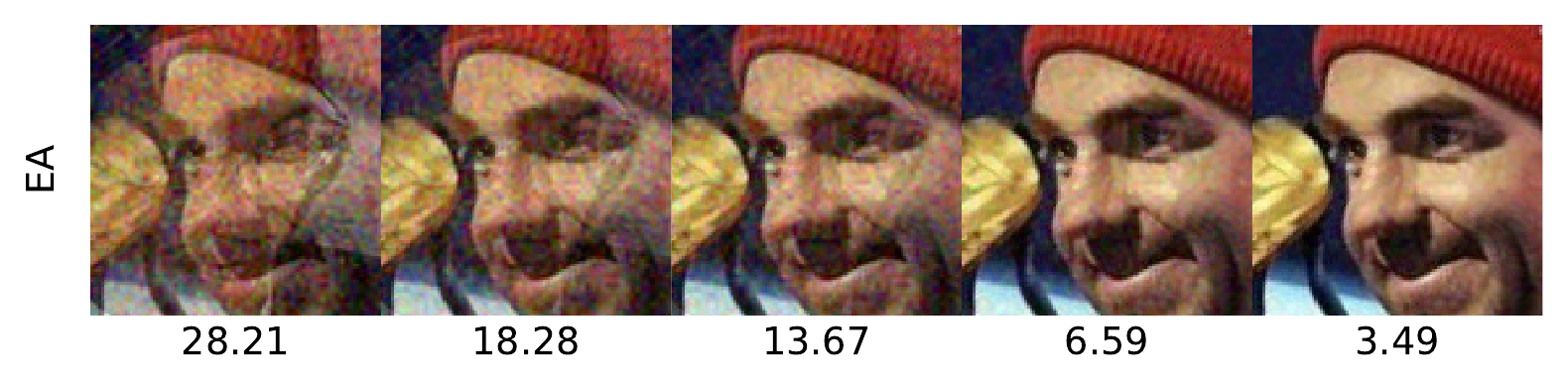}
         \includegraphics[width=\textwidth,trim={0cm 0.3cm 0cm 0cm},clip]{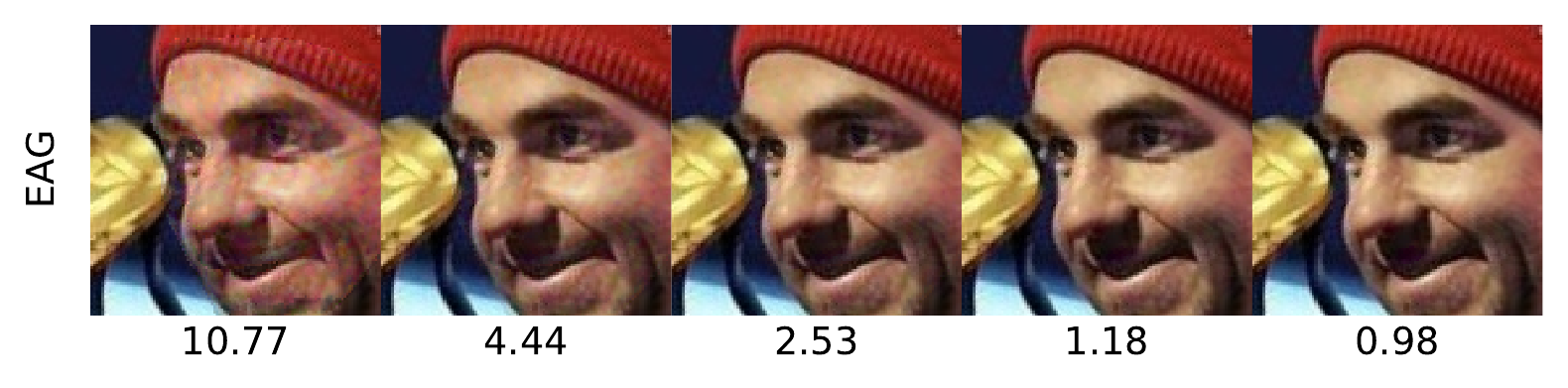}
                  \includegraphics[width=\textwidth,trim={0cm 0.3cm 0cm 0cm},clip]{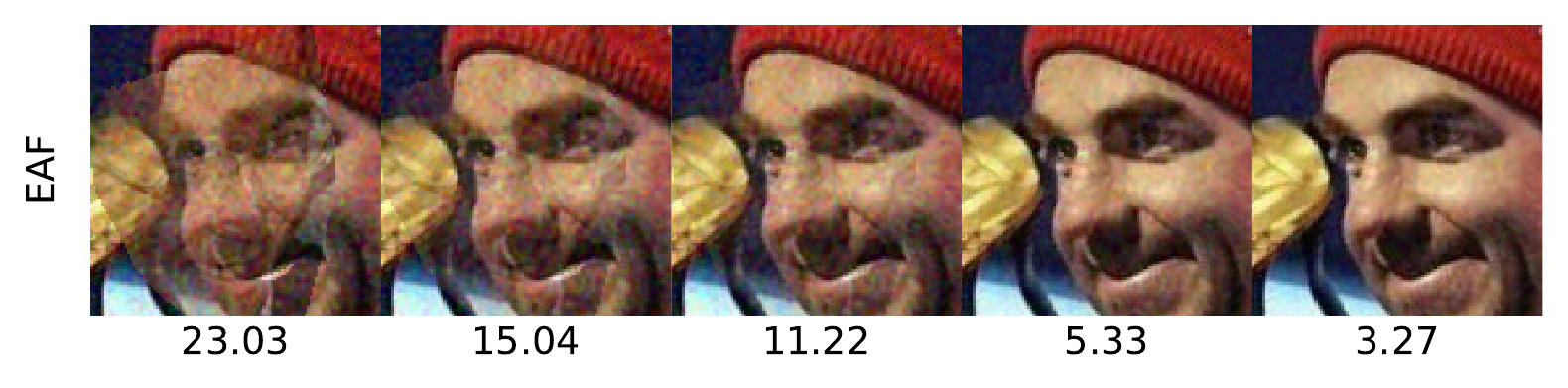}
     \end{subfigure}
        %\vspace{-0.2cm}
        \caption{Qualitative results of impersonation attacks on the CPLFW dataset \cite{CPLFWTech}. For each attack, we illustrate the minimum norm-adversarial examples in each query budget. The $\ell_2$ norm of perturbation is displayed under each image.}
        \label{fig:impersonation_cplfw1}
\end{figure}

\begin{figure}[t]
     \centering
     \begin{subfigure}[b]{0.47\textwidth}
         \centering
         \includegraphics[width=\textwidth,trim={0cm 0.3cm 0cm 0cm},clip]{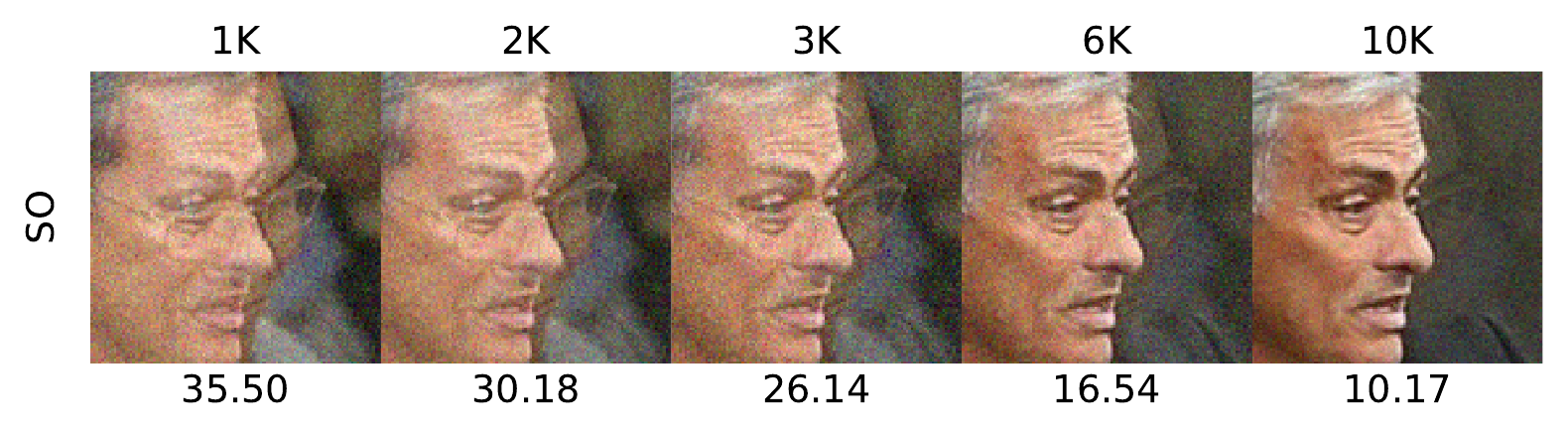}
         \includegraphics[width=\textwidth,trim={0cm 0.3cm 0cm 0cm},clip]{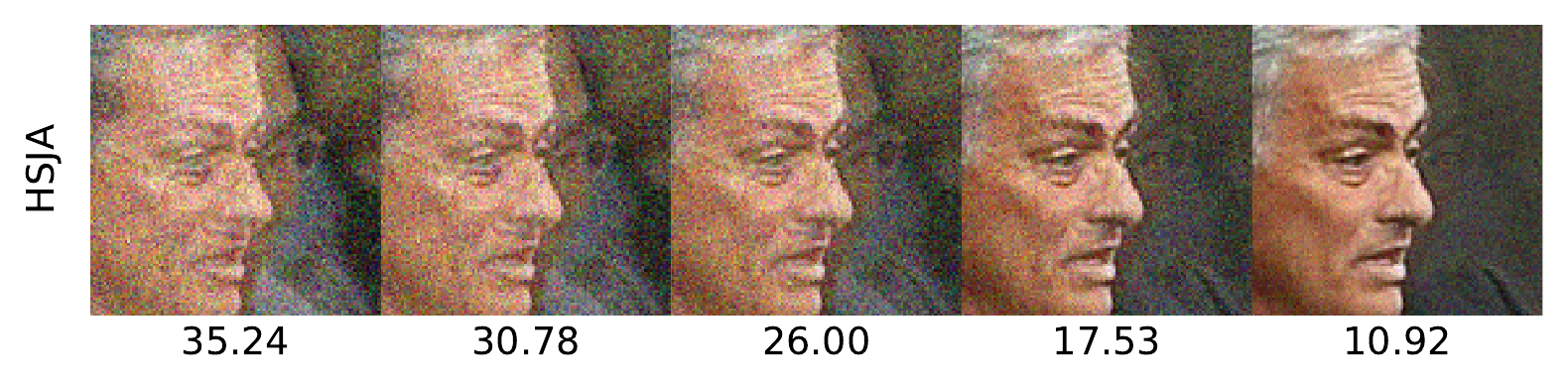}
         \includegraphics[width=\textwidth,trim={0cm 0.3cm 0cm 0cm},clip]{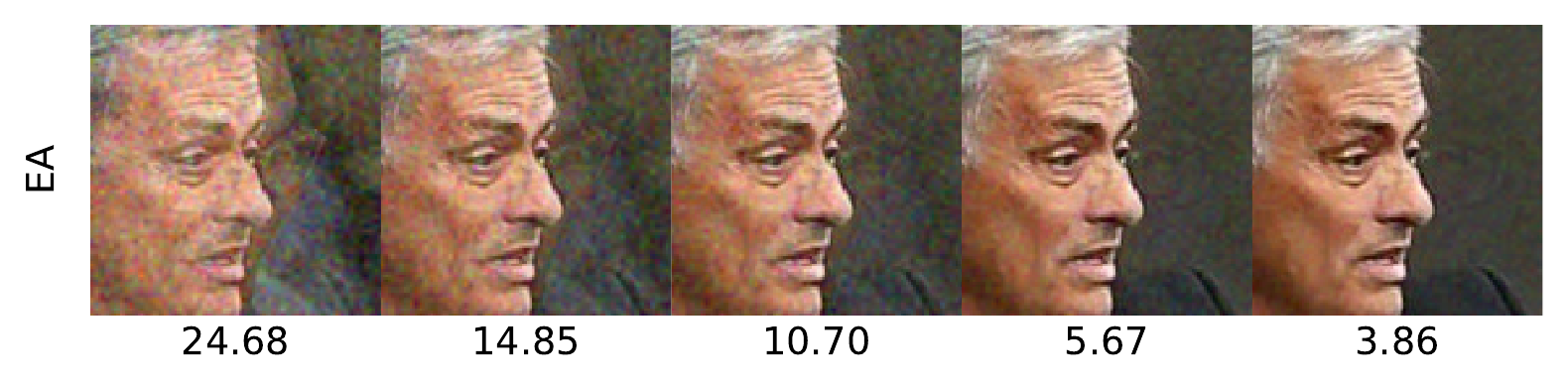}
         \includegraphics[width=\textwidth,trim={0cm 0.3cm 0cm 0cm},clip]{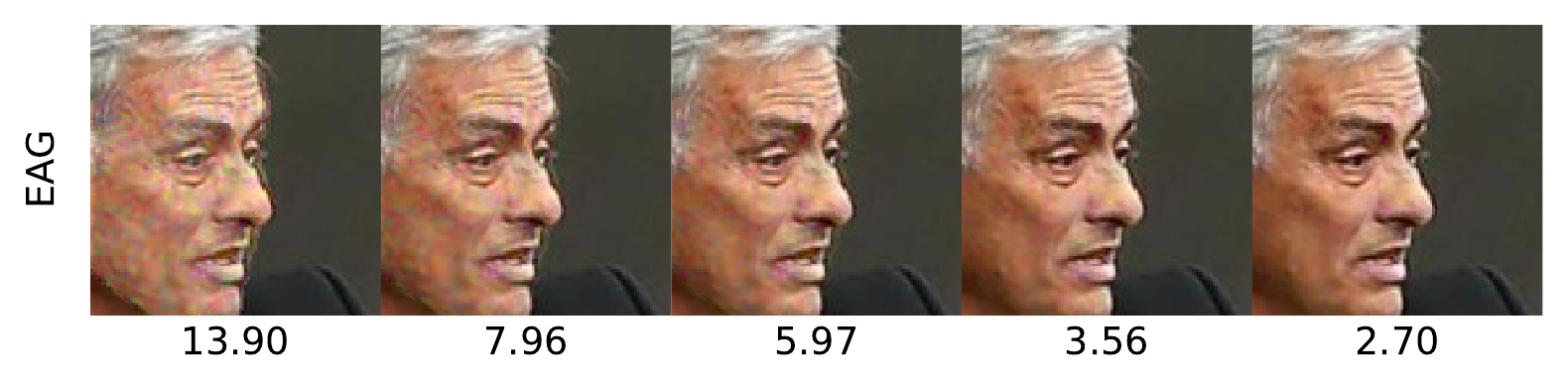}
              \includegraphics[width=\textwidth,trim={0cm 0.3cm 0cm 0cm},clip]{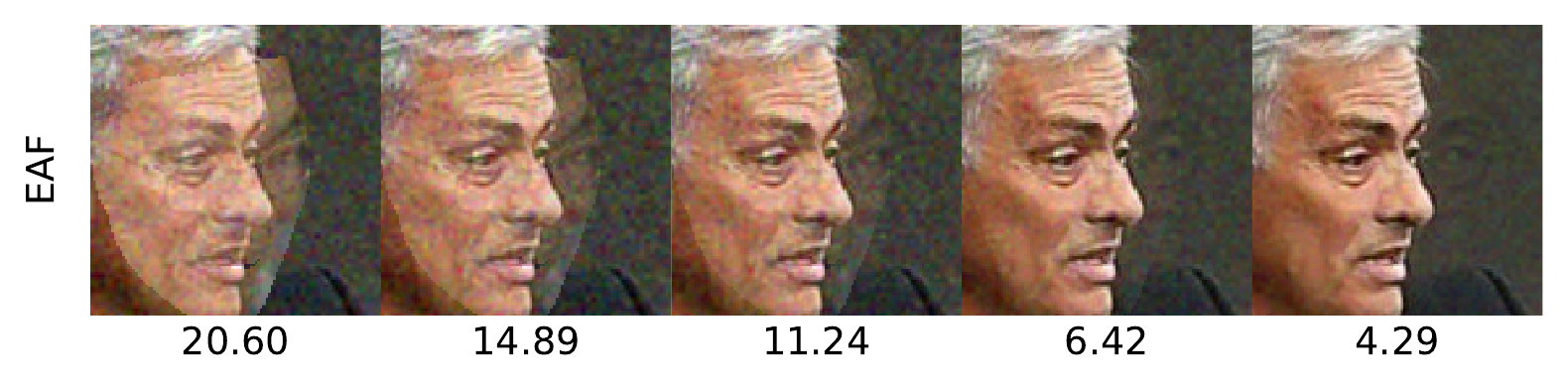}
     \end{subfigure}
        %\vspace{-0.2cm}
        \caption{Qualitative results of impersonation attacks on the CPLFW dataset \cite{CPLFWTech}. For each attack, we illustrate the minimum norm-adversarial examples in each query budget. The $\ell_2$ norm of perturbation is displayed under each image.}
        \label{fig:impersonation_cplfw2}
\end{figure}

\begin{figure}[t]
     \centering
     \begin{subfigure}[b]{0.47\textwidth}
         \centering
         \includegraphics[width=\textwidth,trim={0cm 0.3cm 0cm 0cm},clip]{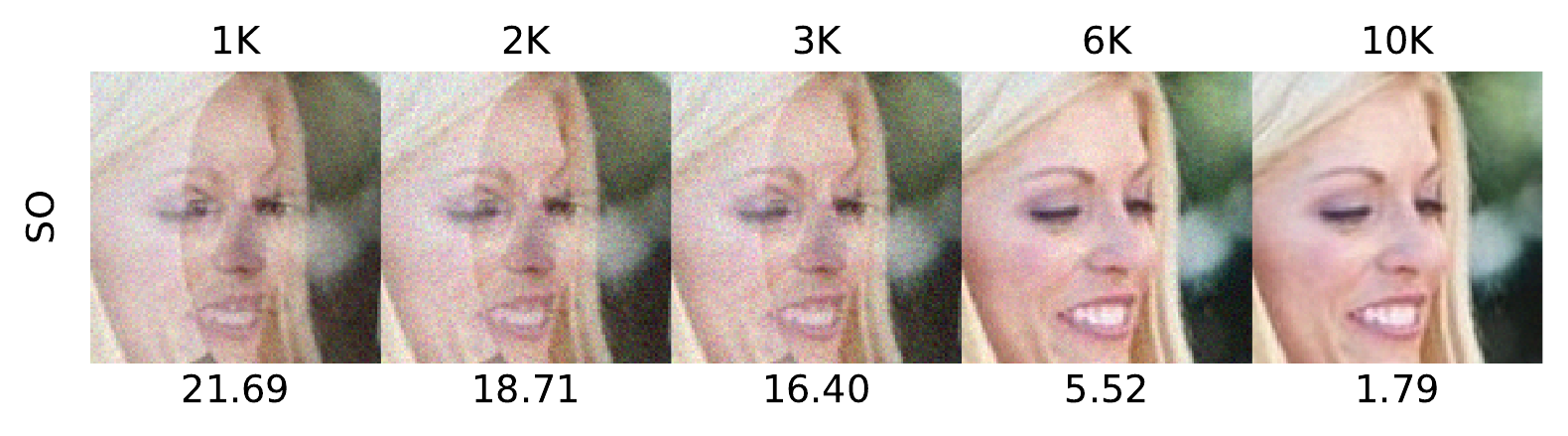}
         \includegraphics[width=\textwidth,trim={0cm 0.3cm 0cm 0cm},clip]{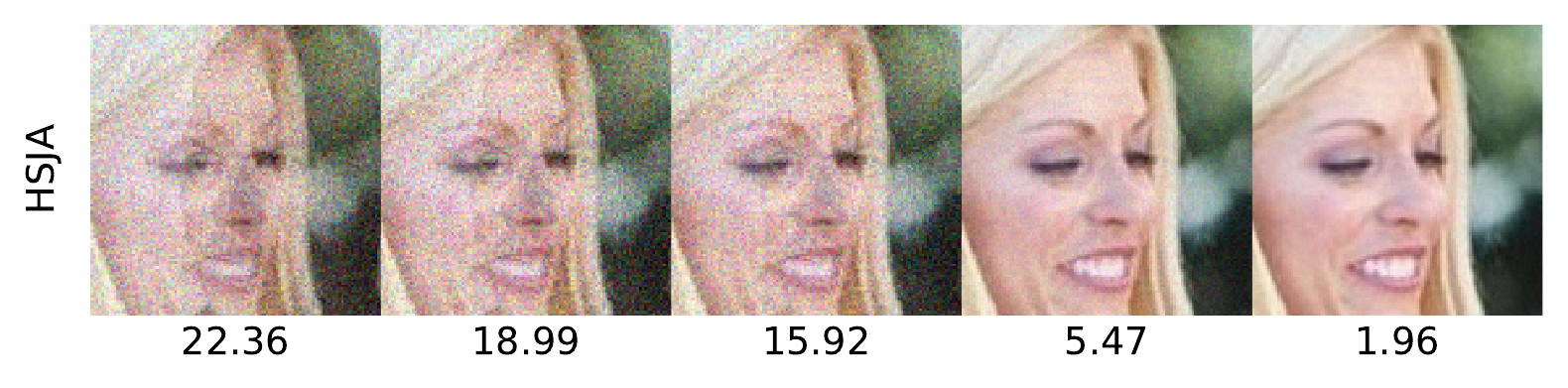}
         \includegraphics[width=\textwidth,trim={0cm 0.3cm 0cm 0cm},clip]{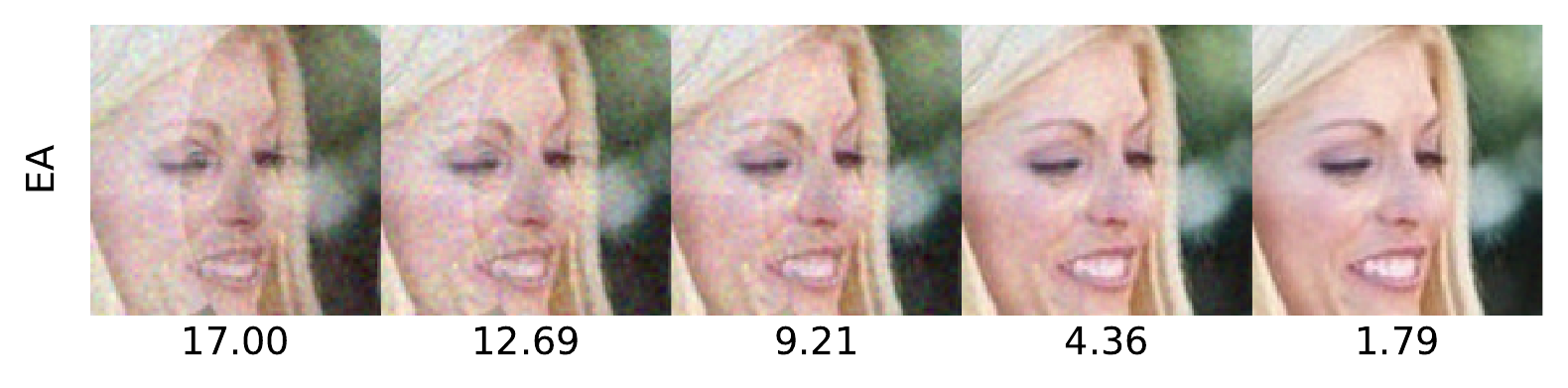}
         \includegraphics[width=\textwidth,trim={0cm 0.3cm 0cm 0cm},clip]{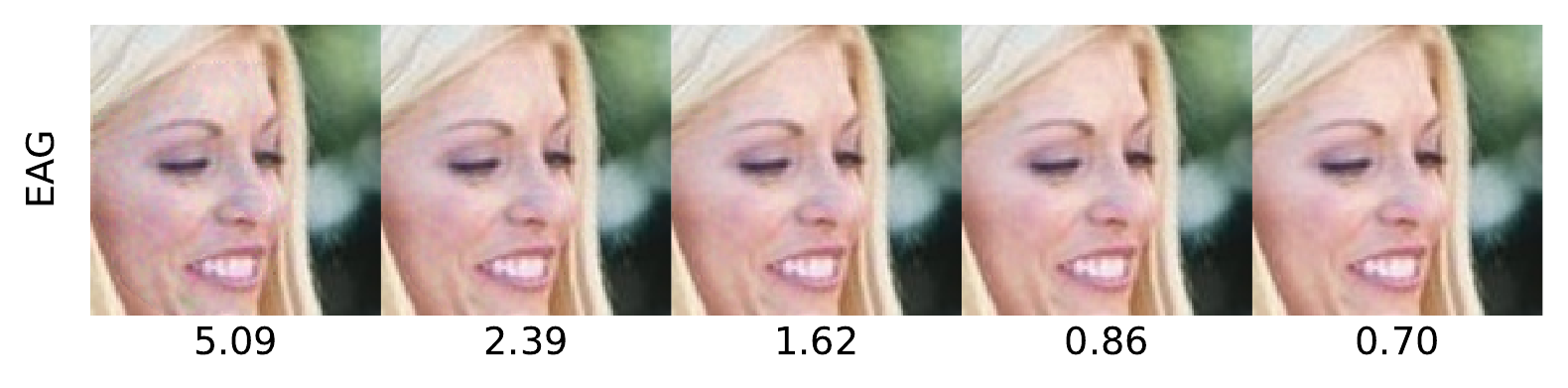}
             \includegraphics[width=\textwidth,trim={0cm 0.3cm 0cm 0cm},clip]{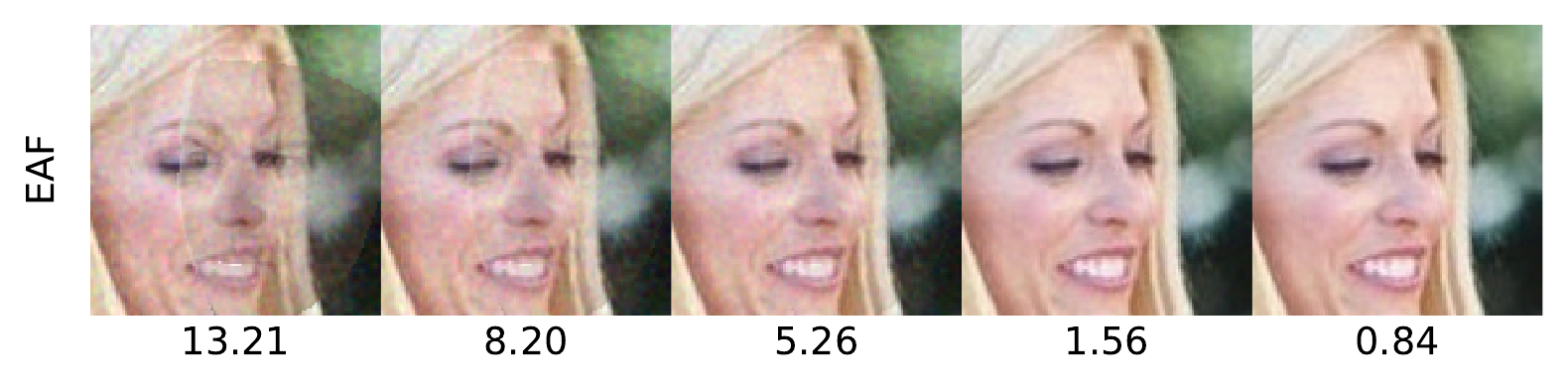}
     \end{subfigure}
        %\vspace{-0.2cm}
        \caption{Qualitative results of impersonation attacks on the CPLFW dataset \cite{CPLFWTech}. For each attack, we illustrate the minimum norm-adversarial examples in each query budget. The $\ell_2$ norm of perturbation is displayed under each image.}
        \label{fig:impersonation_cplfw3}
\end{figure}

\begin{figure}[t]
     \centering
     \begin{subfigure}[b]{0.47\textwidth}
         \centering
         \includegraphics[width=\textwidth,trim={0cm 0.3cm 0cm 0cm},clip]{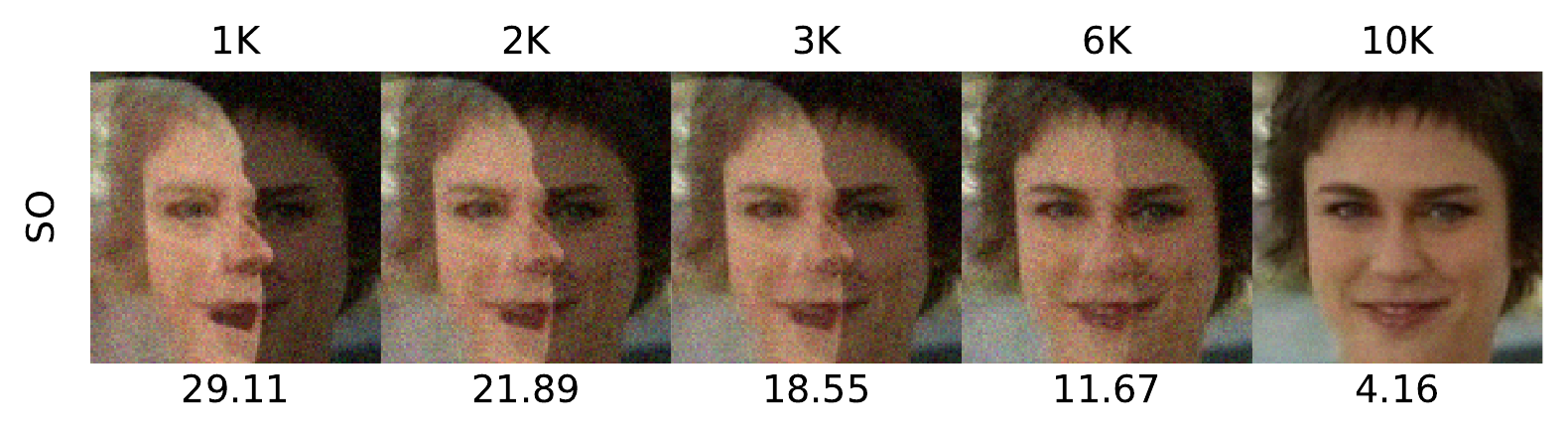}
         \includegraphics[width=\textwidth,trim={0cm 0.3cm 0cm 0cm},clip]{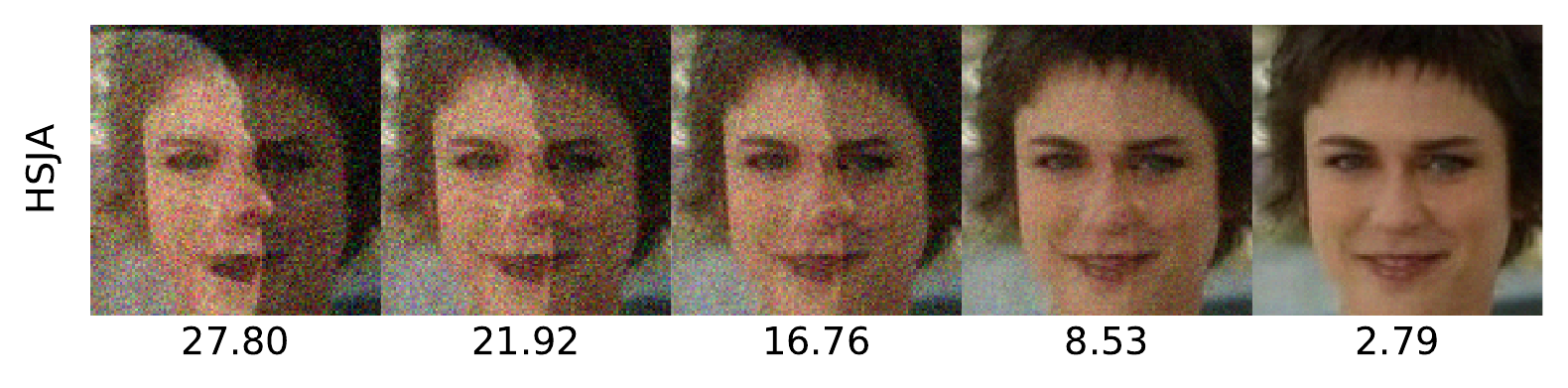}
         \includegraphics[width=\textwidth,trim={0cm 0.3cm 0cm 0cm},clip]{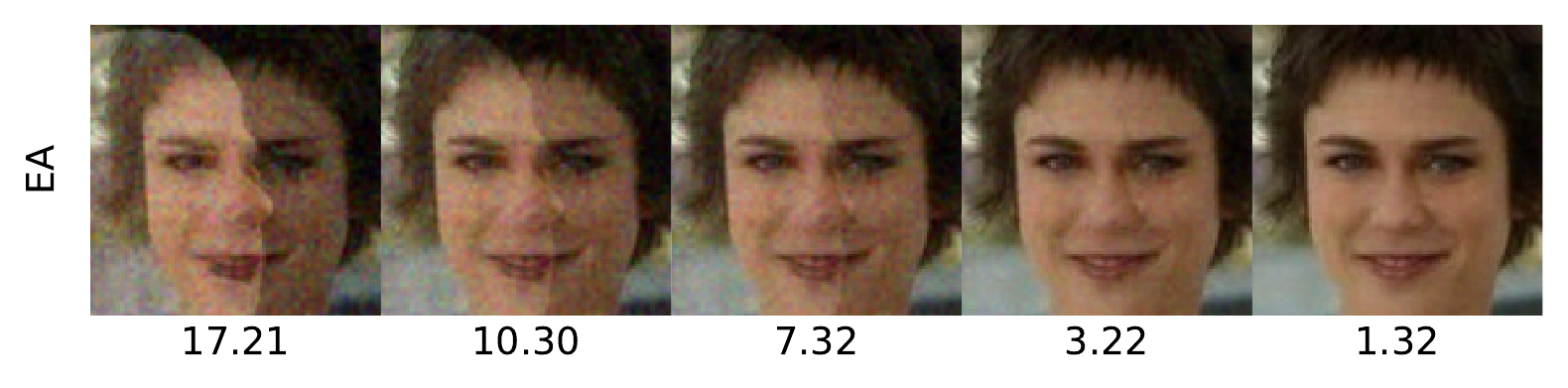}
         \includegraphics[width=\textwidth,trim={0cm 0.3cm 0cm 0cm},clip]{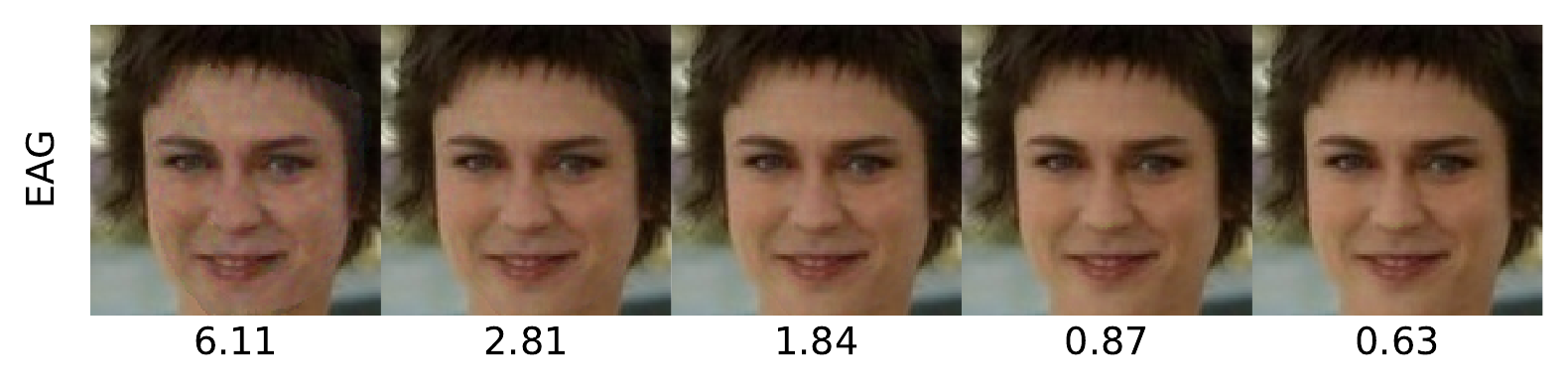}
         \includegraphics[width=\textwidth,trim={0cm 0.3cm 0cm 0cm},clip]{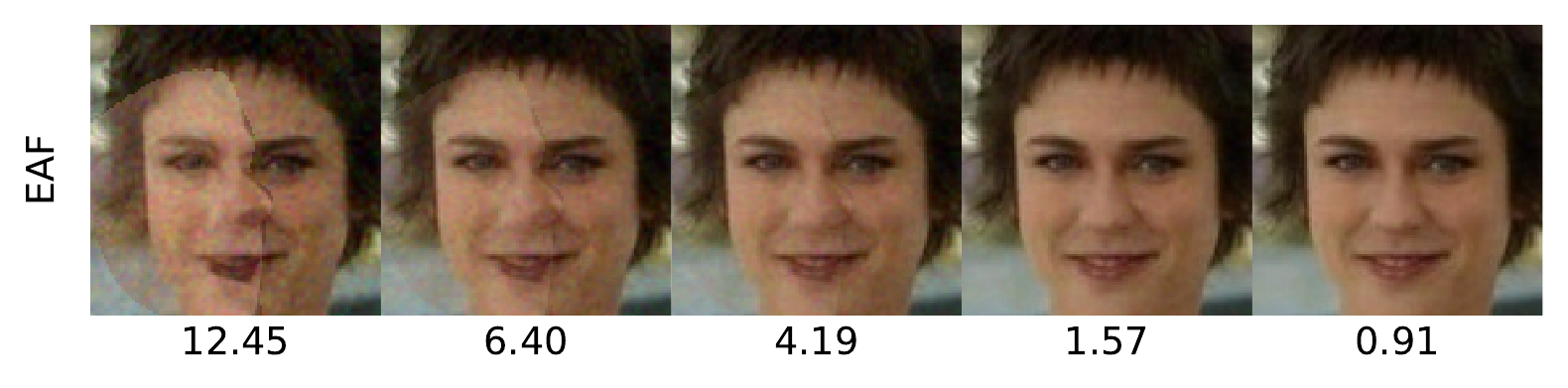}
     \end{subfigure}
        %\vspace{-0.2cm}
        \caption{Qualitative results of impersonation attacks on the CPLFW dataset \cite{CPLFWTech}. For each attack, we illustrate the minimum norm-adversarial examples in each query budget. The $\ell_2$ norm of perturbation is displayed under each image.}
        \label{fig:impersonation_cplfw4}
\end{figure}

\begin{figure}[t]
     \centering
     \begin{subfigure}[b]{0.47\textwidth}
         \centering
         \includegraphics[width=\textwidth,trim={0cm 0.3cm 0cm 0cm},clip]{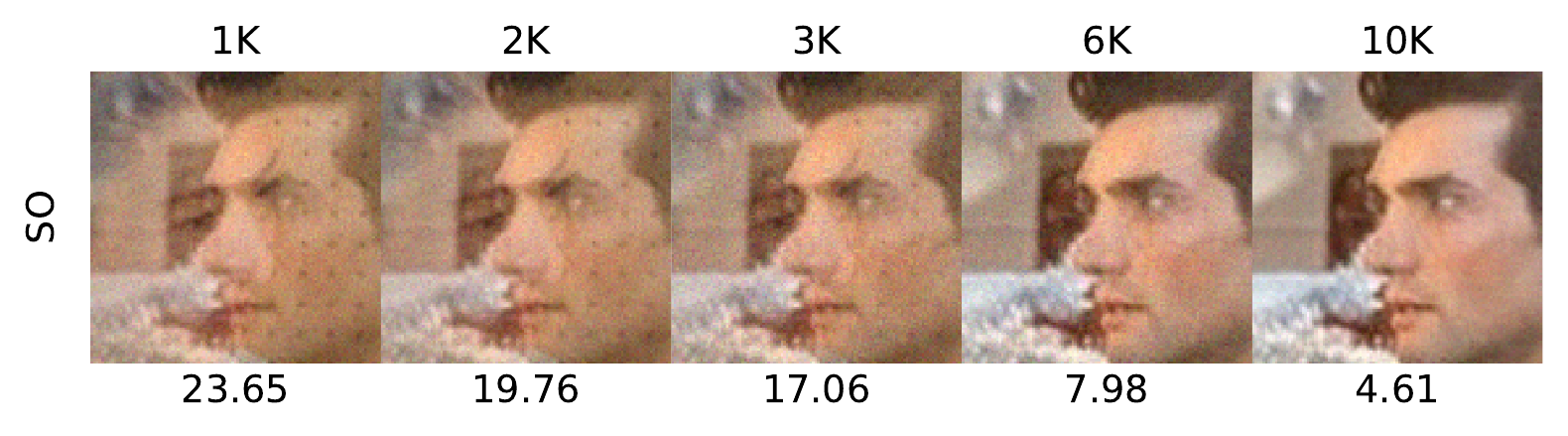}
         \includegraphics[width=\textwidth,trim={0cm 0.3cm 0cm 0cm},clip]{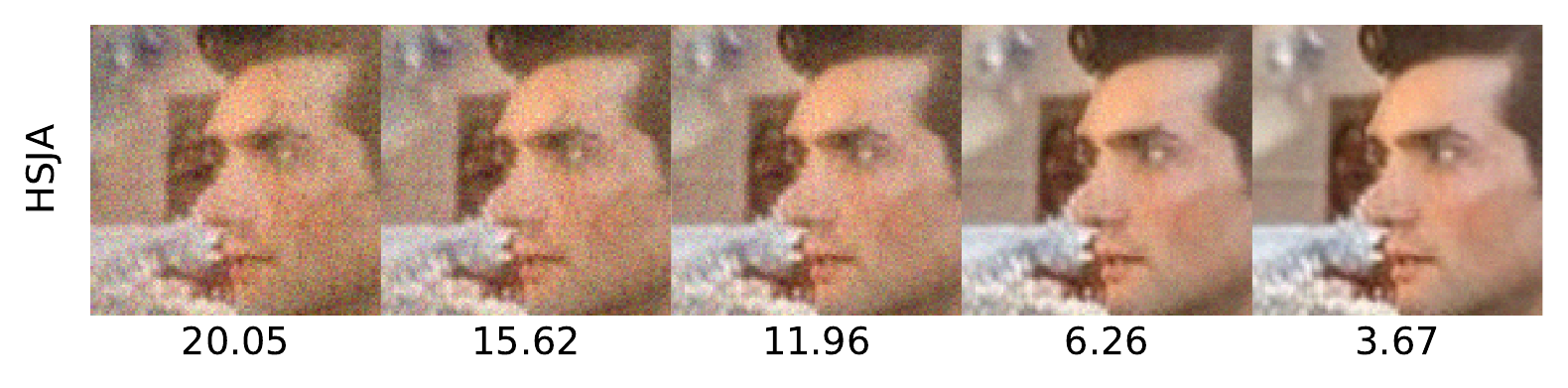}
         \includegraphics[width=\textwidth,trim={0cm 0.3cm 0cm 0cm},clip]{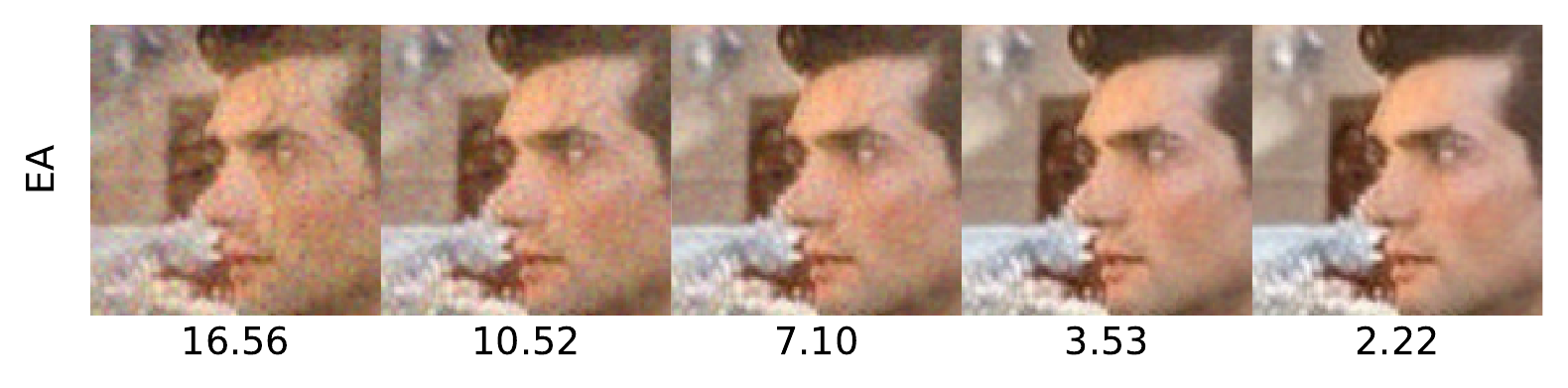}
         \includegraphics[width=\textwidth,trim={0cm 0.3cm 0cm 0cm},clip]{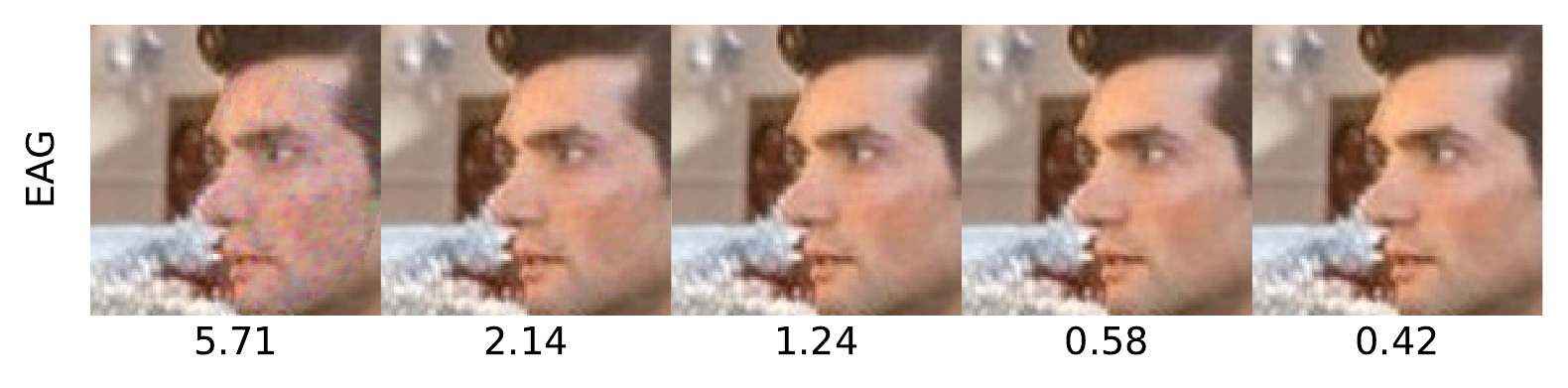}
              \includegraphics[width=\textwidth,trim={0cm 0.3cm 0cm 0cm},clip]{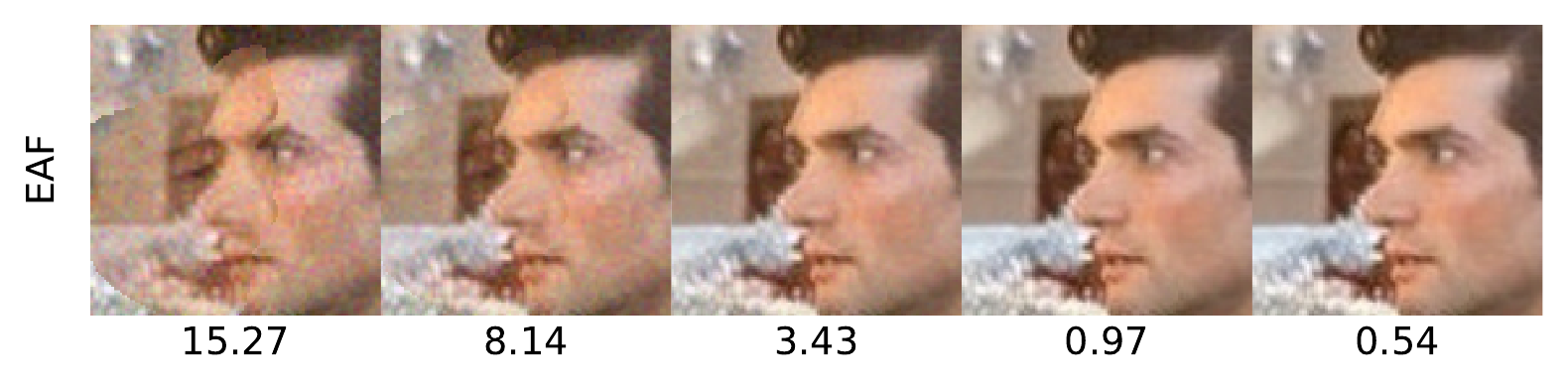}
     \end{subfigure}
        %\vspace{-0.2cm}
        \caption{Qualitative results of impersonation attacks on the CPLFW dataset \cite{CPLFWTech}. For each attack, we illustrate the minimum norm-adversarial examples in each query budget. The $\ell_2$ norm of perturbation is displayed under each image.}
        \label{fig:impersonation_cplfw5}
\end{figure}

We illustrate the perturbation norm curves that visually show the query efficiency of each attack method in Fig. \ref{fig:ir50_lfw} and Fig. \ref{fig:ir50_cplfw}. We show additional examples for more extensive qualitative comparison in Fig. \ref{fig:dodging_lfw1} to Fig. \ref{fig:impersonation_cplfw5}. In detail, Fig. \ref{fig:dodging_lfw1} and Fig. \ref{fig:dodging_lfw2} show the results of dodging attacks on the LFW dataset. Fig. \ref{fig:impersonation_lfw1} through Fig. \ref{fig:impersonation_lfw4} display the examples of impersonation attacks on the LFW dataset. 
 Fig. \ref{fig:dodging_cplfw1} to Fig. \ref{fig:dodging_cplfw4} show the results of dodging attacks on the CPLFW dataset.  Fig. \ref{fig:impersonation_cplfw1} to Fig. \ref{fig:impersonation_cplfw5} show the examples of impersonation attacks on the CPLFW dataset.

\section{Experimental results on the CurricularFace ResNet-100 model}
We conduct the decision-based black-box attacks against the CurricularFace ResNet-100\footnote{We use the pretrained model from \url{https://github.com/HuangYG123/CurricularFace}} \cite{huang2020curricularface, he2016deep} which is trained on the refined MS1MV2 dataset \cite{deng2019arcface}. We arrange the experimental results in Table \ref{table:curricularface}. 
Clearly, GADA greatly improves the query efficiency of EA. 

\begin{figure}[t]
    \centering
     \begin{subfigure}[b]{0.47\textwidth}
         \centering
         \includegraphics[width=\textwidth]{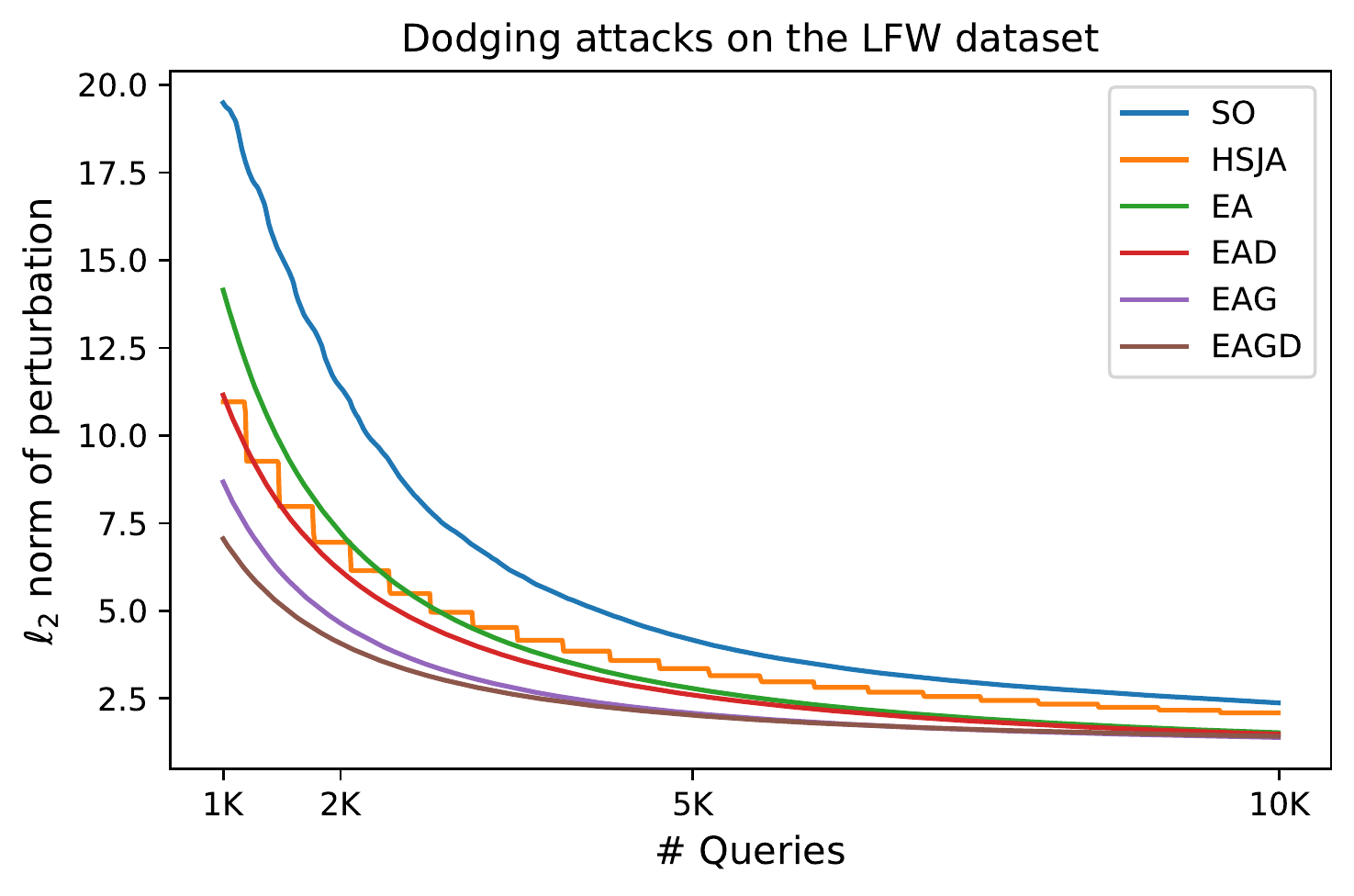}
     \end{subfigure}
          \begin{subfigure}[b]{0.47\textwidth}
         \centering
         \includegraphics[width=\textwidth]{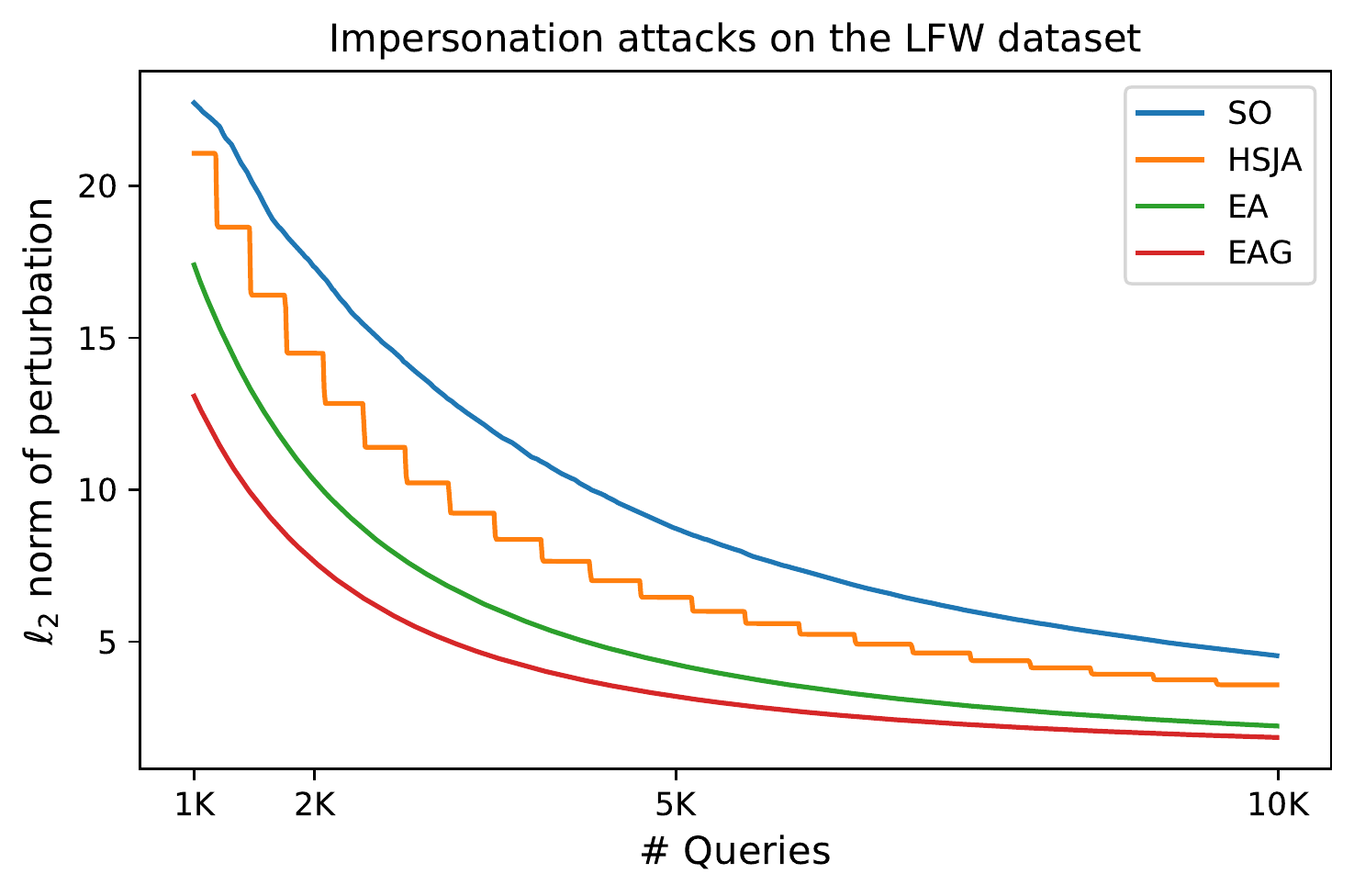}
     \end{subfigure}
        \caption{Perturbation norm curves of the decision-based attacks against the CurricularFace ResNet-100 with the LFW dataset \cite{LFWTech}.
        }
    \label{fig:curr_lfw}
\end{figure}

\begin{figure}[t]
    \centering
     \begin{subfigure}[b]{0.47\textwidth}
         \centering
         \includegraphics[width=\textwidth]{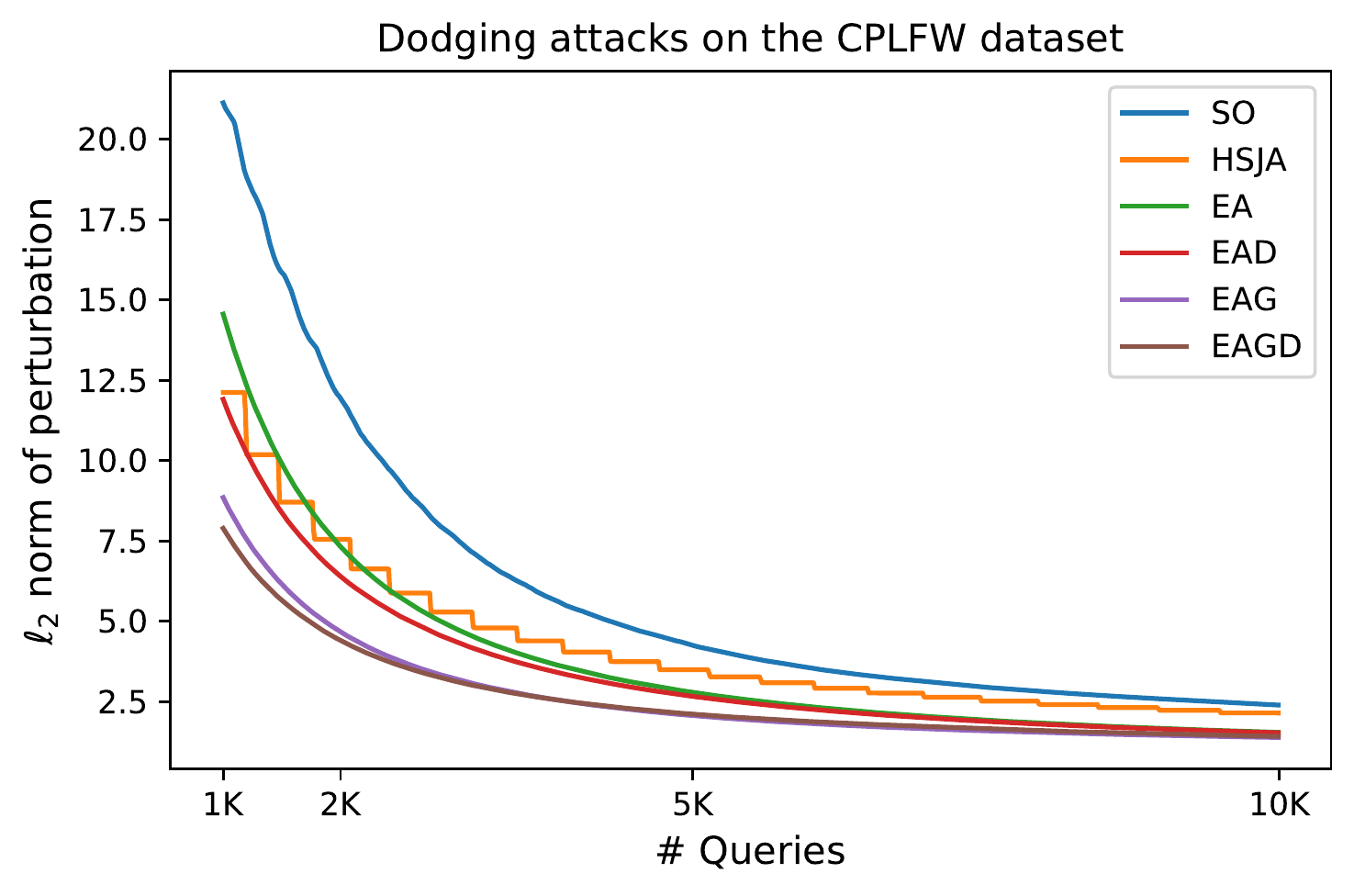}
     \end{subfigure}
         \begin{subfigure}[b]{0.47\textwidth}
         \centering
         \includegraphics[width=\textwidth]{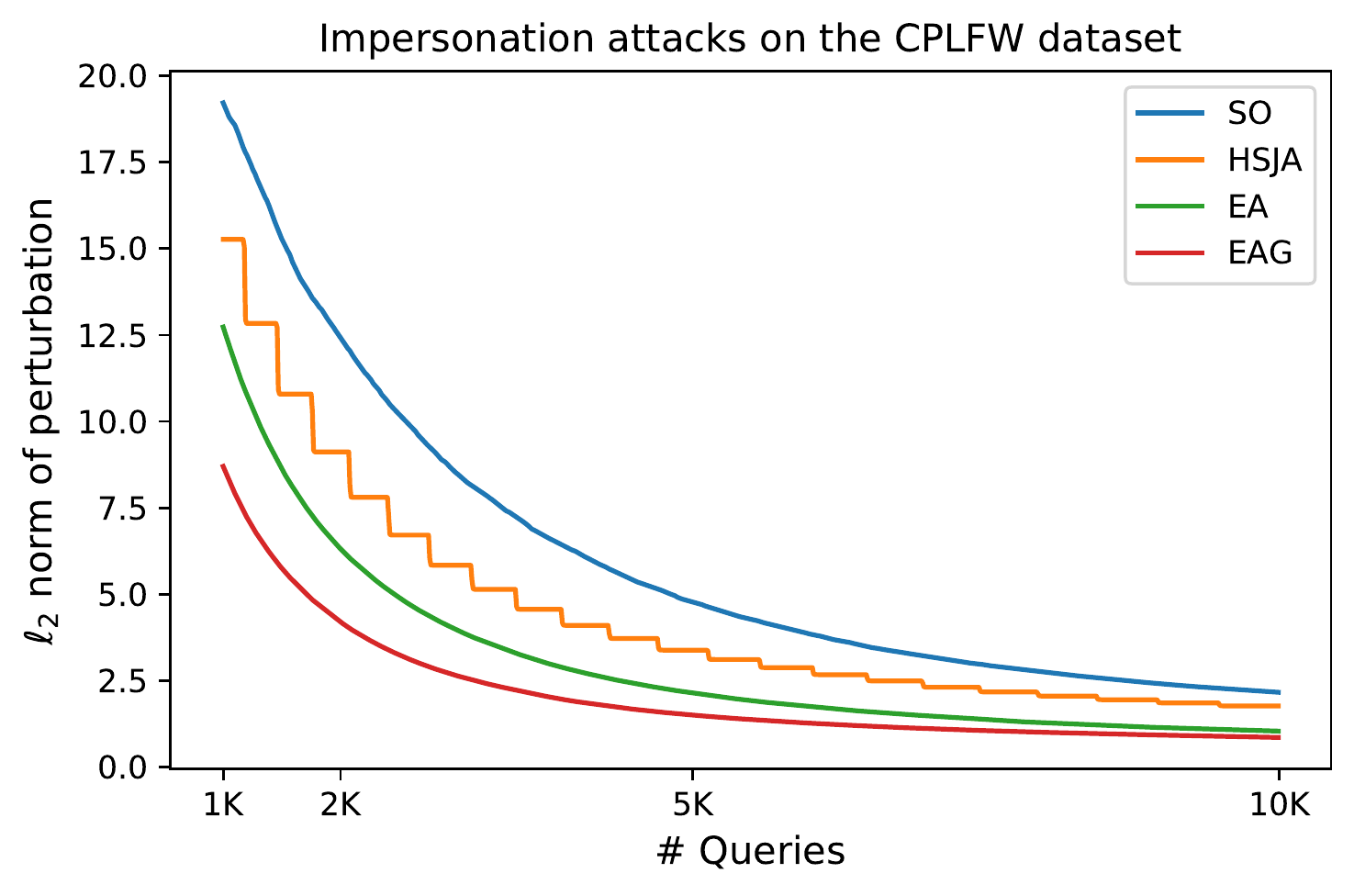}
     \end{subfigure}
        \caption{Perturbation norm curves of the decision-based attacks against the CurricularFace ResNet-100 with the CPLFW dataset \cite{CPLFWTech}.
        }
    \label{fig:curr_cplfw}
\end{figure}

As the model is deeper and more accurate, the robustness to attack increases, so the minimum perturbation norm is higher in all datasets than that of the results of the ArcFace ResNet-50 model \cite{deng2019arcface, he2016deep}. We illustrate the perturbation norm curves that visually show each attack method's query efficiency in Fig. \ref{fig:curr_lfw} and Fig. \ref{fig:curr_cplfw}.

%%%%%%%%%%%%%%%%%%%%%%%%%%%%%%%%%%%




\newcolumntype{P}[1]{>{\centering\arraybackslash}p{#1}}
\setlength{\tabcolsep}{4pt} % Default value: 6pt
\begin{table*}[t]
	\centering
	\begin{tabular}{lcccc P{1.5cm}P{1.5cm} cccc P{1.5cm}P{1.5cm}}
		\specialrule{.1em}{.05em}{.05em} 
		\multicolumn{1}{c}{} & \multicolumn{12}{c}{Dodging attacks} \\
		& \multicolumn{6}{c}{LFW dataset \cite{LFWTech}} & \multicolumn{6}{c}{CPLFW dataset \cite{CPLFWTech}} \\ 
		\specialrule{.1em}{.05em}{.05em} 
		%\\  finding $\bm{\delta}$ with norm
		% \\ norm with query budget
		% \\ finding $\bm{\delta}$ with norm
		& \multicolumn{4}{c}{\begin{tabular}[c]{@{}c@{}}Minimum perturbation\\ norm with query budget\end{tabular}} & \multicolumn{2}{c}{\begin{tabular}[c]{@{}c@{}}Avg. \# queries for\\  perturbation with norm\end{tabular}} & \multicolumn{4}{c}{\begin{tabular}[c]{@{}c@{}}Minimum perturbation\\ norm with query budget \end{tabular}} & \multicolumn{2}{c}{\begin{tabular}[c]{@{}c@{}}Avg. \# queries for\\ perturbation with norm \end{tabular}} \\
		Attack method & 1K & 2K & 5K & 10K & 4 & 2 & 1K & 2K & 5K & 10K & 4 & 2 \\ \hline
		Sign-OPT \cite{cheng2019sign} &  19.50 & 11.38 & 4.17 & 2.37 & 5284 & 8678 & 21.14 & 11.95 & 4.24 & 2.39 & 5206 & 8256  \\
		HSJA \cite{chen2020hopskipjumpattack}&  10.96 & 6.96 & 3.35 & 2.09 & 3990 & 7796 & 12.12 & 7.55 & 3.49 & 2.15 & 4090 & 7362 \\
		EA \cite{dong2019efficient}&  14.16 & 7.23 & 2.78 & 1.52 & 3550 & 6802 &  14.57 & 7.33 & 2.79 & 1.54 & 3486 & 6419  \\
		EAD &11.16 & 6.15 & 2.60 & 1.48 & 3195 & 6538 & 11.92 & 6.40 & 2.66 & 1.53 & 3236 & 6255   \\
		EAG & 8.68 & 4.65 & 2.08 & \textbf{1.39} & 2424 & 5494 &  8.86 & 4.67 & \textbf{2.06} & \textbf{1.38} & 2407 & 5101 \\
		EAGD &  \textbf{7.06} & \textbf{4.07} & \textbf{2.02} & 1.42 & \textbf{2104} & \textbf{5166} & \textbf{7.90} & \textbf{4.40} & 2.11 & 1.42 & \textbf{2251} & \textbf{4954}  \\
		\specialrule{.1em}{.05em}{.05em} 
		
		\multicolumn{1}{c}{} & \multicolumn{12}{c}{Impersonation attacks} \\
		& \multicolumn{6}{c}{LFW dataset \cite{LFWTech}} & \multicolumn{6}{c}{CPLFW dataset \cite{CPLFWTech}} \\
		\specialrule{.1em}{.05em}{.05em} 
		& \multicolumn{4}{c}{\begin{tabular}[c]{@{}c@{}}Minimum perturbation\\ norm with query budget\end{tabular}} & \multicolumn{2}{c}{\begin{tabular}[c]{@{}c@{}}Avg. \# queries for\\  perturbation with norm\end{tabular}} & \multicolumn{4}{c}{\begin{tabular}[c]{@{}c@{}}Minimum perturbation\\ norm with query budget \end{tabular}} & \multicolumn{2}{c}{\begin{tabular}[c]{@{}c@{}}Avg. \# queries for\\ perturbation with norm \end{tabular}} \\
		Attack method & 1K & 2K & 5K & 10K & 4 & 2 & 1K & 2K & 5K & 10K & 4 & 2 \\ \hline
		Sign-OPT \cite{cheng2019sign}& 22.74 & 17.33 & 8.73 & 4.53 & 7944 & 9561 &  19.22 & 12.43 & 4.79 & 2.17 & 4876 & 6827  \\
		HSJA \cite{chen2020hopskipjumpattack}& 21.08 & 14.49 & 6.46 & 3.58 & 6709 & 9102 &  15.26 & 9.12 & 3.39 & 1.77 & 3718 & 5799 \\
		EA \cite{dong2019efficient}& 17.41 & 10.29 & 4.24 & 2.22 & 5226 & 8169 & 12.74 & 6.32 & 2.15 & 1.05 & 2931 & 4803 \\
		EAG &  \textbf{13.09} & \textbf{7.63} & \textbf{3.19} & \textbf{1.84} & \textbf{3887} & \textbf{7217} &  \textbf{8.71} & \textbf{4.21} & \textbf{1.51} & \textbf{0.86} & \textbf{1999} & \textbf{3546} \\
		\specialrule{.1em}{.05em}{.05em} 
	\end{tabular}
	\caption{Evaluation of decision-based black-box attacks against the CurricularFace ResNet-100 \cite{huang2020curricularface, he2016deep} with the two datasets.}
	\label{table:curricularface}
\end{table*}

{\small
\bibliographystyle{ieee_fullname}
\bibliography{egbib}
}